%% file: CVPR_GW_Camera.tex
\documentclass[final]{cvpr}

\usepackage{times}
\usepackage{epsfig}
\usepackage{graphicx}
\usepackage{amsmath}
\usepackage{amssymb}

\usepackage{url}            % simple URL typesetting
\usepackage{booktabs}       % professional-quality tables
\usepackage{amsfonts}       % blackboard math symbols
\usepackage{nicefrac}       % compact symbols for 1/2, etc.
\usepackage{microtype}      % microtypography
\usepackage{graphicx} % more modern
\usepackage{subfigure}
\usepackage{color}

\usepackage{algorithm}
\usepackage{algorithmic}
\usepackage{array}
\usepackage{enumitem}
\newtheorem{defn}{Definition}

\newcommand{\TODO}[1]{\textcolor{blue}{}\textcolor{blue}{\emph{#1}}}

\newcommand{\D}[1]{\frac{\partial \mathcal{L}}{\partial #1 }}

\newcommand{\x}[1]{\mathbf{x}_{#1}}
\newcommand{\X}[1]{\mathbf{X}_{#1}}

\newcommand{\Nx}[1]{\hat{\mathbf{x}}_{#1}} %normalized output
\newcommand{\NX}[1]{\widehat{\mathbf{X}}_{#1}} %normalized output

\def\eg{\emph{e.g.}}

\def\ie{\emph{i.e.}}

\def\etal{\emph{et al.}}
\def\SM{Appendix}
\usepackage[pagebackref=true,breaklinks=true,colorlinks,bookmarks=false]{hyperref}

\def\cvprPaperID{2687} % *** Enter the CVPR Paper ID here
\def\confYear{CVPR 2021}

\begin{document}

%%%%%%%%% TITLE
\title{Group Whitening: Balancing Learning Efficiency and  Representational Capacity}
%\title{Group Whitening: Revisiting the Constraint Imposed by Normalization}
%\title{Group Whitening}
%\title{Group Whitening: Investigating the Constraint Imposed by Normalization}
%\title{Group Whitening: Balancing the Learning Efficiency and  Representation Ability of Normalization}

\author{Lei Huang$^{1,3}$ \quad Yi Zhou$^{2}$ \quad  Li Liu$^{3}$ \quad Fan Zhu$^{3}$ \quad  Ling Shao$^{3}$\\
	$^{1}$SKLSDE, Institute of Artificial Intelligence,  Beihang University, Beijing, China\\
	$^{2}$MOE Key Laboratory of Computer Network and Information Integration, Southeast University, China\\
	$^{3}$Inception Institute of Artificial Intelligence (IIAI), Abu Dhabi, UAE\\
	%{\tt\small $\{$huanglei, xlliu, blonster, langbo$\}$@nlsde.buaa.edu.cn, dacheng.tao@sydney.edu.au}
	%$^{2}$University of Electronic Science and Technology of China, Chengdu, China\\
%	{\tt\small $\{$lei.huang, yi.zhou,  fan.zhu, li.liu, ling.shao$\}$ @inceptioniai.org}
}

\maketitle

\thispagestyle{empty}
%%%%%%%%% ABSTRACT
\begin{abstract}
	Batch normalization (BN) is an important technique commonly incorporated into deep learning models to perform standardization within mini-batches. The merits of BN in improving a model's learning efficiency can be further amplified by applying whitening, while its drawbacks in estimating population statistics for inference can be avoided through group normalization (GN).
	This paper proposes group whitening (GW), which exploits the advantages of the whitening operation and avoids the disadvantages of normalization within mini-batches.
	In addition, we  analyze the constraints imposed on features by normalization, and show how the batch size (group number) affects the performance of batch (group) normalized networks, from the perspective of model's representational capacity.   This analysis provides theoretical guidance for  applying GW in practice.
	Finally, we apply the proposed GW to ResNet and ResNeXt architectures and conduct experiments on the ImageNet and COCO benchmarks. Results show that GW consistently improves the performance of different architectures, with absolute gains of  $1.02\%$ $\sim$ $1.49\%$ in top-1 accuracy on ImageNet and  $1.82\%$  $\sim$  $3.21\%$ in bounding box AP on COCO.
\end{abstract}
%qualitatively and quantitatively
%Finally, we apply the proposed GW to ResNet and ResNeXt architectures and conduct experiments on the ImageNet and COCO benchmarks. Results show that GW consistently improves the performance of different architectures, with absolute gains of  1.02% ~1.49% in top-1 accuracy on ImageNet and  1.82%~3.21% in bounding box AP on COCO.

% avoids BN's drawbacks in estimating population statistics with insufficient batch size while batch whitening improves BN's performance by decorrelating the standardized output.
\vspace{-0.05in}
\section{Introduction}
\label{sec_intro}
\vspace{-0.05in}

Batch normalization (BN) \cite{2015_ICML_Ioffe} represents a milestone technique in deep learning \cite{2015_CVPR_He,2015_CoRR_Szegedy,2018_ECCV_Wu}, and has been extensively used in various network architectures~ \cite{2015_CVPR_He,2015_CoRR_Szegedy,2016_CoRR_Zagoruyko,2016_CoRR_Szegedy,2016_CoRR_Huang_a}.
BN standardizes the activations within a mini-batch of data, which  improves the conditioning of optimization and accelerates training~\cite{2015_ICML_Ioffe,2016_CoRR_Ba,2018_NIPS_shibani}. Further, the stochasticity of normalization introduced along the batch dimension is believed to benefit generalization \cite{2018_ECCV_Wu,2018_ACCV_Alexander,2019_CVPR_Huang}. However,
this stochasticity also results in differences between the training distribution (using mini-batch statistics) and the test distribution (using estimated population statistics) \cite{2017_NIPS_Ioffe}, which is believed to be the main cause of BN's small-batch-size problem --- BN's error increases rapidly as the batch size becomes smaller~\cite{2018_ECCV_Wu}. To address this issue, a number of  approaches have been proposed \cite{2018_ECCV_Wu,2017_ICLR_Ren,2018_arxiv_luo,2017_NIPS_Ioffe,2018_NIPS_Wang,2019_ICCV_Singh,2020_arxiv_Cai}.
One representative method is group normalization (GN), which divides the neurons into groups and then applies the standardization operation over the neurons of each group, for each sample, independently. GN provides a flexible solution to avoid normalization along the batch dimension,
and benefits visual tasks limited to small-batch-size training~\cite{2018_ECCV_Wu,2020_ECCV_Kolesnikov}.
%Furthermore, it has been observed that BN also encounters difficulties in
%optimization during training \cite{2018_ACCV_Alexander,2019_CVPR_Huang}.
%is a more general operation of standardization that.
%. The benefits of whitened data has been demonstrated in whitening improves the conditioning of the Hessian,

As a widely used operation in data pre-processing, whitening not only standardizes but also decorrelates the data~\cite{1998_NN_Yann}, which further improves the  conditioning of the optimization problem~\cite{1998_NN_Yann,2011_NIPS_Wiesler,2015_NIPS_Desjardins,2018_CVPR_Huang}. A whitened input has also been shown to
make the gradient descent updates similar to the Newton updates for linear models~\cite{1998_NN_Yann, 2011_NIPS_Wiesler, 2018_CVPR_Huang}.
Motivated by this, Huang~\etal~\cite{2018_CVPR_Huang} proposed batch whitening (BW) for deep models, which performs whitening on the activations of each layer within a mini-batch. BW has been shown to  achieve better optimization efficiency and generalization than BN~\cite{2018_CVPR_Huang,2019_CVPR_Huang,2019_ICCV_Pan}.  However, BW further amplifies the disadvantage of BN in estimating the population statistics, where the number of parameters to be estimated with BW is quadratic to the number of neurons/channels. Thus, BW requires a sufficiently large batch size to work well.
%heavily limits the usage of BW.
%benefits of whitening operaimmake BW require sufficient large batch-size to achieve good performance.
% We experimentally find that such an estimation not only has problem in small batch size scenarios (e.g. Object dection), but also suffer the problems in moderate batch size.

%\paragraph{Contributions}
To exploit whitening's advantage in optimization, while avoiding its disadvantage in normalization along the batch dimension, this paper proposes group whitening (GW).
%For each sample, GW divides the neurons into groups for standardization over the neurons of each group (like GN), and then meanwhile decorrelates the groups.
 GW divides the neurons of a sample into groups for standardization over the neurons in each group, and then decorrelates the groups.
Unlike BW, GW has stable performance for a wide range of batch sizes, like GN, and thus can be applied to a variety of tasks.
GW further improves the conditioning of optimization of GN with its whitening operation.
%, which is experimentally validated in this paper

One important hyperparameter of GW is the group number.
We observe that GW/GN has a significantly degenerated training performance when the group number is large, which is similar to the small-batch-size problem of BW/BN.
We attribute this to the constraints on the output imposed by the normalization operation, which affect the model's representational capacity.
As such, this paper defines the \textbf{constraint number} of normalization (as will be discussed in Section~\ref{sec_theory}) to quantitatively measure the magnitude of the constraints provided by normalization methods.
With the support of the constraint number, we analyze how the batch size (group number) affects the model's representational capacity for batch (group) normalized networks.
Our analysis also presents a new viewpoint for understanding the small-batch-size problem of BN.
%, by constraining the representation of feature in the internal layers

We apply the proposed GW to two representative deep network architectures (ResNet~\cite{2015_CVPR_He} and ResNeXt~\cite{2017_CVPR_Xie}) for ImageNet classification~\cite{2015_IJCV_ImageNet} and  COCO object detection and instance segmentation~\cite{2014_ECCV_COCO}. GW consistently improves the performance for both architectures, with absolute gains of $1.02\%$ \textasciitilde $1.49\%$ in top-1 accuracy for ImageNet and  $1.82\%$ \textasciitilde $3.21\%$ in bounding box AP for COCO.

\vspace{-0.05in}
\section{Preliminaries}
\label{sec_pre}
%\vspace{-0.2in}

For simplicity, we first consider the $d$-dimensional input vector $\mathbf{x}$, which will be generalized to a convolutional input in the subsequent section.
Let $\mathbf{X} \in \mathbb{R}^{d \times m}$ be a data matrix denoting the mini-batch input  of size $m$ in a given layer.
\vspace{-0.12in}
\paragraph{Standardization.}
During training, batch normalization (BN) \cite{2015_ICML_Ioffe} standardizes the layer input  within a mini-batch, for each neuron, as\footnote{BN and other normalization methods discussed in this paper all use extra learnable scale and shift parameters~\cite{2015_ICML_Ioffe}.  We omit this for simplicity.}:
{\setlength\abovedisplayskip{3pt}
	\setlength\belowdisplayskip{3pt}
	\begin{equation}
	\label{eqn:BN}
	\widehat{\mathbf{X}}=\phi_{BN}(\mathbf{X})
	= \Lambda_{d}^{-\frac{1}{2}} (\mathbf{X} - \mathbf{\mu}_d  \mathbf{1}^T).
	\end{equation}
}
\hspace{-0.04in}Here, $\mathbf{\mu}_d = \frac{1}{m} \mathbf{X} \mathbf{1}$ and  $\Lambda_{d}=\mbox{diag}(\sigma_1^2, \ldots,\sigma_d^2) + \epsilon \mathbf{I}$, where $\sigma_i^2$ is the variance over mini-batches for the \emph{i}-th neuron, $\mathbf{1}$ is a column vector of all ones, and $\epsilon>0$ is a small number to prevent numerical instability. During inference, the population statistics $\{\widehat{\Lambda }_{d}^{-\frac{1}{2}}, \hat{\mathbf{\mu}}_d  \}$ are required for deterministic inference, and they are usually calculated by running average over the training iterations, as follows:
\begin{small}
	\setlength\abovedisplayskip{3pt}
	\setlength\belowdisplayskip{3pt}
	\begin{equation}
	\label{eqn:running_average}
	\begin{cases}
	\hat{\mathbf{\mu}}_d  = (1-\lambda)  \hat{\mathbf{\mu}}_d   + \lambda  \mathbf{\mu}_d, \\
	\widehat{\Lambda }_{d}^{-\frac{1}{2}} = (1-\lambda) \widehat{\Lambda }_{d}^{-\frac{1}{2}} + \lambda \Lambda_{d}^{-\frac{1}{2}}.
	\end{cases}
	\end{equation}
\end{small}
\hspace{-0.04in}Such an estimation process can limit the usage of BN  in recurrent neural networks~\cite{2016_ICASSP_Laurent,2016_CoRR_Cooijmans}, or harm the performance for small-batch-size training~\cite{2017_NIPS_Ioffe,2018_ECCV_Wu}.

%\begin{small}
%	\setlength\abovedisplayskip{0.03in}
%	\setlength\belowdisplayskip{0.03in}
%	\begin{eqnarray}
%	\label{eqn:running_average}
%	%	\hat{\mu} &=& (1-\lambda) ~\hat{\mu} + \lambda ~\mathbf{\mu} \nonumber \\
%	\{\widehat{\Lambda }_{d}^{-\frac{1}{2}}, \hat{\mathbf{\mu}}_d  \} & = &(1-\lambda) \{\widehat{\Lambda }_{d}^{-\frac{1}{2}}, \hat{\mathbf{\mu}}_d  \} + \lambda \{\Lambda_{d}^{-\frac{1}{2}}, \mathbf{\mu}_d  \}.
%	\end{eqnarray}
%\end{small}
To avoid the estimation of population statistics shown in Eqn.~\ref{eqn:running_average}, Ba \etal ~proposed layer normalization (LN)~\cite{2016_CoRR_Ba} to standardize the layer input within the neurons for each training sample, as:
{\setlength\abovedisplayskip{3pt}
	\setlength\belowdisplayskip{3pt}
	\begin{equation}
	\label{eqn:LN}
	\widehat{\mathbf{X}}=\phi_{LN}(\mathbf{X})
	=  (\mathbf{X} -   \mathbf{1}  \mathbf{\mu}_m^T) \Lambda_{m}^{-\frac{1}{2}}.
	\end{equation}
}
\hspace{-0.04in}Here, $\mathbf{\mu}_m = \frac{1}{d}  \mathbf{X}^T   \mathbf{1} $ and  $\Lambda_{m}=\mbox{diag}(\sigma_1^2, \ldots,\sigma_m^2) + \epsilon \mathbf{I}$, where $\sigma_i^2$ is the variance over the neurons for the \emph{i}-th sample. LN has the same formulation during training and inference, and is extensively used in natural language processing tasks~\cite{2017_NIPS_Vaswani,2018_ICLR_Yu,2019_NIPS_Xu}.
%since it does not normalize within a mini-batch.

Group normalization (GN)~\cite{2018_ECCV_Wu} further generalizes LN, dividing the neurons into groups and performing the standardization within the neurons of each group independently, for each sample. Specifically, defining the group division operation as $\Pi: \mathbb{R}^{d \times m} \mapsto \mathbb{R}^{c \times gm}$, where $g$ is the group number and $d=g c$, GN can be represented as follows:
{\setlength\abovedisplayskip{5pt}
	\setlength\belowdisplayskip{5pt}
	\begin{equation}
	\label{eqn:GN}
	\widehat{\mathbf{X}}=\phi_{GN}(\mathbf{X}; g)
	= \Pi^{-1}(\phi_{LN} (\Pi(\mathbf{X}))),
	\end{equation}
}
\hspace{-0.04in}where $\Pi^{-1}: \mathbb{R}^{c \times gm} \mapsto \mathbb{R}^{d \times m} $ is the inverse operation of $\Pi$. It is clear from Eqn.~\ref{eqn:GN} that LN is a special case of GN with $g=1$. By changing the group number $g$, GN is more flexible than LN, enabling it to achieve good performance on visual tasks limited to small-batch-size training (\eg, object detection and segmentation~\cite{2018_ECCV_Wu}).

\begin{figure}[t]
	\centering
	\vspace{-0.08in}
\hspace{-0.25in}	\subfigure[]{
		\begin{minipage}[c]{.43\linewidth}
			\centering
			\includegraphics[width=4.0cm]{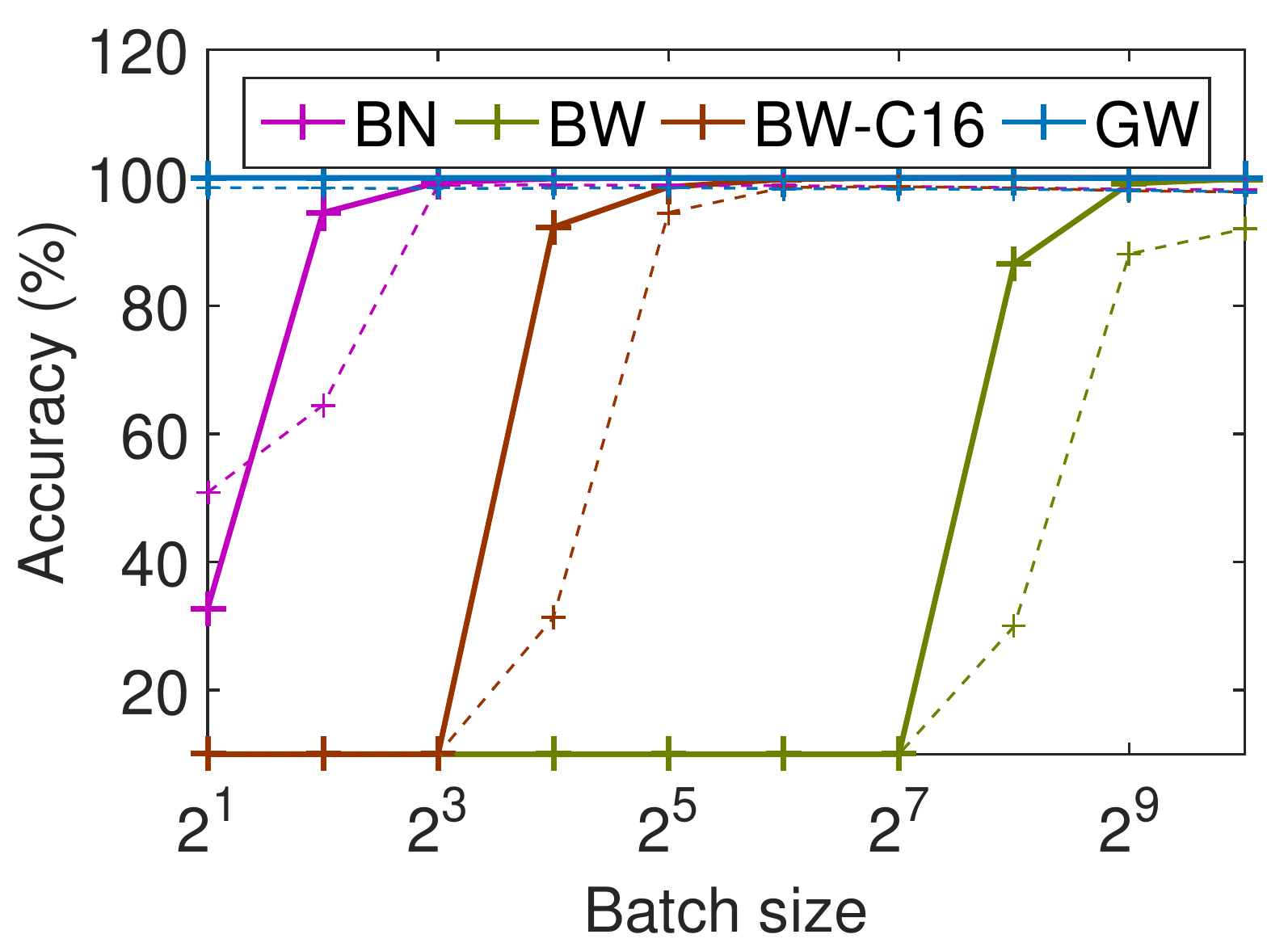}
		\end{minipage}
	}
\hspace{0.15in}		\subfigure[]{
		\begin{minipage}[c]{.43\linewidth}
			\centering
			\includegraphics[width=4.0cm]{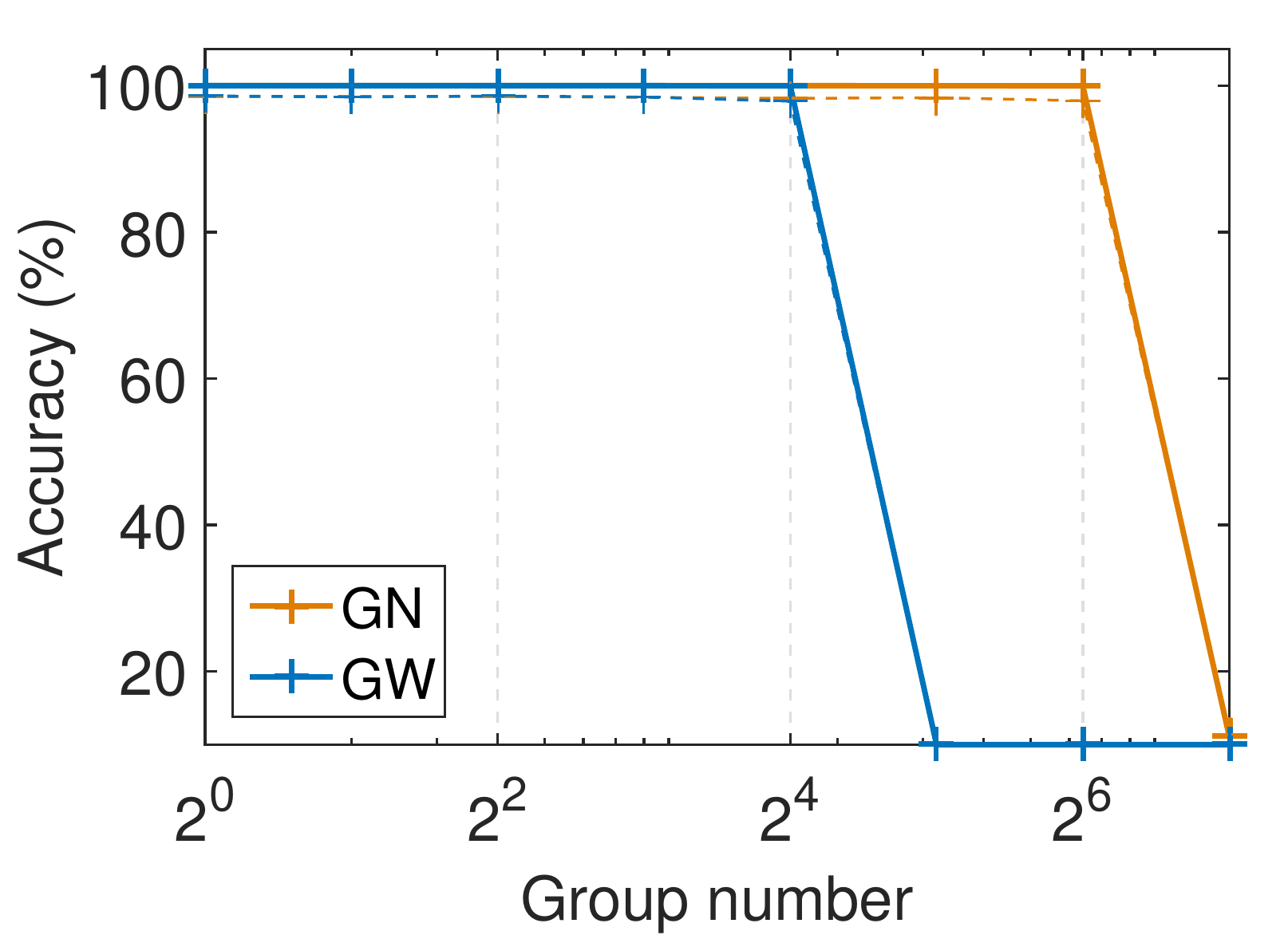}
		\end{minipage}
	}
	%\vspace{-0.08in}
	\caption{Effects of batch size (group number) for batch (group) normalized networks. We train a four-layer multilayer perceptron (MLP) with 256 neurons in each layer, for MNIST classification. We  evaluate the training (thick `plus' with solid line) and validation (thin `plus' with dashed line) accuracies at the end of 50 training epochs. Note that `BW-C16' indicates  group-based BW with 16 neurons in each group.  We vary the batch size and group number in (a) and (b), respectively.  These results are obtained using a learning rate of 0.1, but we also obtain similar observations for other learning rates.  Please see the \TODO{\SM}~\ref{sup:batchSize} for details.}
	\label{fig:MLP_BatchSize}
	\vspace{-0.17in}
\end{figure}

\vspace{-0.1in}
\paragraph{Whitening.}
%BN, LN and GN all normalize the activations with standardization operation, which is a special case of the more general whitening operation that further decorrelating the standardized output.
To exploit the advantage of whitening over standardization in improving the conditioning of optimization, Huang~\etal~ proposed
decorrelated BN ~\cite{2018_CVPR_Huang}, which performs zero-phase component analysis (ZCA) whitening  to normalize the layer input within a mini-batch, as:
\begin{small}
	{\setlength\abovedisplayskip{5pt}
		\setlength\belowdisplayskip{5pt}
		\begin{equation}
		\label{eqn:whiten}
		\phi^{W}_{ZCA}(\mathbf{X})= \Sigma^{-\frac{1}{2}}_{d}(\mathbf{X} - \mathbf{\mu}_d  \mathbf{1}^T)
		= \mathbf{D}\Lambda^{-\frac{1}{2}} \mathbf{D}^T (\mathbf{X} - \mathbf{\mu}_d  \mathbf{1}^T),
		\end{equation}
	}
\end{small}
\hspace{-0.09in}where  $\Lambda=\mbox{diag}(\tilde{\sigma}_1, \ldots,\tilde{\sigma}_d)$ and $\mathbf{D}=[\mathbf{d}_1, ...,
\mathbf{d}_d]$ are the eigenvalues and associated eigenvectors of $\Sigma$, \ie~$\Sigma = \mathbf{D}
\Lambda \mathbf{D}^T$, and $\Sigma = \frac{1}{m} (\mathbf{X} - \mathbf{\mu}_d
\mathbf{1}^T) (\mathbf{X} - \mathbf{\mu}_d  \mathbf{1}^T)^T +
\epsilon \mathbf{I}$ is the covariance matrix of the centered input.
One crucial problem in Eqn.~\ref{eqn:whiten} is the eigen-decomposition, which is computationally expensive on a GPU and numerically instable. To address this issue, iterative normalization (`ItN')~\cite{2019_CVPR_Huang} was proposed to approximate the ZCA whitening matrix $\Sigma^{-\frac{1}{2}}_{d}$ using Newton's iteration~\cite{2005_NumerialAlg}:
\begin{small}
	{\setlength\abovedisplayskip{3pt}
		\setlength\belowdisplayskip{3pt}
		\begin{equation}
		\label{eqn:IterNorm}
		\phi^{W}_{ItN}(\mathbf{X})= \Sigma^{-\frac{1}{2}}_{d}(\mathbf{X} - \mathbf{\mu}_d  \mathbf{1}^T)
		= \frac{\mathbf{P}_T}{\sqrt{tr(\Sigma_d)}} (\mathbf{X} - \mathbf{\mu}_d  \mathbf{1}^T),
		\end{equation}
	}
\end{small}
\hspace{-0.05in}where $tr(\Sigma_d)$ indicates the trace of $\Sigma_d$ and $\mathbf{P}_T$ is calculated iteratively as:
\begin{small}
	{\setlength\abovedisplayskip{3pt}
		\setlength\belowdisplayskip{3pt}
		\begin{equation}
		\label{eqn:Iteration}
		\begin{cases}
		\mathbf{P}_0=\mathbf{I} \\
		\mathbf{P}_{k}=\frac{1}{2} (3 \mathbf{P}_{k-1} - \mathbf{P}_{k-1}^{3} \Sigma^{N}_{d}), ~~ k=1,2,...,T.
		\end{cases}
		\end{equation}
	}
\end{small}
\hspace{-0.08in}Here, $\Sigma^{N}_{d} = \frac{\Sigma_d} {tr(\Sigma_d)}$. Other BW methods also exist for calculating the whitening matrix~\cite{2018_CVPR_Huang, 2019_ICLR_Siarohin}; please refer to~\cite{2018_AS_Kessy,2020_CVPR_Huang} for more details.

%The whitening operation has better advantage in improving the conditioning of optimization problem than standardization \TODO{cite},\TODO{SWitening}.
It is necessary for BW to estimate the population statistics of the whitening matrix $\widehat{\Sigma}^{-\frac{1}{2}}_{d}$ during inference, like BN. However, the number of independent parameters in $\widehat{\Sigma}^{-\frac{1}{2}}_{d}$ of BW is $\frac{d(d+1)}{2}$, while $\widehat{\Lambda }_{d}^{-\frac{1}{2}}$ of BN is $d$. This amplifies the difficulty in estimation and requires a sufficiently large batch size for BW to work well (Figure~\ref{fig:MLP_BatchSize}). Although group-based BW~\cite{2018_CVPR_Huang} --- where neurons are divided into groups and BW is performed within each one --- can relieve this issue, it  is still sensitive to the batch size (Figure~\ref{fig:MLP_BatchSize}) due to its inherent drawback of normalizing along the batch dimension.

\vspace{-0.05in}
\section{Group Whitening}
\vspace{-0.05in}
\label{sec_method}
%For each sample $\mathbf{x} \in \mathbb{R}^{d}$,  Group Whitening (GW)  first divides the neurons into groups: $\mathbf{X}_G= \Pi(\mathbf{x}; g) \in \mathbb{R}^{g \times c}$ where $\Pi: \mathbb{R}^{d} \mapsto \mathbb{R}^{g \times c}$, and then perform whiten operation over the reshaped data $\mathbf{x}_G$ as:
We propose group whitening (GW).
%For each sample, GW divides the neurons into groups and perform standardization over the neurons of each group (like GN), and further decorrelates the groups. Specifically,
Given a sample $\mathbf{x} \in \mathbb{R}^{d}$,  GW performs normalization as:
{\setlength\abovedisplayskip{3pt}
	\setlength\belowdisplayskip{3pt}
	\begin{align}
	\label{eqn:GW-1}
&	Group ~ division: \mathbf{X}_G= \Pi(\mathbf{x}; g) \in \mathbb{R}^{g \times c}, \\
	\label{eqn:GW}
&	Whitening:\widehat{\mathbf{X}}_G = \Sigma^{-\frac{1}{2}}_{g}(\mathbf{X}_G - \mathbf{\mu}_g  \mathbf{1}^T), \\
%&	Whitening:\widehat{\mathbf{X}}_G = \phi^{W}(\mathbf{X}_G) = \Sigma^{-\frac{1}{2}}_{g}(\mathbf{X}_G - \mathbf{\mu}_g  \mathbf{1}^T), \\
	\label{eqn:GW-2}
&	Inverse ~ group~ division: \hat{\mathbf{x}}= \Pi^{-1}(\widehat{\mathbf{X}}_G) \in \mathbb{R}^{d},
	\end{align}
}
\hspace{-0.05in}where $\Pi: \mathbb{R}^{d} \mapsto \mathbb{R}^{g \times c}$ and its inverse transform $\Pi^{-1}: \mathbb{R}^{g \times c} \mapsto \mathbb{R}^{d}$.
%The full algorithom are shown in \TODO{algorithm **},
We can use different whitening operations~\cite{2018_CVPR_Huang,2019_ICLR_Siarohin,2018_AS_Kessy} in Eqn.~\ref{eqn:GW}. Here, we use ZCA whitening (Eqn.~\ref{eqn:whiten}) and its efficient approximation `ItN' (Eqn~\ref{eqn:IterNorm}), since they work well on discriminative tasks~\cite{2018_CVPR_Huang,2019_CVPR_Huang,2019_ICCV_Pan,2020_arxiv_Shao,2020_ICLR_Ye}. We provide the full algorithms (forward and backward passes) and PyTorch~\cite{2017_NIPS_pyTorch} implementations in the \TODO{\SM}~\ref{sup:algorithm}.

\begin{figure}[t]
	\centering
\vspace{-0.10in}
\hspace{-0.25in}	\subfigure[]{
		\begin{minipage}[c]{.43\linewidth}
			\centering
			\includegraphics[width=4.2cm]{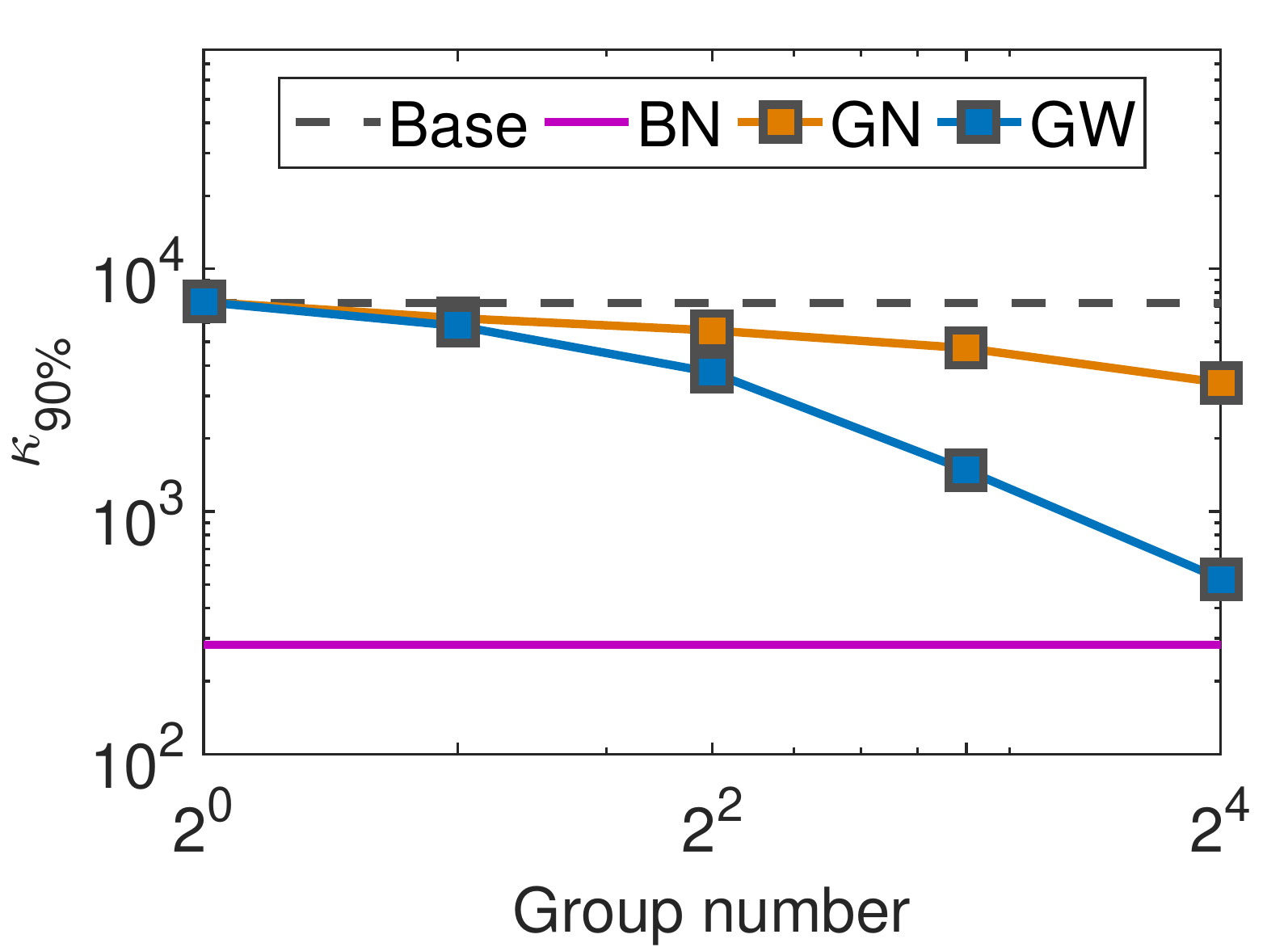}
		\end{minipage}
	}
\hspace{0.15in}		\subfigure[]{
		\begin{minipage}[c]{.43\linewidth}
			\centering
			\includegraphics[width=4.2cm]{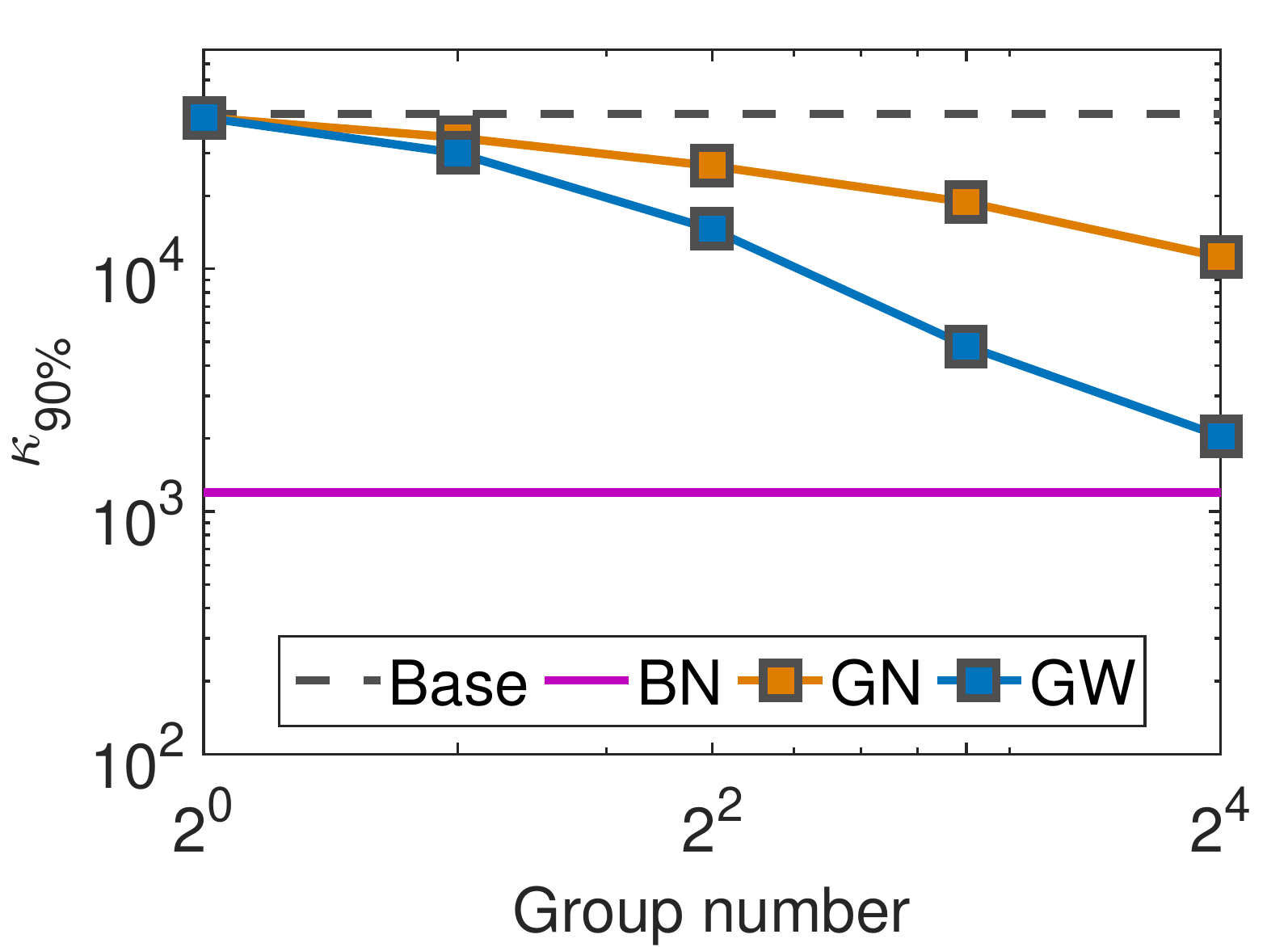}
		\end{minipage}
	}	
	\vspace{-0.05in}
	\caption{Conditioning analysis on the normalized output. We simulate the activations $\mathbf{X}=f(\mathbf{X}_0) \in \mathbb{R}^{256 \times 1024}$ using a network $f(\cdot)$, where $\mathbf{X}_0$ is sampled from a Gaussian distribution.  We evaluate the more general condition number with respect to the percentage: $\kappa_{p}=\frac{\lambda_{max}}{\lambda_{p}}$, where $\lambda_{p}$ is the $pd$-th eigenvalue (in descending order) and $d$ is the total number of eigenvalues. We show the $\kappa_{90\%}$ of the covariance matrix of $\widehat{\mathbf{X}}$ normalized by GN/GW, while varying the group number. `Base' and `BN' indicate the condition number for $\mathbf{X}$ and the batch normalized output, respectively. We use a one-layer and two-layer MLP as $f(\cdot)$, in (a) and (b), respectively. Please refer to \TODO{\SM}~\ref{sup:conditioning} for more results.  }
	\label{fig:conditioning}
	\vspace{-0.16in}
\end{figure}

GW avoids normalization along the batch dimension, and it works stably across a wide range of batch sizes (Figure~\ref{fig:MLP_BatchSize}).
GW also ensures that the normalized activation for each sample has the properties: $\widehat{\mathbf{X}}_G  \mathbf{1} = \mathbf{0}$ and $\frac{1}{c} \widehat{\mathbf{X}}_G \widehat{\mathbf{X}}_G^T = \mathbf{I}$, which should improve the conditioning, like BW, and benefit training.
%\TODO{Show the results of improved conditioning:}
 %We experimentally show that whitening along the group
%A better conditioned input will have better conditioning of the optimization.
 We conduct several experiments to validate this,  and the results in Figure~\ref{fig:conditioning} show that the group whitened  output (by GW) has significantly better conditioning than the group standardized one (by GN), which is similar to normalization along the batch dimension~\cite{2018_CVPR_Huang}. Note that the condition number of BW is $1$. We also find that GN/GW has better conditioning with increasing group number. Besides, we find that BN has better conditioning than GN/GW, which suggests that normalizing along the batch dimension is better for decorrelating the data than normalizing along the channel dimension.

%\paragraph{Implementation}
%The core in here of our method is to calcluate $\Sigma_g^{-\frac{1}{2}}$. We can use the ZCA whitneing method \TODO{ref}, as shown in Eqn \TODO{ref}, to implement it. We also can use its more efficient approximation IterNorm. We provide the algorithm and the pytroch code in \TODO{\SM}
\vspace{-0.1in}
\paragraph{Convolutional layer.}
For the convolutional input $\mathbf{X} \in \mathbb{R}^{d \times m \times H \times W}$, where H and W are the height and width of the feature maps, BN and BW  consider each spatial position in a feature map as a sample~\cite{2015_ICML_Ioffe} and normalize over the unrolled input $\mathbf{X} \in \mathbb{R}^{d \times m  H  W}$. In contrast, LN and GN  view each spatial position in a feature map as a neuron~\cite{2018_ECCV_Wu} and normalize over the unrolled input $\mathbf{X} \in \mathbb{R}^{d  H  W \times m }$. Following GN, GW also views each spatial position as a neuron, \ie, GW operations (Eqns.~\ref{eqn:GW-1},~\ref{eqn:GW} and~\ref{eqn:GW-2}) are performed for each sample with unrolled input $\mathbf{x} \in \mathbb{R}^{dHW}$.
%There are some alternation in applying our GW on Convoluaitonal layers, like the Positional normalization, or division normalization, over standardization operation for specific task/domain. We leave it to our further research direction.
\vspace{-0.1in}
\paragraph{Computational complexity.}
For a convolutional mini-batch input $\mathbf{X} \in \mathbb{R}^{d \times m \times H \times W}$, GW using ZCA whitening (Eqn.~\ref{eqn:whiten}) costs  $2mHWdg + m O(g^3)$. Using the more efficient `ItN' operation (Eqn.~\ref{eqn:IterNorm}), GW costs $2mHWdg+ mTg^3$, where T is the iteration number. The $3 \times 3$ convolution with the same input and output feature maps costs $9mHWd^2$. The relative cost of GW for a $3 \times 3$ convolution is $\frac{2g}{9d} + \frac{Tg^3}{9HWd^2}$.
%We can see the introduced computational cost of GW is not
%$2/(9g) + Tg^3/9HWd^2$
%compared to the group-based whitening with: $2mHWd^2/g + O((d/g)^3)g$. We can find that with small batch size training, Group whitneing is more efficient, compared to group-based whitening.

\vspace{-0.1in}
\paragraph{Difference from group-based BW.}
Our method is significantly different from the group-based BW~\cite{2018_CVPR_Huang}, in which the whitening operation is also applied within mini-batch data. Specifically, group-based BW has difficulty in estimating the population statistics, as discussed in Section~\ref{sec_pre}. Note that group-based BW is reduced to BN if the channel number in each group $c=1$, while GW is reduced to GN if the group number $g=1$.
%, they still a whitening over mini-batch methods, which has the disadvantage as shown in \TODO{Figure}. Note that in Group-based DBN, NC=1, the methods is reduced to the original batch normalization. for our method, if g=1, our methods is reduced to the group normalization.

%\begin{figure}[t]
%	\centering
%	\vspace{-0.15in}
%	\hspace{-0.3in}	\subfigure[Train performance]{
%		\begin{minipage}[c]{.44\linewidth}
%			\centering
%			\includegraphics[width=5.8cm]{./figures/MLP/exp1_batchSize_Test.pdf}
%		\end{minipage}
%	}
%	\hspace{0.1in}	\subfigure[Test performance]{
%		\begin{minipage}[c]{.44\linewidth}
%			\centering
%			\includegraphics[width=5.8cm]{./figures/MLP/exp1_batchSize_Train.pdf}
%		\end{minipage}
%	}
%	\vspace{-0.06in}
%	\caption{Ablation Study on MNISAT. Here we vary the batch size during training. We show how the normalization over Batch dimesion is affected by the batch size while not affected by the Group-based normalization. Here we use the whiteing matrix as a estimator for the BW during trianing following \TODO{cite}. We also tried the more accreate estaimitaotn methods usrinty covariance matrix as introduced in \TODO{cite}. We find 2 can imporve the perfomance but it still has significnatly dengearte peorfiamnce compared to group-based method }
%	\label{fig:SND_analysis}
%	\vspace{-0.2in}
%\end{figure}

\begin{figure*}[t]
	\centering
		\vspace{-0.22in}
			\subfigure[c=2]{
		\begin{minipage}[c]{.17\linewidth}
			\centering
			\includegraphics[width=3.0cm]{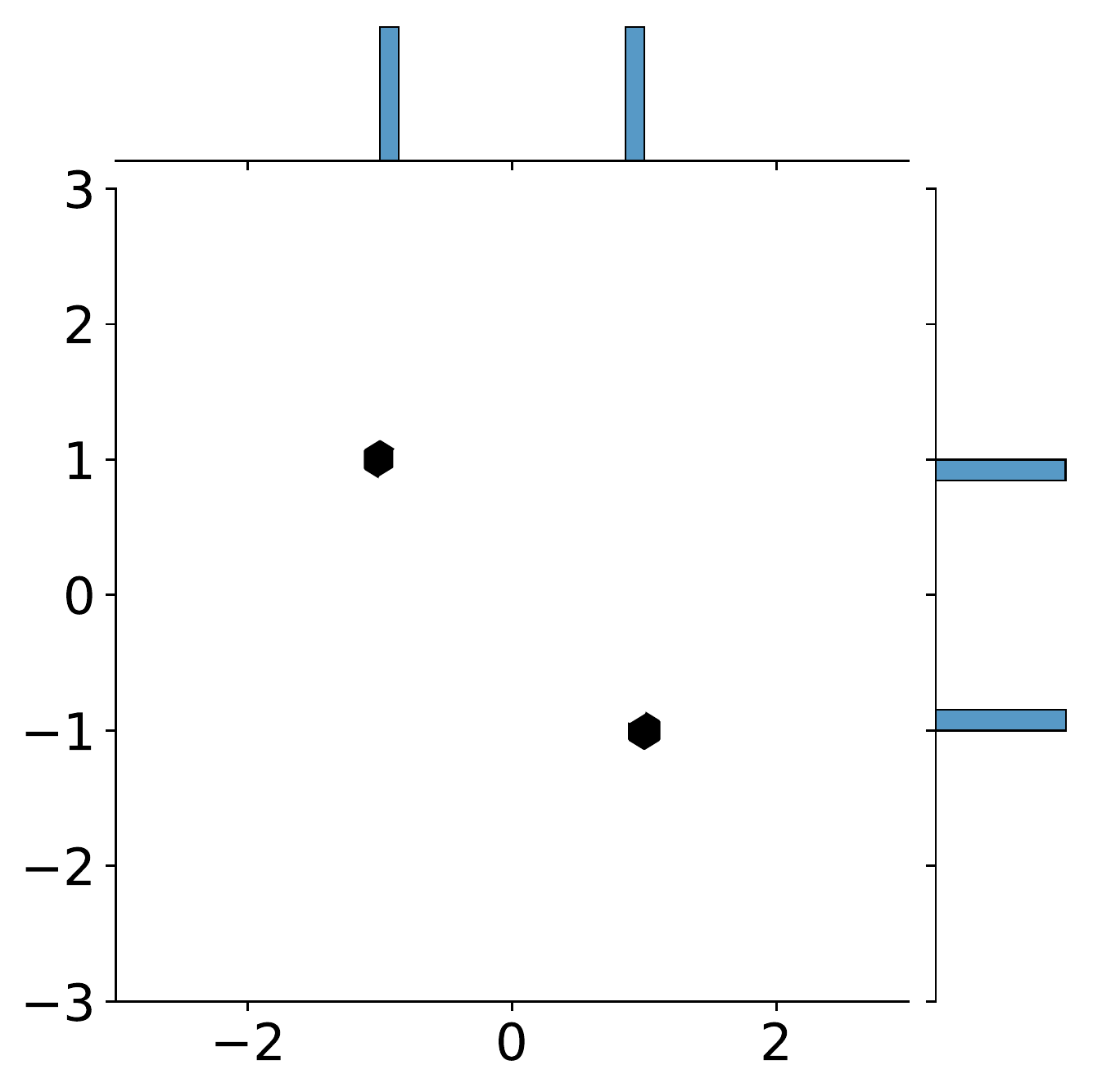}
		\end{minipage}
	}
	\subfigure[c=3]{
		\begin{minipage}[c]{.17\linewidth}
			\centering
			\includegraphics[width=3.0cm]{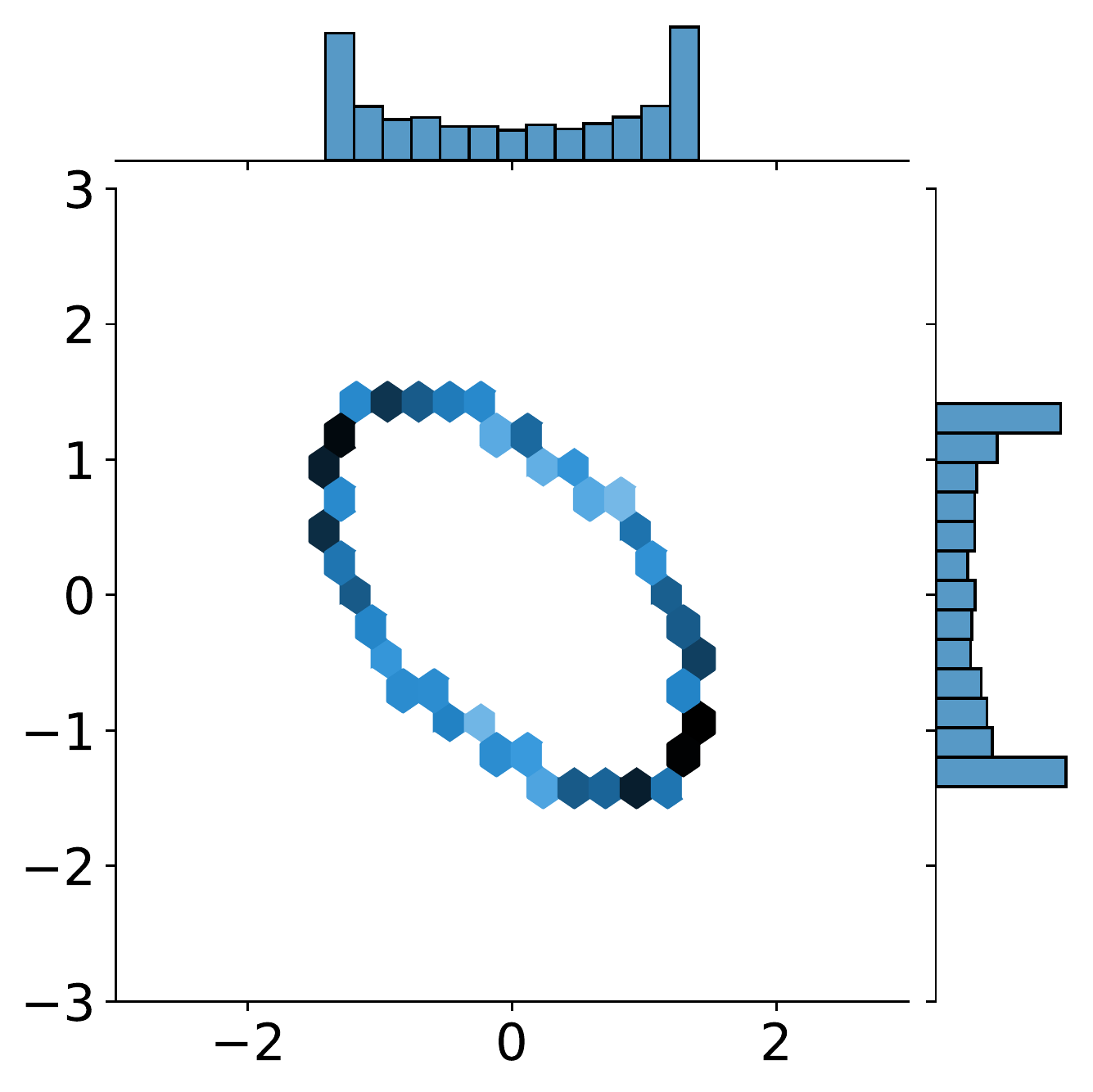}
		\end{minipage}
	}
	\subfigure[c=4]{
		\begin{minipage}[c]{.17\linewidth}
			\centering
			\includegraphics[width=3.0cm]{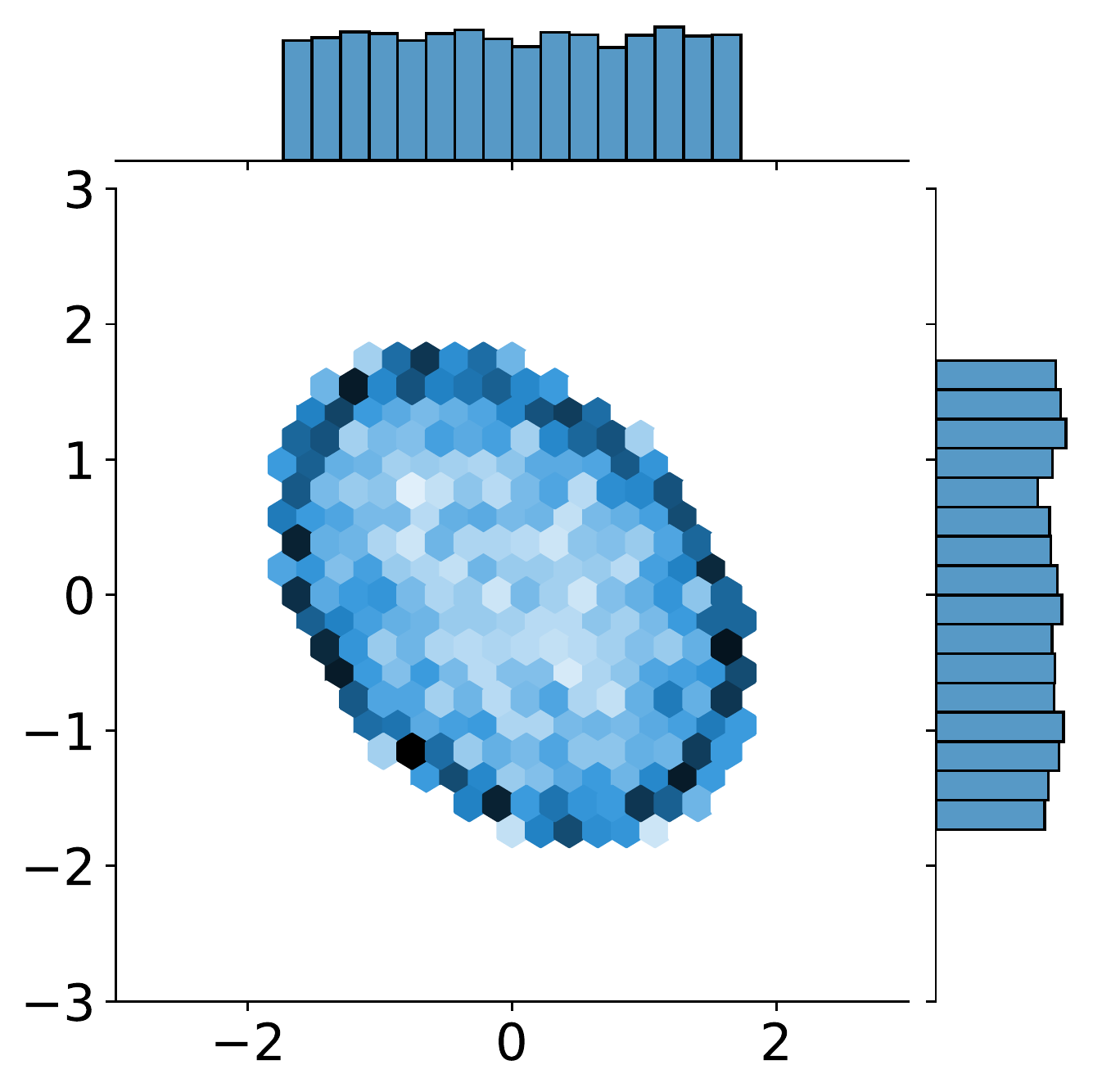}
		\end{minipage}
	}
	\subfigure[c=8]{
		\begin{minipage}[c]{.17\linewidth}
			\centering
			\includegraphics[width=3.0cm]{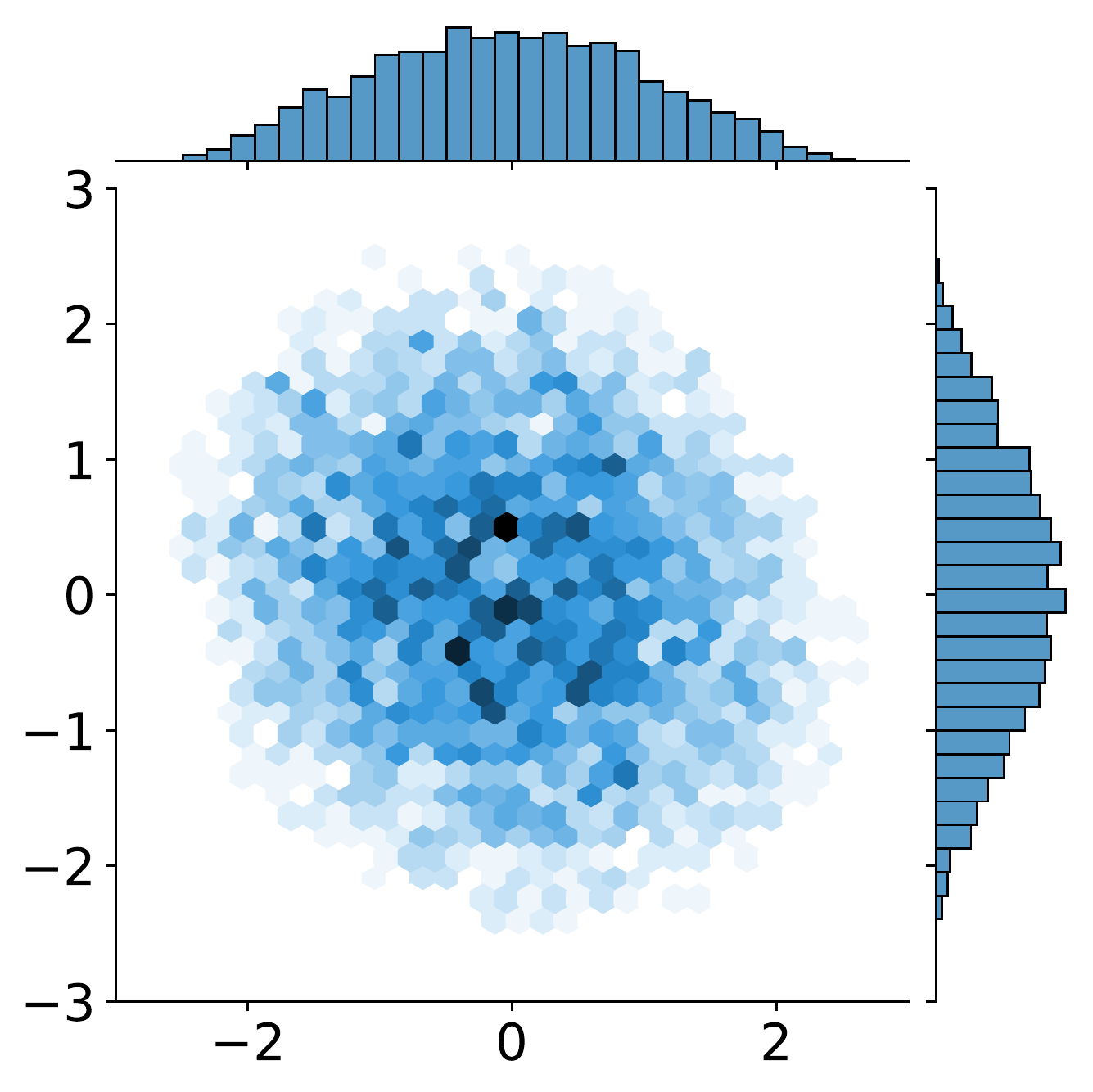}
		\end{minipage}
	}
	\subfigure[c=16]{
		\begin{minipage}[c]{.17\linewidth}
			\centering
			\includegraphics[width=3.0cm]{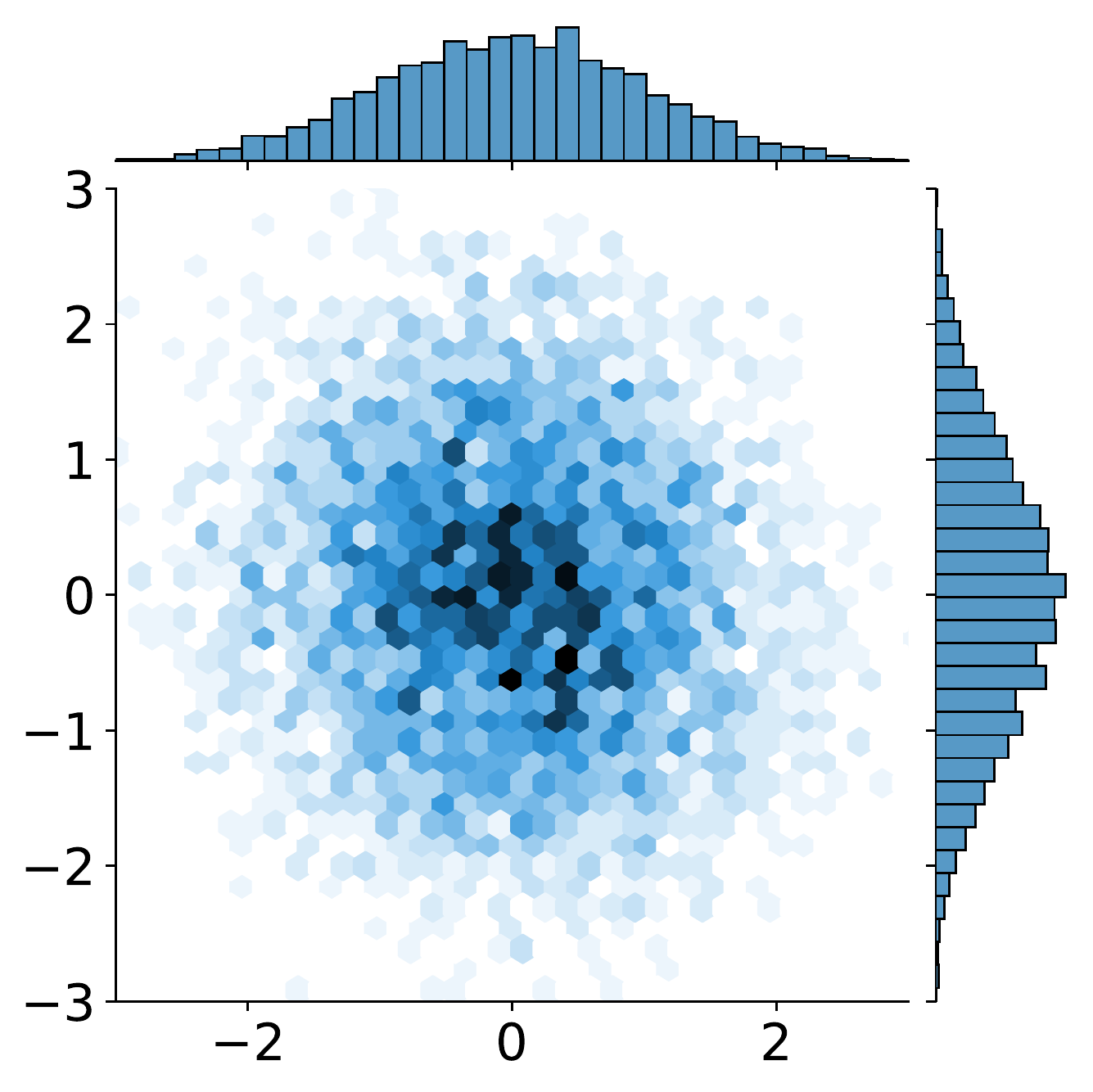}
		\end{minipage}
	}
\vspace{-0.05in}
		\subfigure[m=2]{
		\begin{minipage}[c]{.17\linewidth}
			\centering
			\includegraphics[width=3.0cm]{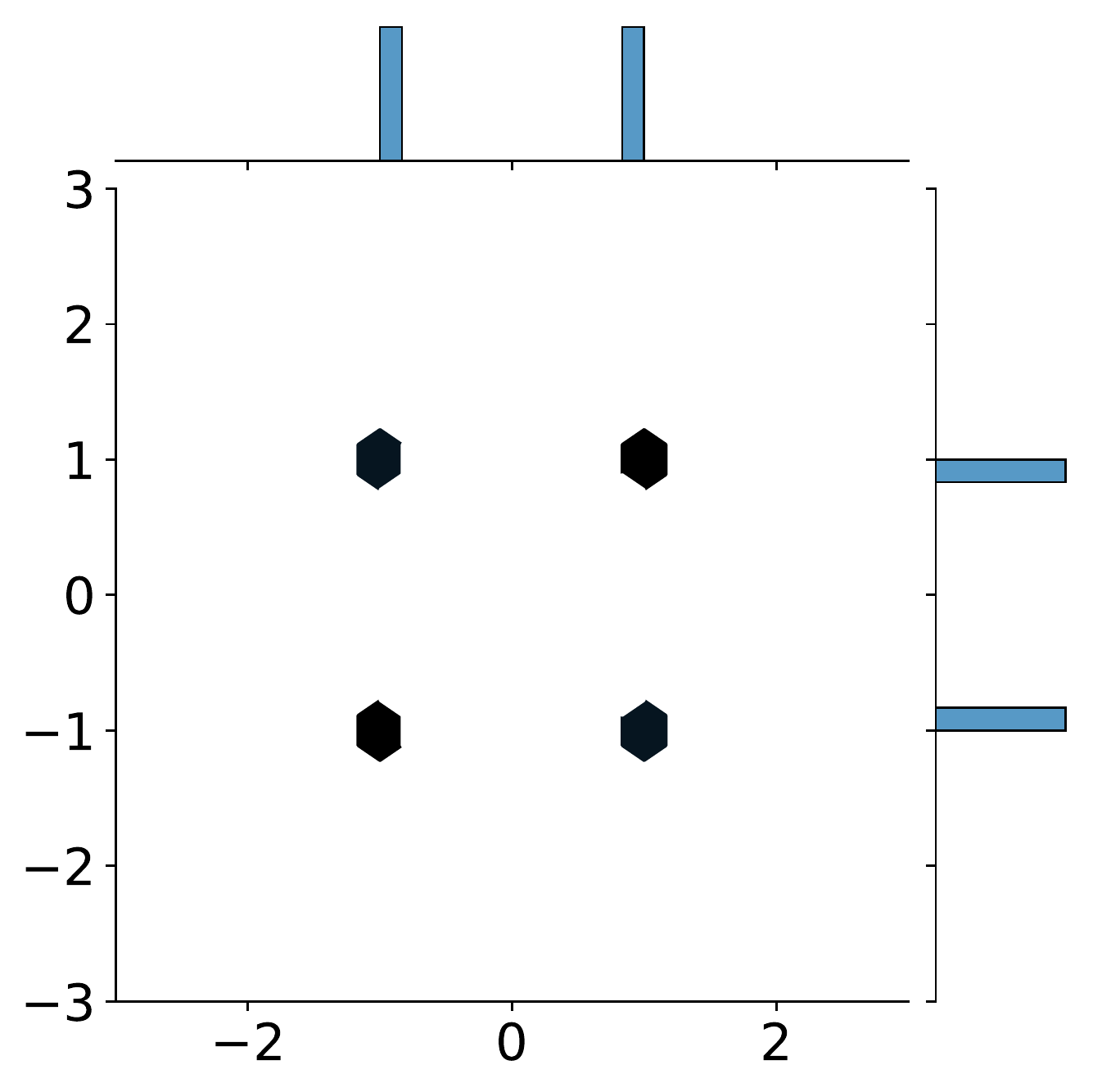}
		\end{minipage}
	}
	\subfigure[m=3]{
		\begin{minipage}[c]{.17\linewidth}
			\centering
			\includegraphics[width=3.0cm]{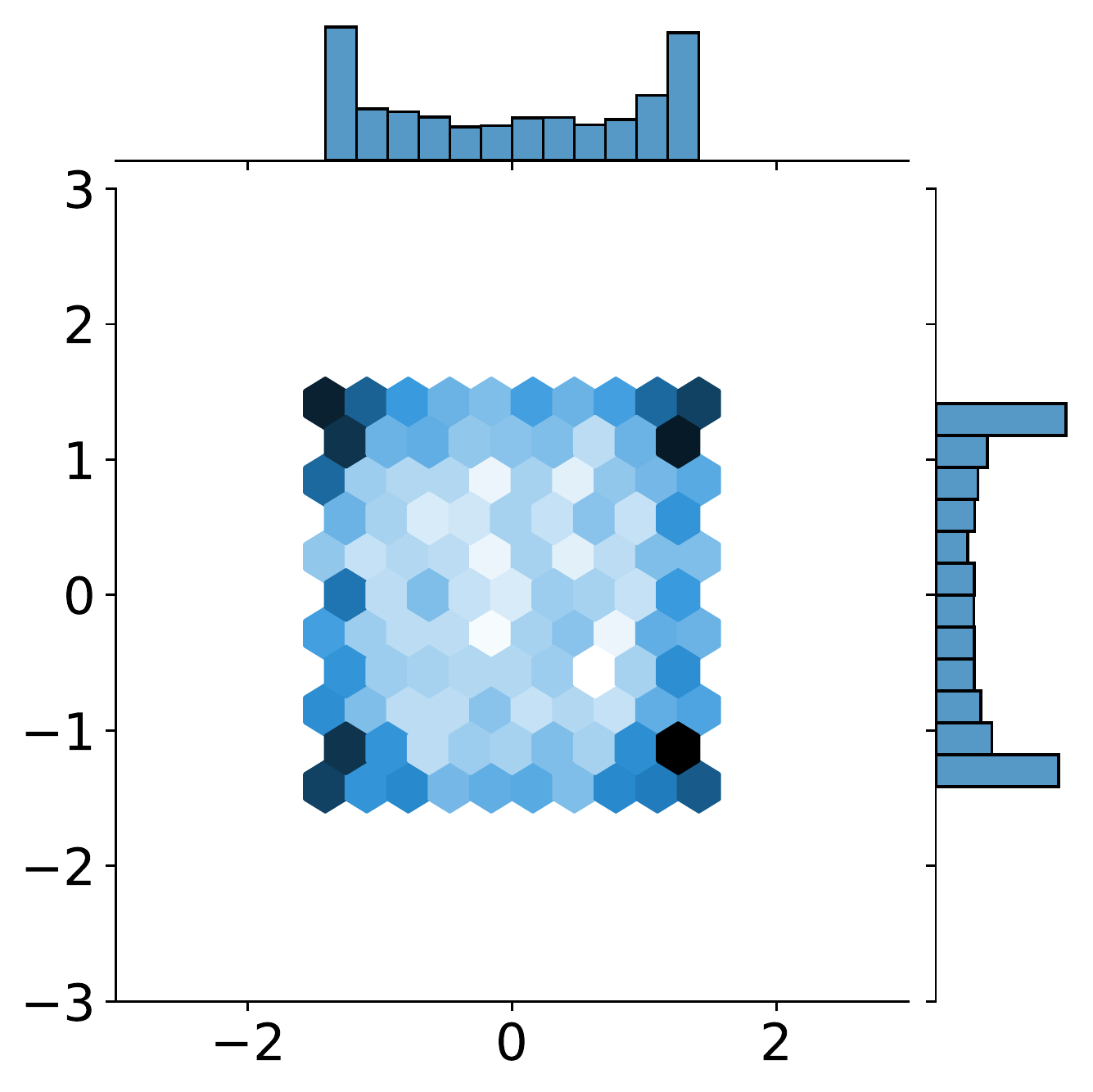}
		\end{minipage}
	}
	\subfigure[m=4]{
		\begin{minipage}[c]{.17\linewidth}
			\centering
			\includegraphics[width=3.0cm]{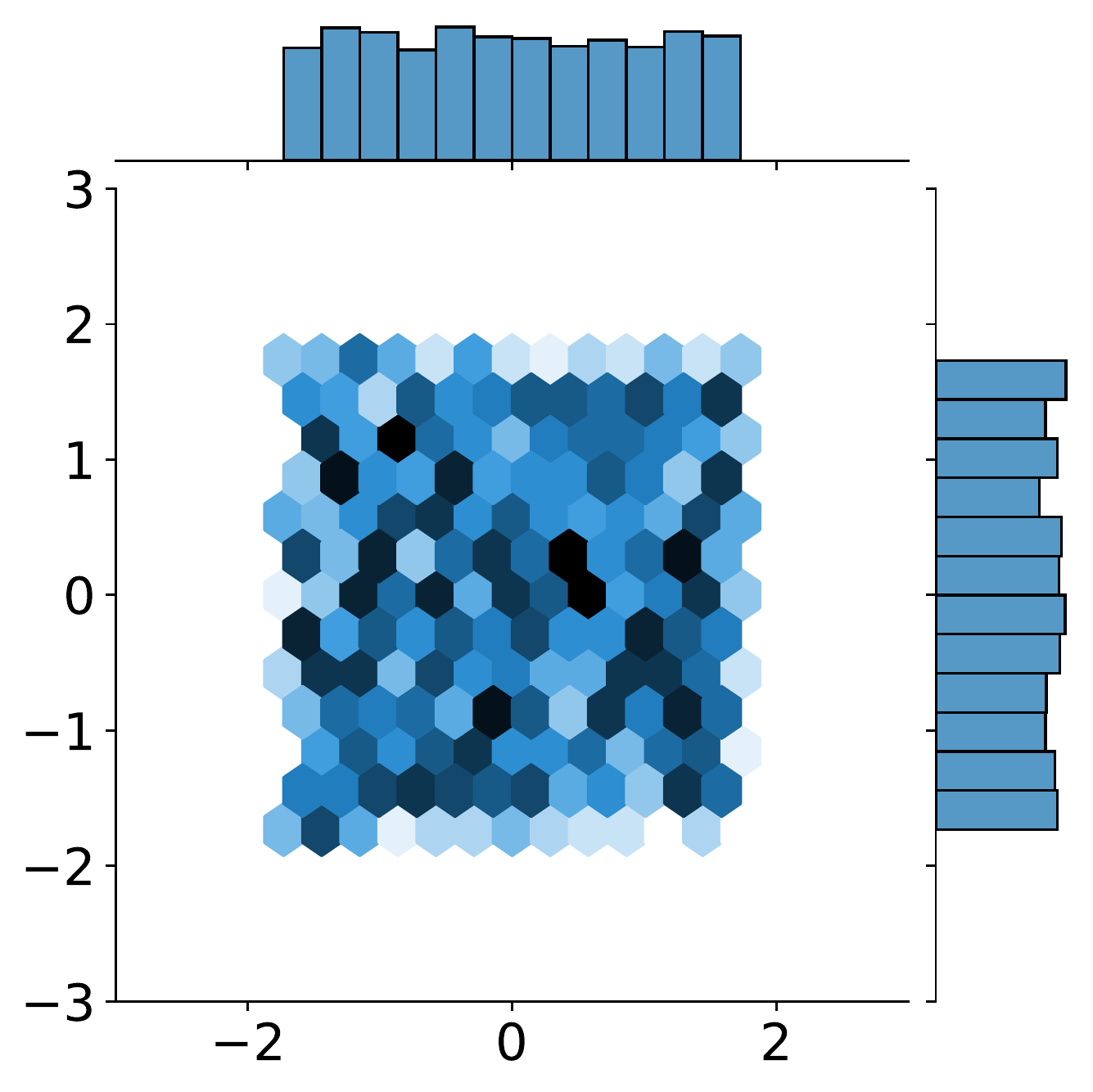}
		\end{minipage}
	}
   \subfigure[m=8]{
		\begin{minipage}[c]{.17\linewidth}
			\centering
			\includegraphics[width=3.0cm]{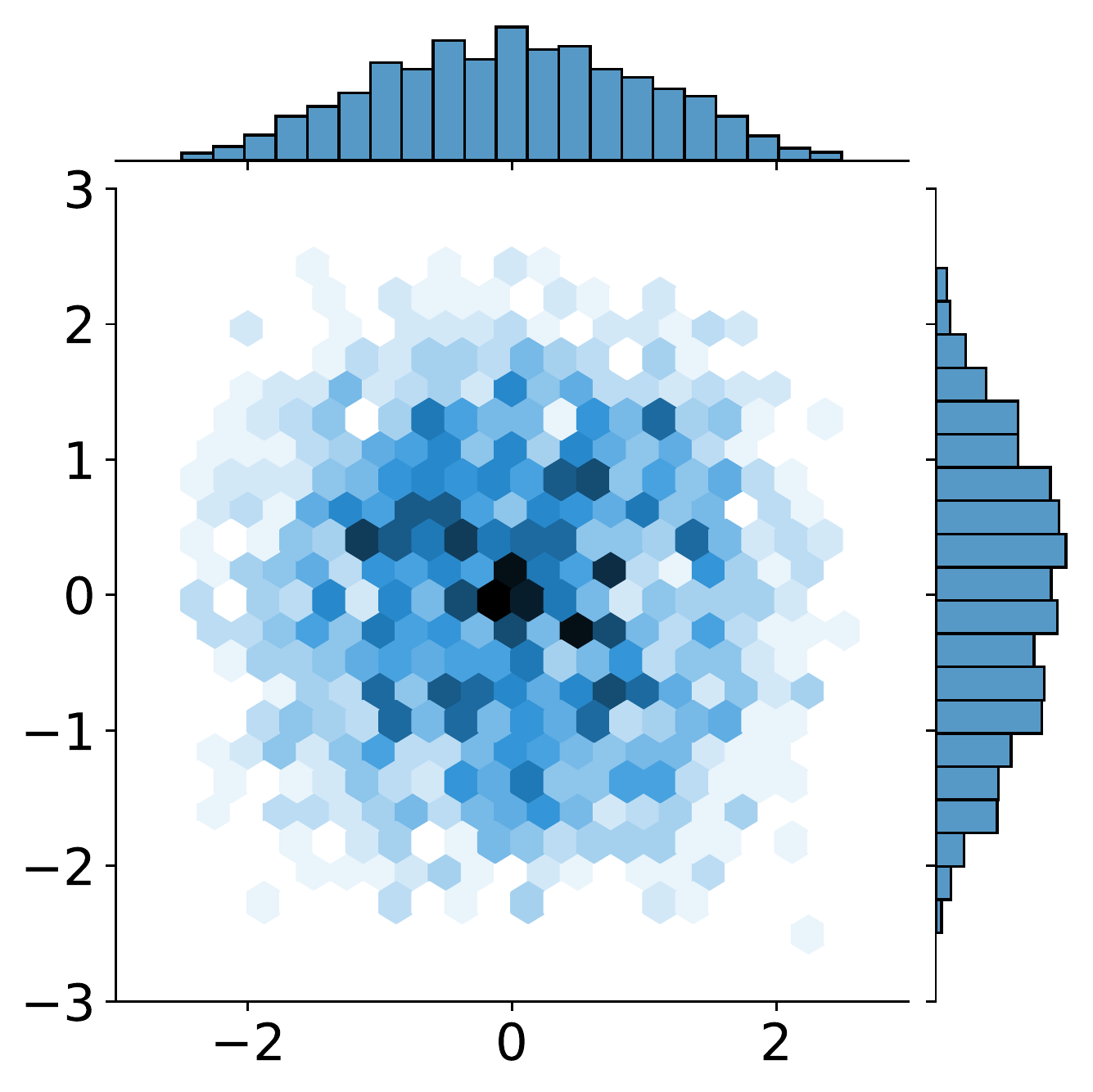}
		\end{minipage}
	}
	\subfigure[m=16]{
		\begin{minipage}[c]{.17\linewidth}
			\centering
			\includegraphics[width=3.0cm]{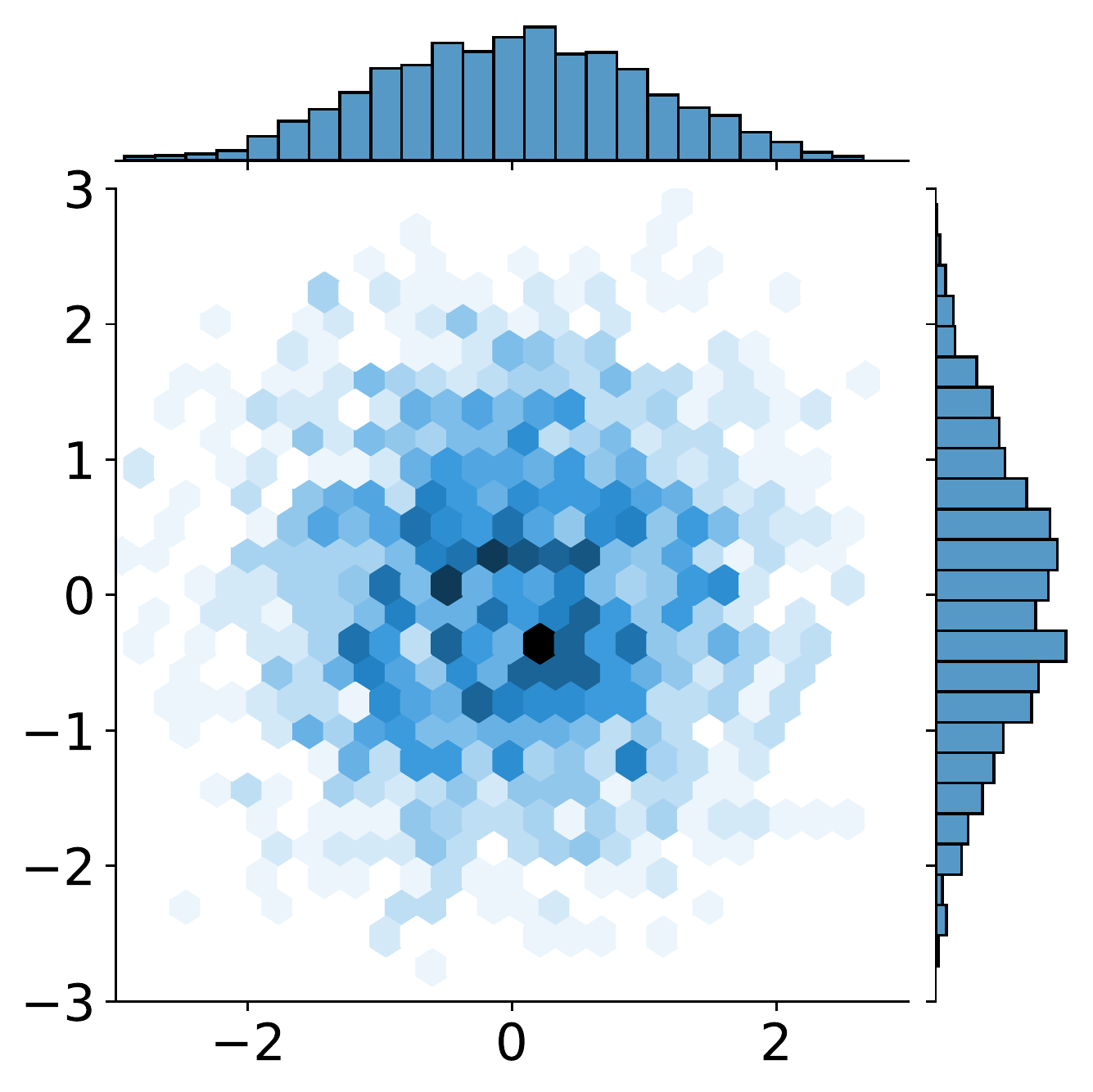}
		\end{minipage}
	}
	%\vspace{-0.08in}
	\caption{Illustration of the normalized output of GN/BN. We perform normalization over  1,680 examples sampled from a Gaussian distribution, varying the channel number for each group $c$ of GN (the upper subfigures) and the batch size $m$ of BN (the lower subfigures). We plot the bivariate histogram (using hexagonal bins)  of the  normalized output in the two-dimensional subspace, and marginal histograms (using rectangular bins) in the one-dimensional subspace.  }
	\label{fig:constrain_plot}
	\vspace{-0.0in}
\end{figure*}

\begin{table*}[t]
	\centering
	%\vspace{-0.15in}
	%\begin{footnotesize}
	\begin{tabular}{c|p{0.8in}<{\centering}  p{0.75in}<{\centering}|p{0.75in}<{\centering} p{0.8in}<{\centering}}
		%	\begin{tabular}{p{1in} p{0.8in}<{\centering}  p{0.8in}<{\centering}  p{0.8in}<{\centering} p{0.8in}<{\centering} p{0.8in}<{\centering}}
		\toprule[1pt]
		& \multicolumn{2}{c| }{Normalization along a batch} &    \multicolumn{2}{c}{Normalization along a group of neurons}  \\
		& BN    & BW &  GN &  GW\\
		\hline
		$\zeta(\phi; \mathbf{X})$ &  $2d$ &  $\frac{d(d+3)}{2}$  &  $2gm$ & $\frac{mg(g+3)}{2}$  \\
		$\zeta(\phi; \mathbb{D})$ &  $\frac{2Nd}{m}$ &    $\frac{Nd(d+3)}{2m}$ & $2gN$ & $\frac{Ng(g+3)}{2}$  \\
		Ranges of $m/g$ &  $m \geq 2$ &  $m \geq \frac{d+3}{2}$ & $g \leq \frac{d}{2}$ &    $g \leq \frac{\sqrt{8d+9}-3}{2}$ \\
		\hline	
	\end{tabular}
	%	\end{footnotesize}
	\vspace{0.05in}
	\caption{Summary of $\zeta(\phi; \mathbf{X})$, $\zeta(\phi; \mathbb{D})$ and ranges of $m/g$ for normalization methods. The analysis can be naturally extended to CNN, following how BN (GN) extents from MLP to CNN shown in Section~\ref{sec_method}. For examples,
  the number of neurons to be normalized for GN/GW  is $d=d'HW$, and the number of samples to be normalized for BN/BW is $m=m'HW$, given the input $\mathbf{X}\in\mathbb{R}^{d'\times m'\times H\times W}$ for CNN.}
	\label{table:Constrains}
	\vspace{-0.1in}
	%\vspace{-0.12in}
\end{table*}

\vspace{-0.05in}
\section{Revisiting the Constraint of Normalization}
%\vspace{-0.05in}
\label{sec_theory}

The normalization operation ensures  that the normalized output  $\widehat{\mathbf{X}}=\phi(\mathbf{X}) \in \mathbb{R}^{d \times m}$ has a stable distribution. This stability of  distribution can be implicitly viewed as the constraints imposed on  $\widehat{\mathbf{X}}$, which can be represented as a system of equations $\Upsilon_{\phi}(\widehat{\mathbf{X}})$. For example, BN provides the constraints $\Upsilon_{\phi_{BN}}(\widehat{\mathbf{X}})$ as:
{\setlength\abovedisplayskip{3pt}
	\setlength\belowdisplayskip{3pt}
	\begin{eqnarray}
	\label{eqn:constrain-BN}
	\sum_{j=1}^{m} \widehat{\mathbf{X}}_{ij}=0 ~ and ~ \sum_{j=1}^m \widehat{\mathbf{X}}_{ij}^2 - m = 0, ~ for  ~i=1, ..., d.
	\end{eqnarray}
}
%It is interesting to further investigate the magnitude of the constraints of normalization methods.
%  Here, we define the \textbf{constraint number} $\zeta(\phi; \mathbf{X}) $ for the normalization operation $\phi()$, as the number of equations in $\Upsilon_{\phi}(\widehat{\mathbf{X}})$. Clearly, the constrain number can
\hspace{-0.05in}Here, we define the \textbf{constraint number} of normalization to quantitatively measure the magnitude of the constraints provided by the normalization method.
%thus the stability of the distribution of the normalized output.
\vspace{-0.02in}
%	\begin{small}
\begin{defn}
	\label{def1:cn}
	Given the input data $\mathbf{X} \in \mathbb{R}^{d \times m}$, the \textbf{constraint number} of a normalization operation $\phi(\cdot)$, referred to as $\zeta(\phi; \mathbf{X})$,  is the number of independent equations in $\Upsilon_{\phi}(\widehat{\mathbf{X}})$.
\end{defn}
%	\end{small}
\vspace{-0.02in}
%It's clear these constraints are represented as $2d$ equations
% Here, we will provide an  analysis on this constraints relating to diffent normalization methods, and we expect the analysis will benefits better understanding and applying normalization methods.
As an example, we have $\zeta(\phi_{BN}; \mathbf{X})= 2 d$ from Eqn.~\ref{eqn:constrain-BN}. Furthermore, given training data  $\mathbb{D}$ of size $N$, we consider the optimization algorithm with batch size $m$ (we assume $N$ is divisible by $m$). We calculate the constraint number of normalization over the entire training data $\zeta(\phi; \mathbb{D})$.
Table \ref{table:Constrains} summarizes the constraint numbers of the main normalization methods discussed in this paper (please refer to the \TODO{\SM}~\ref{sup:constraint} for derivation details). We can see that the whitening operation provides significantly stronger constraints than the standardization operation. Besides, the constrains get stronger for BN (GN), when reducing (increasing) the batch size (group number).

%The results for different normalization methods are shown in Table \ref{table:Constrains}.
%With the constrain number in hand, we first provide the analysis on how the batch size $m$ (group number $g$) affects
% It's interesting to further consider $\Upsilon(\phi; \mathbb{D})$ where the constraints number is over the whole dataset $\mathbb{D}$ and the training is based on mini-batch $m$ (We assume $N$ is divided by $m$).
%\vspace{-0.08in}
%\paragraph{Connection to solution of system of equations}

\subsection{Constraint on  Feature Representation }
\label{sec_theory_Feature}
BN's benefits in accelerating the training of DNNs are mainly attributed to two reasons: 1) The distribution is  more stable  when fixing the first and second momentum of the activations, which reduces the internal covariant shifts~\cite{2015_ICML_Ioffe}; 2) The landscape of the optimization objective is better conditioned~\cite{2018_CVPR_Huang,2018_NIPS_shibani}, by improving the conditioning of the activation matrix with normalization. %he improved conditioning of the activation matrix, which is more foee to improve  the conditioning of the optimization problem.
Based on these arguments, GW/GN should have better performance when increasing the group number, due to the stronger constraints and better conditioning. However, we experimentally observe that GN/GW has significantly degenerated performance when the group number is too large (Figure~\ref{fig:MLP_BatchSize} (b)), which is similar to the small-batch-size problem of BN/BW. We investigate the reason behind this phenomenon.
%(whereas the channel number in each group $c$ is too small)

%\paragraph{Mathematic range of Group Number:}
We first show that the batch size/group number has a value range, which can be mathematically derived.
The normalization operation can be regarded as a way to find a solution $\widehat{\mathbf{X}}$ satisfying the constraints  $\Upsilon_{\phi}(\widehat{\mathbf{X}})$. To ensure the solution is feasible, it must satisfy the following condition:
% feaIt is necessary to ensure that the number of equations in $\Upsilon_{\phi}(\widehat{\mathbf{X}})$, that is $\zeta(\phi; \mathbf{X}) $, larger than or equal to the number of variables in  $\widehat{\mathbf{X}}$, referred to as  $\chi(\widehat{X})= md$.
\begin{small}
{\setlength\abovedisplayskip{5pt}
	\setlength\belowdisplayskip{5pt}
	\begin{eqnarray}
	\label{eqn:constrain-require}
	\zeta(\phi; \mathbf{X}) \leq \chi(\widehat{\mathbf{X}}),
	\end{eqnarray}
}
\end{small}
\hspace{-0.05in}where $\chi(\widehat{\mathbf{X}})= md$ is the number of variables in  $\widehat{\mathbf{X}}$. Based on Eqn. \ref{eqn:constrain-require}, we have $m>=2$ for BN to ensure a feasible solution. We also provide the ranges of batch size/group number for other normalization methods in Table \ref{table:Constrains}. Note that the batch size $m$ should be larger than or equal to $d$ to achieve a numerically stable solution for BW when using ZCA whitening in practice~\cite{2018_CVPR_Huang}. This also applies to GW, where $g$ should be less than or equal to $\sqrt{d}$.

\hspace{-0.05in}We then demonstrate that  normalization eventually affects the feature representation in a certain layer. Figure~\ref{fig:constrain_plot} shows the histogram of normalized output $\widehat{\mathbf{X}}$, by varying $c$ of GN\footnote{Note that the channel number in each group $c=\frac{d}{g}$. We vary $c$, rather than $g$, for simplifying the discussion.} and $m$ of BN.
%Before figuring out this intriguable phenomeno, we plot the normlized output by GN, with vary number of neuarons c in each group, as Shown in Figure.
 We observe that: 1) the values of $\widehat{\mathbf{X}}$ are heavily constrained if $c$ or $m$ is too small, \eg, the value of $\widehat{\mathbf{X}}$ is constrained to be $\{-1, +1 \}$ if $c=2$; 2) $\widehat{\mathbf{X}}$ is not Gaussian if $c$ or $m$ is too small, while BN/GN aims to produce a normalized output with a Gaussian distribution.  We believe that the constrained feature representation caused by GN/GW with a large group number is the main factor leading to the degenerated performance of a network. Besides, we also observe that the normalized output of GN is more correlated than that of BN, which supports the claim that BN is more capable of improving the conditioning of activations than GN, as shown in Section ~\ref{sec_method}.

\begin{figure}[t]
\vspace{-0.1in}
	\centering
\hspace{-0.25in}	\subfigure[]{
		\begin{minipage}[c]{.43\linewidth}
			\centering
			\includegraphics[width=4.2cm]{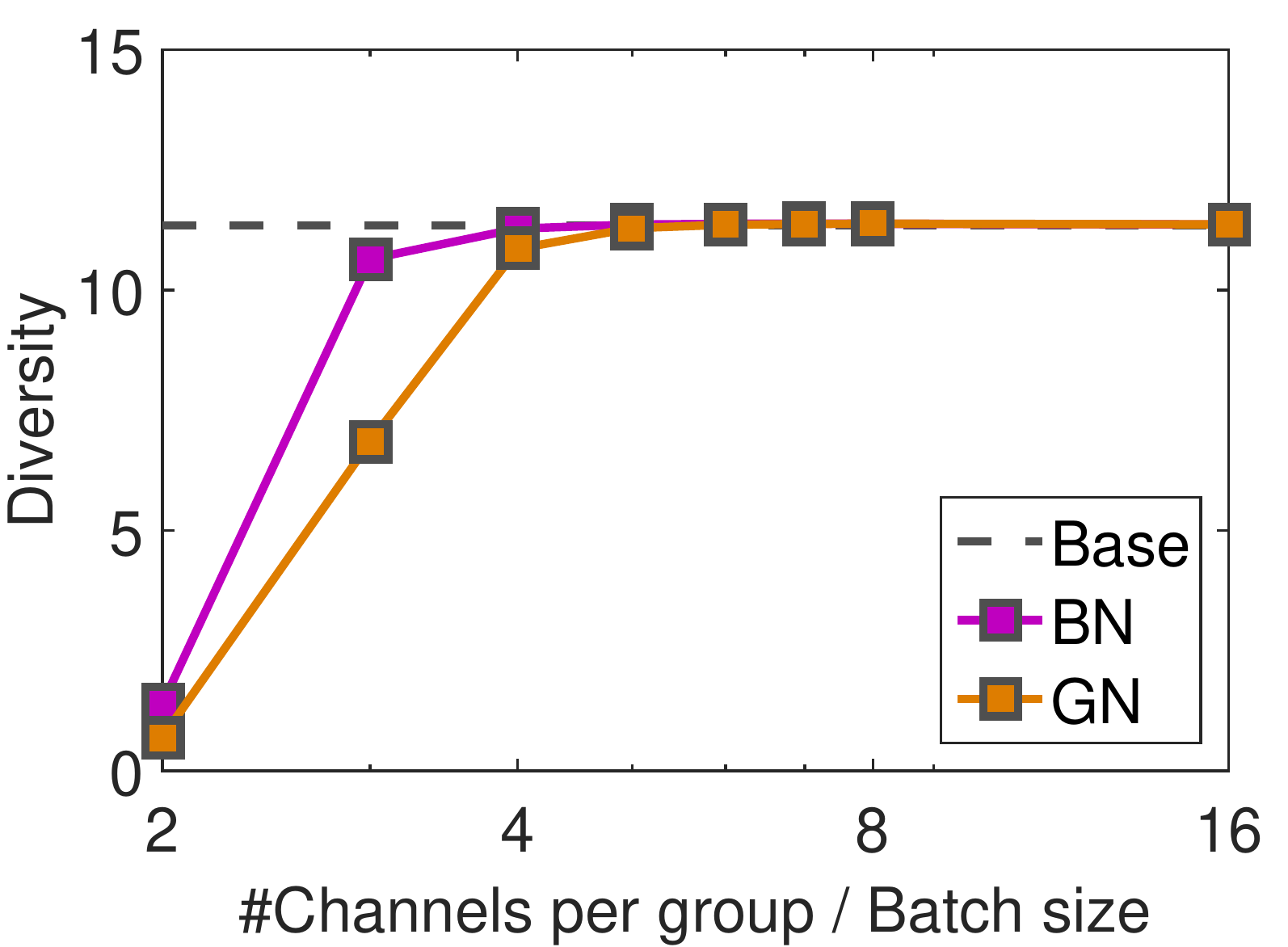}
		\end{minipage}
	}
\hspace{0.15in}		\subfigure[]{
		\begin{minipage}[c]{.43\linewidth}
			\centering
			\includegraphics[width=4.2cm]{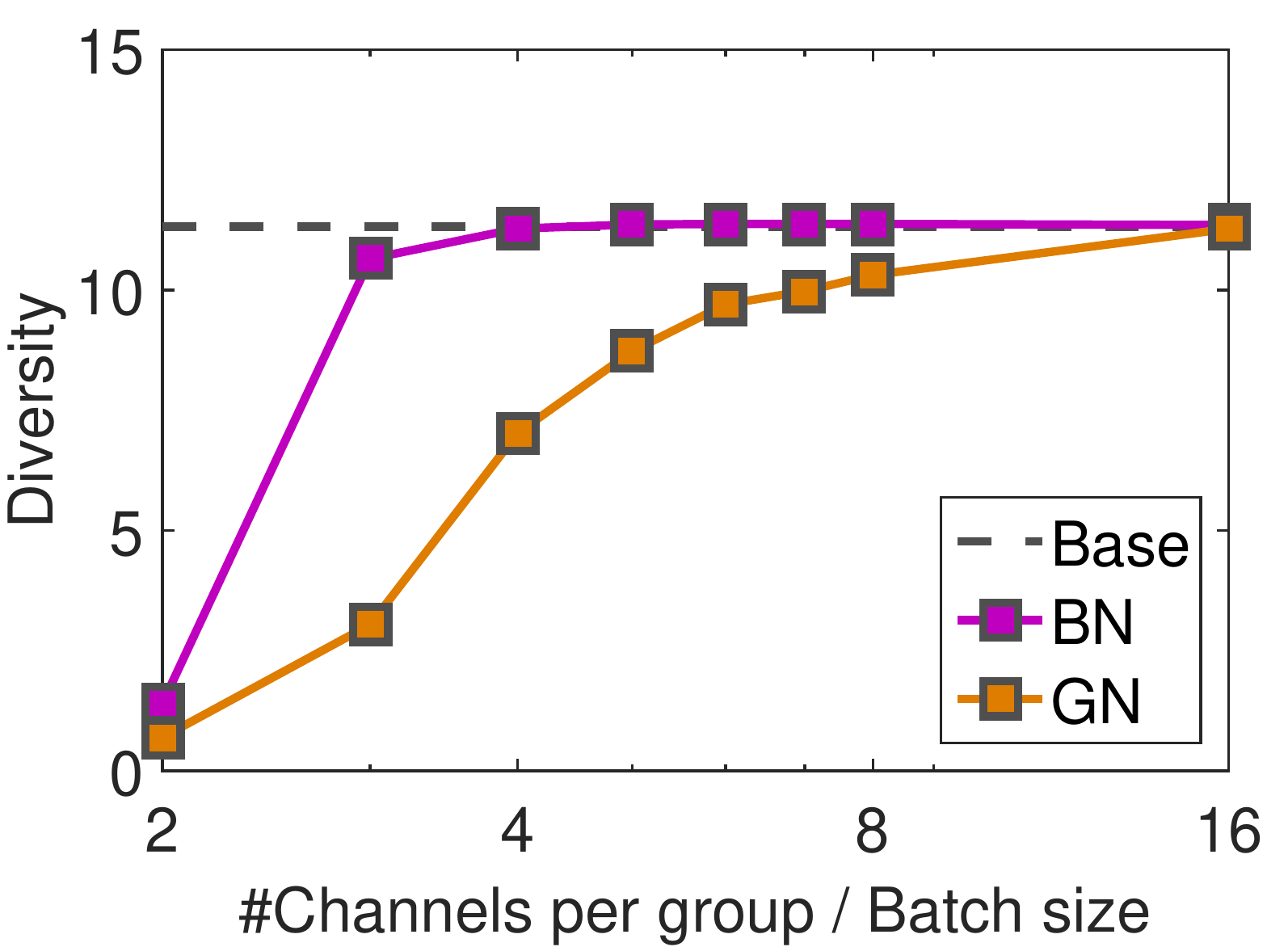}
		\end{minipage}
	}	
	\vspace{-0.05in}
	\caption{Diversity of group (batch) normalized features when varying the channels per group (batch size). We sample $N=1,680,000$ examples and use $1,000^2$ bins. We use the sampled Gaussian data as features in (a) and the output of a one-layer MLP in (b). Here, `Base' indicates the diversity of unnormalized features.}
	\label{fig:diversity}
	\vspace{-0.16in}
\end{figure}

%=\{\tilde{x}_i| \tilde{x}_i= f(x_i)\}, i=1,...,N
We also seek to quantitatively measure the representation of a feature space. Given a set of features $\widetilde{\mathbb{D}}\in \mathbb{R}^{ d \times N}$ extracted by a  network,  we assume the examples of $\widetilde{\mathbb{D}}$ belong in a $d$-dimensional hypercube $V=[-1,1]^d$ (we can ensure that this assumption holds by dividing the maximum absolute value of each dimension). Intuitively, a powerful feature representation    implies that the examples from $\widetilde{\mathbb{D}}$ spread  over $V$  with large diversity, while a weak representation indicates that they are limited to certain values without diversity. We thus define the diversity of $\widetilde{\mathbb{D}}$ based on the information entropy as follows, which can empirically indicate the representation ability of the feature space to some degree:
%We use dismeniton-wise entorpyt to meahichds the posieiant repaeitons. We bin the saplece of the input based on its maximeua and minums value, and divid the rangighe into bins. we count the neumber of
%To  investigate how the constraint number affects the feature representation of normalized output, we define the diversity of a set of data $\mathbb{D}=\{x_1, x_N\}$ as:
\begin{small}
{\setlength\abovedisplayskip{5pt}
	\setlength\belowdisplayskip{5pt}
	\begin{eqnarray}
	\label{eqn:diversity}
	\Gamma_{d,T}(\widetilde{\mathbb{D}})= \sum_{i=1}^{T^d} p_i \log p_i.
	\end{eqnarray}
}
\end{small}
\hspace{-0.05in}Here, $V$ is evenly divided into $T^d$ bins, and $p_i$ denotes the probability of an example belonging to the $i$-th bin.
We  can thus calculate $\Gamma_{d,T}(\widetilde{\mathbb{D}})$ by sampling enough examples. However, calculating $\Gamma_{d,T}(\widetilde{\mathbb{D}})$ with reasonable accuracy  requires  $O(T^d)$ examples  to be sampled from a $d$-dimensional space. We thus only calculate $\Gamma_{2,T}(\widetilde{\mathbb{D}})$  in practice by sampling two dimensions, and average the results. We show the diversity of  group (batch) normalized features by varying the channels of each group (batch size) in Figure~\ref{fig:diversity}, from which we can obtain similar conclusions  as in Figure~\ref{fig:constrain_plot}.

In summary, our qualitative  and quantitative analyses show that group/batch based normalizations have low diversity of feature representations when $c/m$ is  small. We believe these constrained feature representations  affect the performance of a network, and can lead to significantly deteriorated results when the representation is over-constrained.
  %is the main factor that results in degenerated performance.
%2) the Gausssimon assupiton on normlaized outptu is also not true when m/c is too small.
 %We believe  these behojeve affects the prfomanece degiejema, when goubnmuber is small and small-batch size problems.

%
%\TODO{Show the case that large group size}
%As illustrated in Fiague 1, that GN/GW can achieve better conditoing, thus, can more benifeit optimizaoton. Howve, we expeimentatl find that GN/GW has worse perfoamnce when the group size is bigger, as shown in Figure \TODO{cite}. What's the reason behind it? We will explore this prolbme in this sections.
%
%\TODO{Qualitively shows: SHow the figure that the activaiton distribution}
%
%\TODO{Quantitivel yshow the distribution diversity.}
%
%\TODO{Deifent the layer representation ability, which is based on the number of the constrain number.}
%
%\TODO{Connected to the representation ability of the models}

%\begin{table}[t]
%	\centering
%			\caption{The number of constraints for normalization methods for the input $\mathbf{X} \in \mathbb{R}^{d \times m}$.}
%					\label{table:Constrains}
%	%\begin{footnotesize}
%		\begin{tabular}{c|cccccc}
%		\toprule[1pt]
%			Method     & BN   & LN & GN & BW & Group-Based BW & GW\\
%			\hline
%			$\zeta(\phi_{BN}; \mathbf{X})$ &  $2d$ & $2m$ & $2gm$ & $\frac{d(d+3)}{2}$ & $\frac{d(c+3)}{2}$ & $\frac{mg(g+3)}{2}$  \\
%			$\zeta(\phi_{BN}; \mathbb{D})$ &  $\frac{2Nd}{m}$ & $2N$ & $2gN$ & $\frac{Nd(d+3)}{2m}$ & $\frac{Nd(c+3)}{2m}$ & $\frac{Ng(g+3)}{2}$  \\
%			Boundary of $m/g$ &  $m \geq 2$ & --- & $g \leq \frac{d}{2}$ & $m \geq \frac{d+3}{2}$ & $m \geq \frac{c+3}{2}$ &    $g \leq \frac{\sqrt{8d+9}-3}{2}$ \\
%      \bottomrule[1pt]		
%		\end{tabular}
%		%	\vspace{0.1in}
%%	\end{footnotesize}
%	%\vspace{-0.22in}
%\end{table}

\begin{figure*}[t]
	\vspace{-0.2in}
	\centering
\hspace{-0.25in}	\subfigure[Linear classifier]{
		\begin{minipage}[c]{.3\linewidth}
			\centering
			\includegraphics[width=5.6cm]{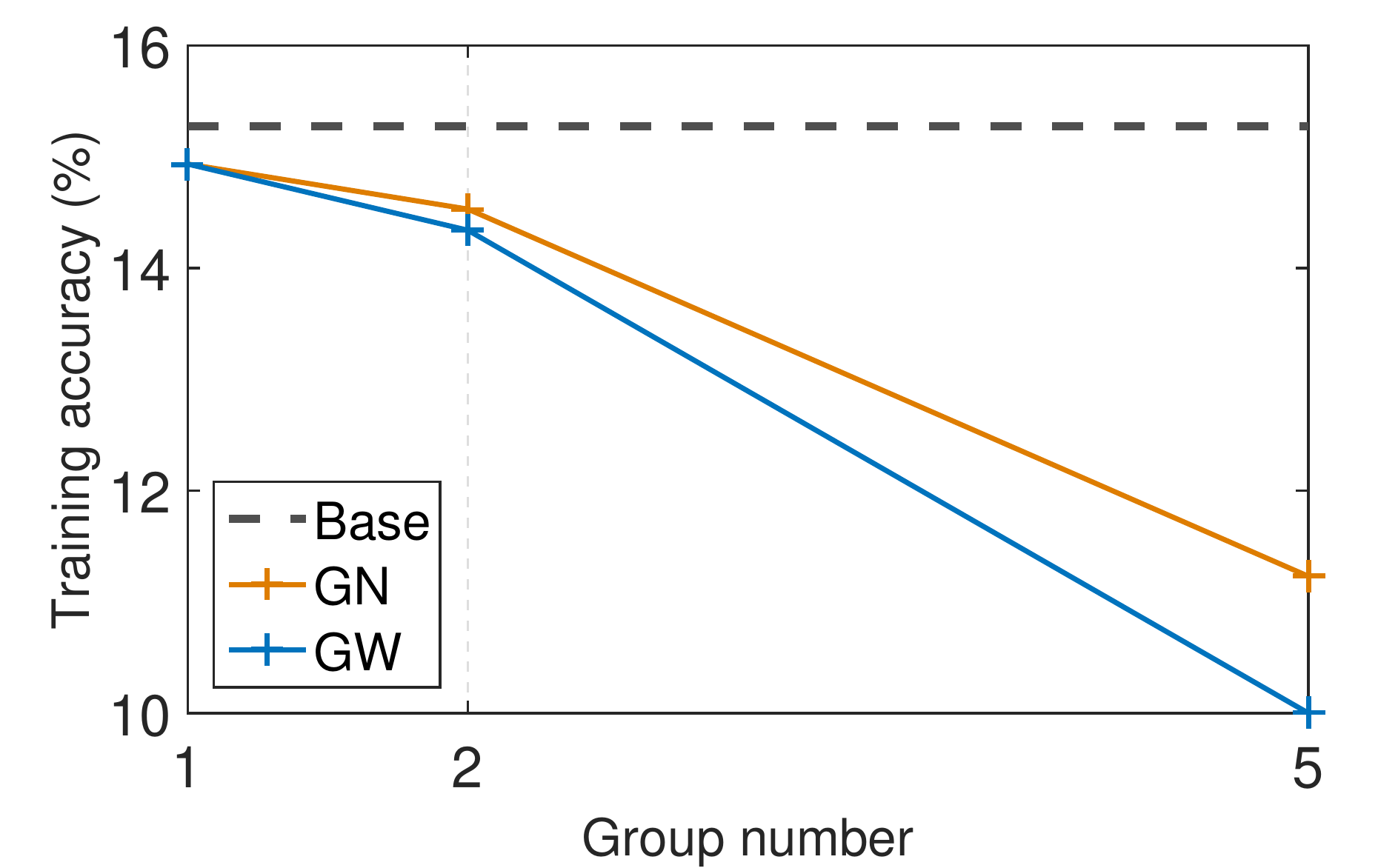}
		\end{minipage}
	}
\hspace{0.15in}		\subfigure[One-layer MLP]{
		\begin{minipage}[c]{.3\linewidth}
			\centering
			\includegraphics[width=5.6cm]{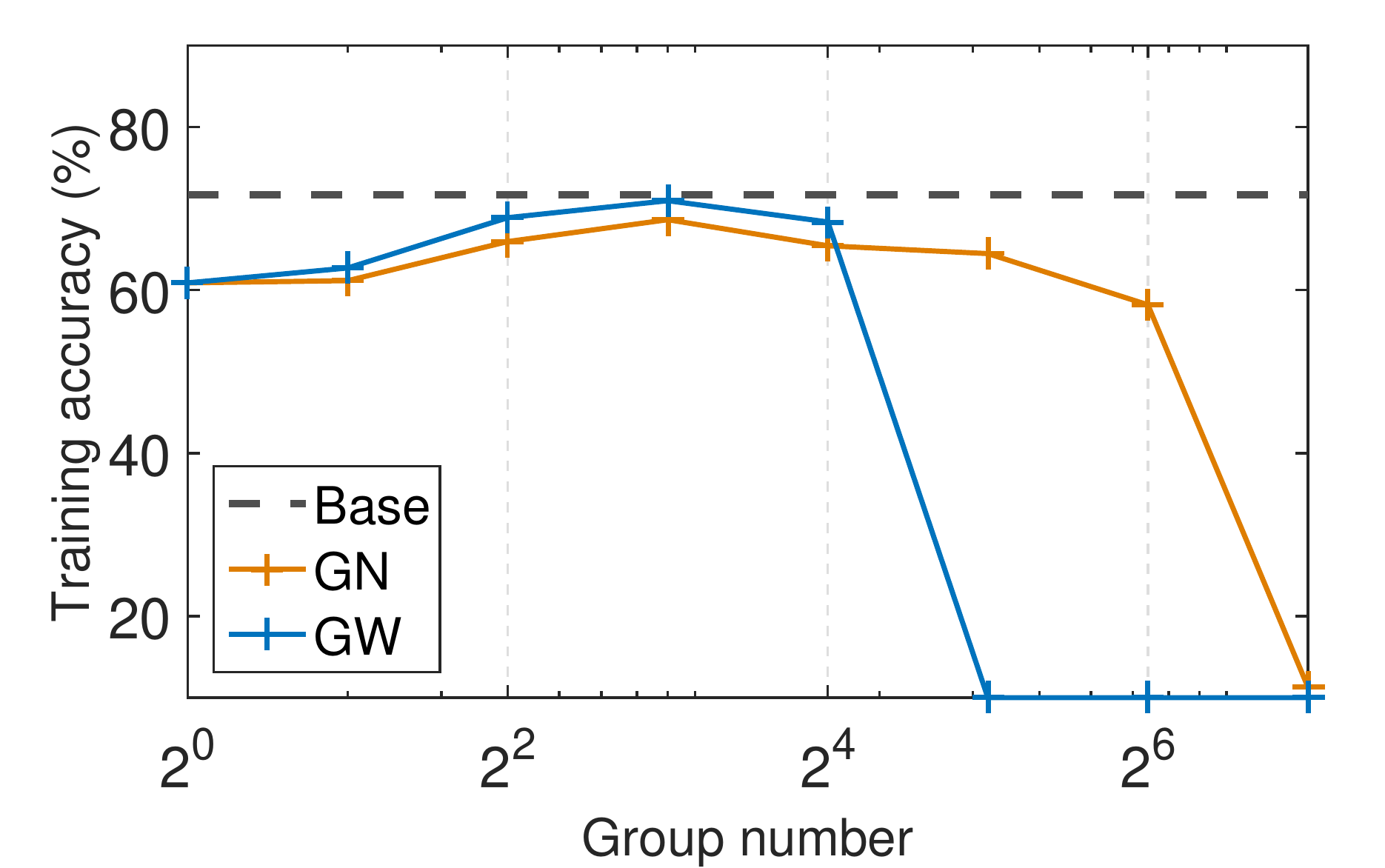}
		\end{minipage}
	}	
\hspace{0.15in}		\subfigure[Four-layer MLP]{
		\begin{minipage}[c]{.3\linewidth}
			\centering
			\includegraphics[width=5.6cm]{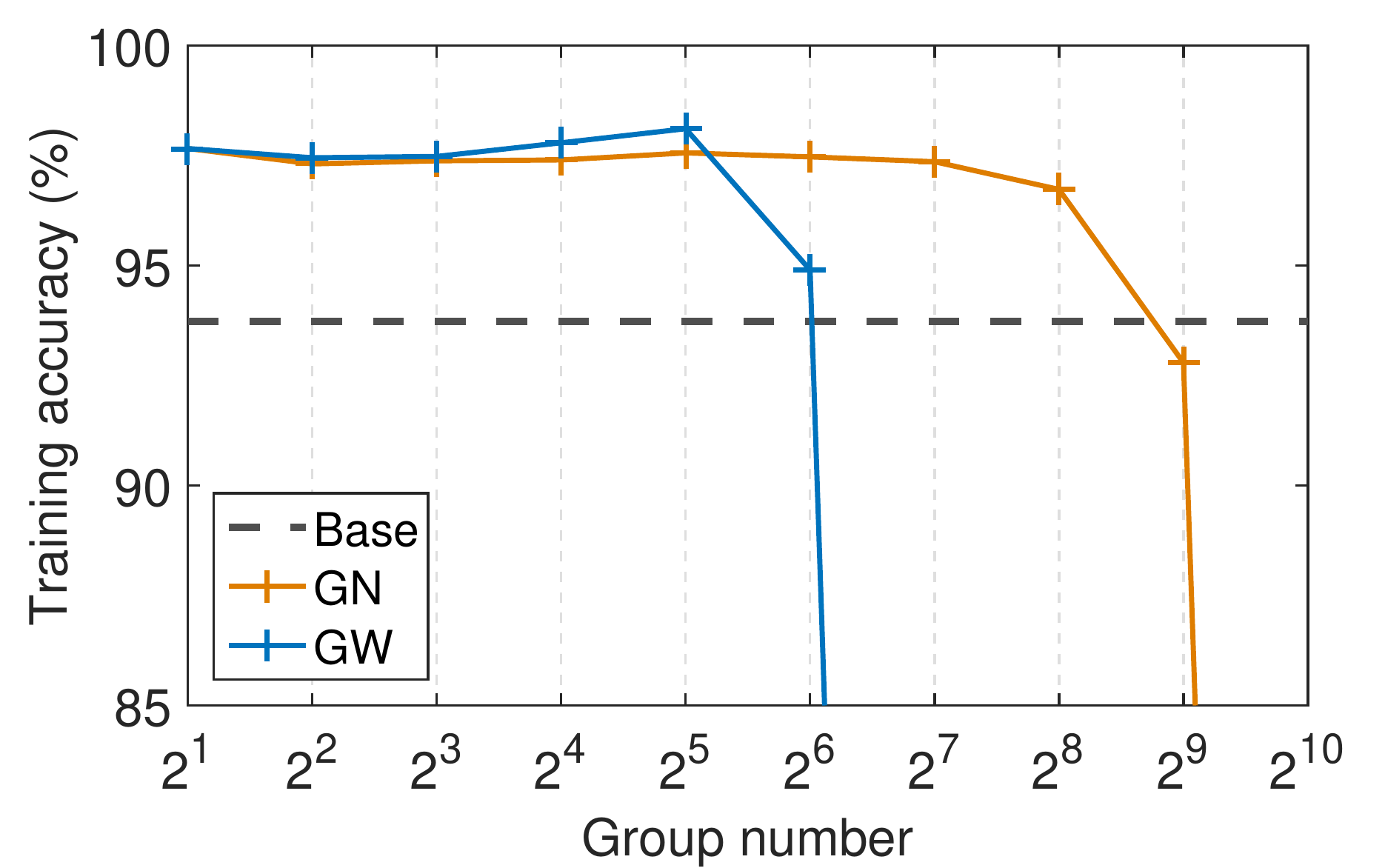}
		\end{minipage}
	}	
	\vspace{-0.05in}
	\caption{Comparison of model representational capacity when fitting random labels~\cite{2017_ICLR_Zhang} on MNIST dataset using different architectures. We vary the group number of GN/GW and evaluate the training accuracy. `Base' indicates the model without normalization. (a) Linear classifier; (b) One-layer MLP with 256 neurons in each layer; (c) Four-layer MLP with 1,280 neurons in each layer.}
	\label{fig:MLP_RandomLabel}
	\vspace{-0.16in}
\end{figure*}

%\begin{figure*}[t]
%	\centering
%	%\vspace{-0.15in}
%	\hspace{-0.3in}	\subfigure[Training accuracy]{
%		\begin{minipage}[c]{.46\linewidth}
%			\centering
%			\includegraphics[width=6.4cm]{./figures/ImageNet/Group-Train.pdf}
%		\end{minipage}
%	}
%	\hspace{0.1in}	\subfigure[Validation accuracy]{
%		\begin{minipage}[c]{.46\linewidth}
%			\centering
%			\includegraphics[width=6.4cm]{./figures/ImageNet/Group-Test.pdf}
%		\end{minipage}
%	}
%	\vspace{-0.06in}
%	\caption{Effects of group number of GW/GN on ResNet-50 for ImageNet classification. We evaluate the top-1 training and validation accuracies.}
%	\label{fig:exp_groupsize}
%	\vspace{-0.15in}
%\end{figure*}

%\subsection{Analysis on Representational Capacity}
\subsection{Effect on Representational Capacity of Model}
\label{sec:representationFeature}
\vspace{-0.05in}
The constraints  introduced by normalization are believed to affect the representational capacity of neural networks~\cite{2015_ICML_Ioffe}, and thus the learnable scale and shift parameters are used to recover the representations~\cite{2015_ICML_Ioffe,2016_CoRR_Ba,2018_CVPR_Huang,2018_ECCV_Wu}. However, such an argument is seldom
validated by either theoretical or empirical analysis.
 Theoretically analyzing the complexity measure (\eg, VC dimensions~\cite{1999_TNN_Vapnik} or the number of linear regions~\cite{2014_NeurIPS_Montufar,2020_ICML_Xionghuan}) of the representational capacity of neural networks with normalization  is a challenging task, because normalized networks do not follow the assumptions for calculating linear regions or VC dimensions. Here, we conduct preliminary experiments, seeking to empirically show how normalization affects the representational capacity of a network, by varying the constraints imposed on the feature.
 %Due to space limitations, we only provide essential components of the experimental setup; for more details, please refer to the \TODO{\SM}.
 %Note that it is more difficult to fit the random label on MNIST than on CIFAR-10 datasets, since the examples of MNIST is more diffcult to tell with the random label (e.g., the very like digit 5 is labeled to 2 or 3, which can be viewed as the noise)

We follow the non-parametric randomization tests fitting random labels~\cite{2017_ICLR_Zhang} to empirically compare the representational capacity of neural networks. To rule out the optimization benefits introduced by normalization,
we first conduct experiments using a linear classifier, where normalization is also inserted after the linear module. We train over $1,000$ epochs using stochastic gradient descent (SGD) with a batch size of 16, and report the best training accuracy among the learning rates in $\{0.001,0.005,0.01,0.05,0.1\}$ in Figure~\ref{fig:MLP_RandomLabel} (a). We observe that GN and GW  have lower training accuracy than when normalization is not used, which suggests that  normalization does indeed reduce the model's representational capacity in this case.  Besides, the accuracy of
GN/GW decreases as the group number increases. This suggests that the model may have weaker representational ability when increasing the constraints on the feature. Note that we have the same observations  regardless of whether or not the learnable scale and shift parameters of GN/GW are used.

To further consider the trade-off between the benefits of normalization on optimization  and its constraints on representation,
we  conduct experiments on the one-layer and four-layer MLPs. The results are shown in Figure~\ref{fig:MLP_RandomLabel} (b) and (c), respectively.
We observe that the model with GN/GW has significantly degenerated training accuracy when $g$ is too large,
which means that a large group number heavily limits the model's representational capacity by constraining the feature representation,  as discussed in Section~\ref{sec:representationFeature}. We note that GW is more sensitive to the group number than GN. The main reason is that $\zeta(\phi_{GW}; \mathbb{D})$ is quadratic to $g$, while $\zeta(\phi_{GN}; \mathbb{D})$ is linear to it, from Table~\ref{table:Constrains}.
 Besides, we observe that GN and GW still have lower training accuracy than `Base' on the one-layer MLP, but higher accuracy on the four-layer MLP if the group number $g$ is not too large. This suggests that the benefits of normalization  on optimization dominate if the model's representation is not too limited.
 We also observe that the best training accuracy of GW is higher than that of GN. We attribute this to the fact that the whitening operation is better for improving the conditioning of optimization, compared to standardization.
We also conduct similar experiments on convolutional neural networks (CNNs). Pleaser refer to the \TODO{\SM}~\ref{sub:representation_CNN} for details.

\vspace{-0.05in}
\subsection{Discussion of Previous Work}
\label{sec_theory_discuss}
\vspace{-0.05in}
Previous analyses on BN  are mainly derived from the perspective of optimization~\cite{2018_NIPS_shibani,2019_AISTATS_Lian,2019_AISTATS_Kohler,2019_ICML_Cai}. One argument is that BN can improve the conditioning of the optimization problem~\cite{2018_NIPS_shibani,2019_ICML_Cai,2019_ICML_Ghorbani,2019_NIPS_Karakida,2020_arxiv_Daneshmand}, either by avoiding the rank collapse of pre-activation matrices~\cite{2020_arxiv_Daneshmand,2020_ECCV_Huang} or alleviating the pathological sharpness of the landscape~\cite{2018_NIPS_shibani,2019_NIPS_Karakida,2020_ECCV_Huang}. This argument has been further investigated by computing the spectrum of the Hessian for a large-scale dataset~\cite{2019_ICML_Ghorbani}. The improved conditioning enables large learning rates, thus improving the generalization~\cite{2018_NIPS_Bjorck,2019_ICLR_Luo2}.
Another argument is that BN is scale invariant~\cite{2015_ICML_Ioffe,2016_CoRR_Ba}, enabling it to adaptively adjust the learning rate \cite{2017_NIPS_Cho,2018_arxiv_Hoffer,2019_ICLR_Arora,2019_ICML_Cai,2019_ICLR_Zhang,2020_ICLR_Li}, which stabilizes and further accelerates training~\cite{2015_ICML_Ioffe,2016_CoRR_Ba}. Other analyses focus on investigating the signal and gradient propagation, either by exploiting mean-field theory \cite{2019_ICLR_Yang,2019_arxiv_Wei}, or a neural
tangent kernel (NTK)~\cite{2019_arxiv_Jacot}.

Different from these works, we are the first to investigate how BN/GN affects a model's representational capacity, by analyzing the constraint on the representation of internal features. This opens new doors in  analyzing and understanding normalization methods. We also investigate how batch size affects the training performance of batch normalized networks (Figure~\ref{fig:MLP_BatchSize} (a)), from the perspective of a model's representational capacity.
% by the introduced constrain number of normalization methods.  open a new venue in analysi the repreiament aboty of batch normlaizaed netowrk, and we show how the batch size affects the peofirannce by affects the model's repreoantioanton abliity.
Several works~\cite{2018_ACCV_Alexander,2019_CVPR_Huang,2020_CVPR_Huang} have shown that batch size is related to the magnitude of stochasticity~\cite{2018_CoRR_Andrei,2018_ICML_Teye} introduced by BN, which also affects the model's training performance.
%the stochaist8c analysis to shown that the  small batch size intrioduce overmuch sthaotcha ticha and afects the triang peramocat.
However, the stochasticity analysis~\cite{2019_CVPR_Huang} is specific to  normalization along the batch dimension, and cannot explain why GN with a large group number has significantly worse performance (Figure~\ref{fig:MLP_BatchSize} (b)). Our work provides a unified analysis for batch and group normalized networks.
% Our work provide a unified framework in analyzing BN/GN to feeef, from the peromant of model's moeoto.
%

%
%NOte that the previous works has show that the stochasticity introduce by normalization over batch data heavily affects the training and inference process. Here our constraint throty, from an another view, show that how the batch size can affects the training performancee. Further, our theory can explain why group normalization  has worse perforce with large group, while the sthocha emopew not. Our method provide a unified characteristic of the batch and group normalization, from the perspective of model representation capacity.

\begin{figure}[t]
	\centering
	%\vspace{-0.15in}
	\hspace{-0.2in}	\subfigure[Training accuracy]{
		\begin{minipage}[c]{.46\linewidth}
			\centering
			\includegraphics[width=4.3cm]{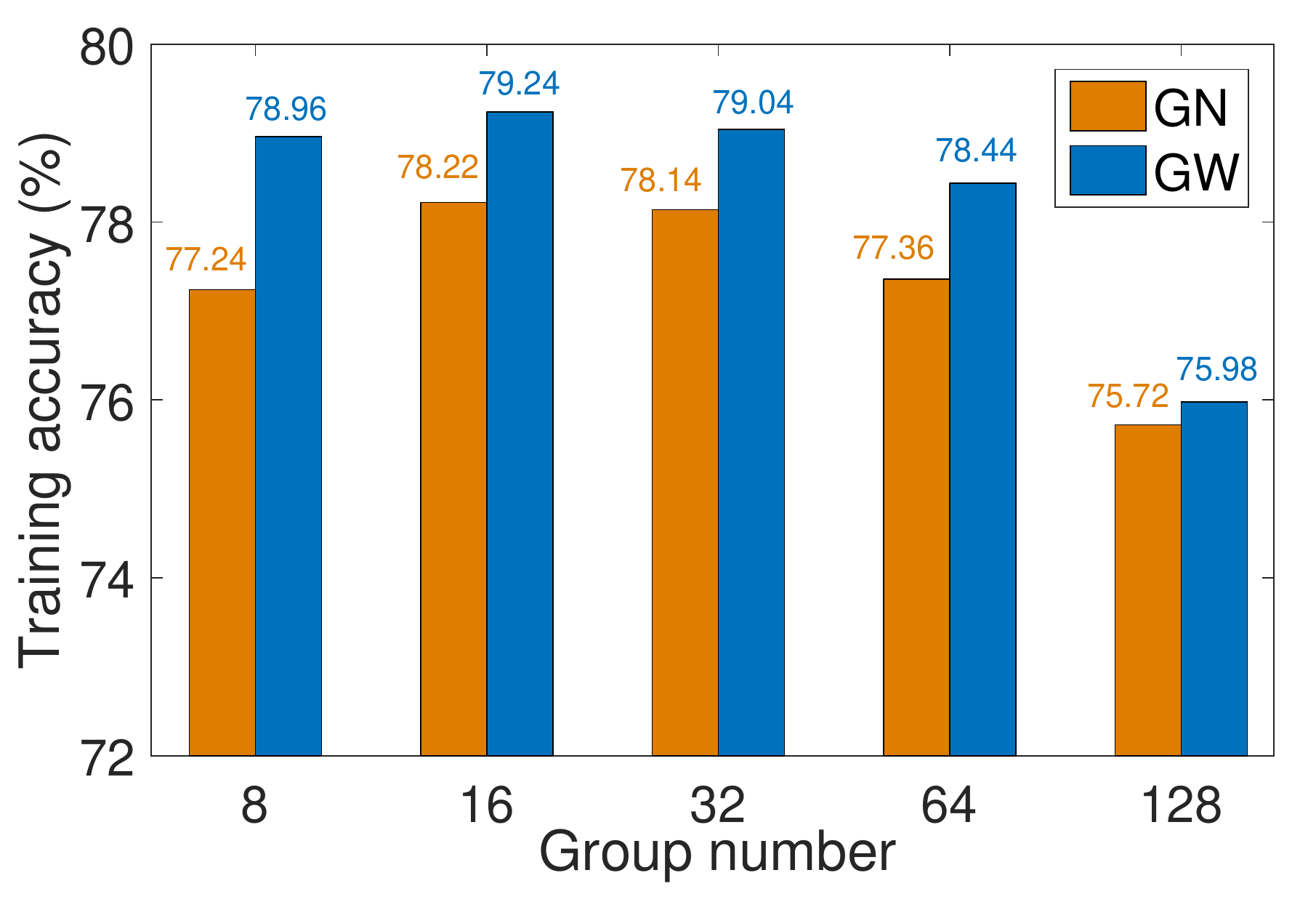}
		\end{minipage}
	}
	\hspace{0.1in}	\subfigure[Validation accuracy]{
		\begin{minipage}[c]{.46\linewidth}
			\centering
			\includegraphics[width=4.3cm]{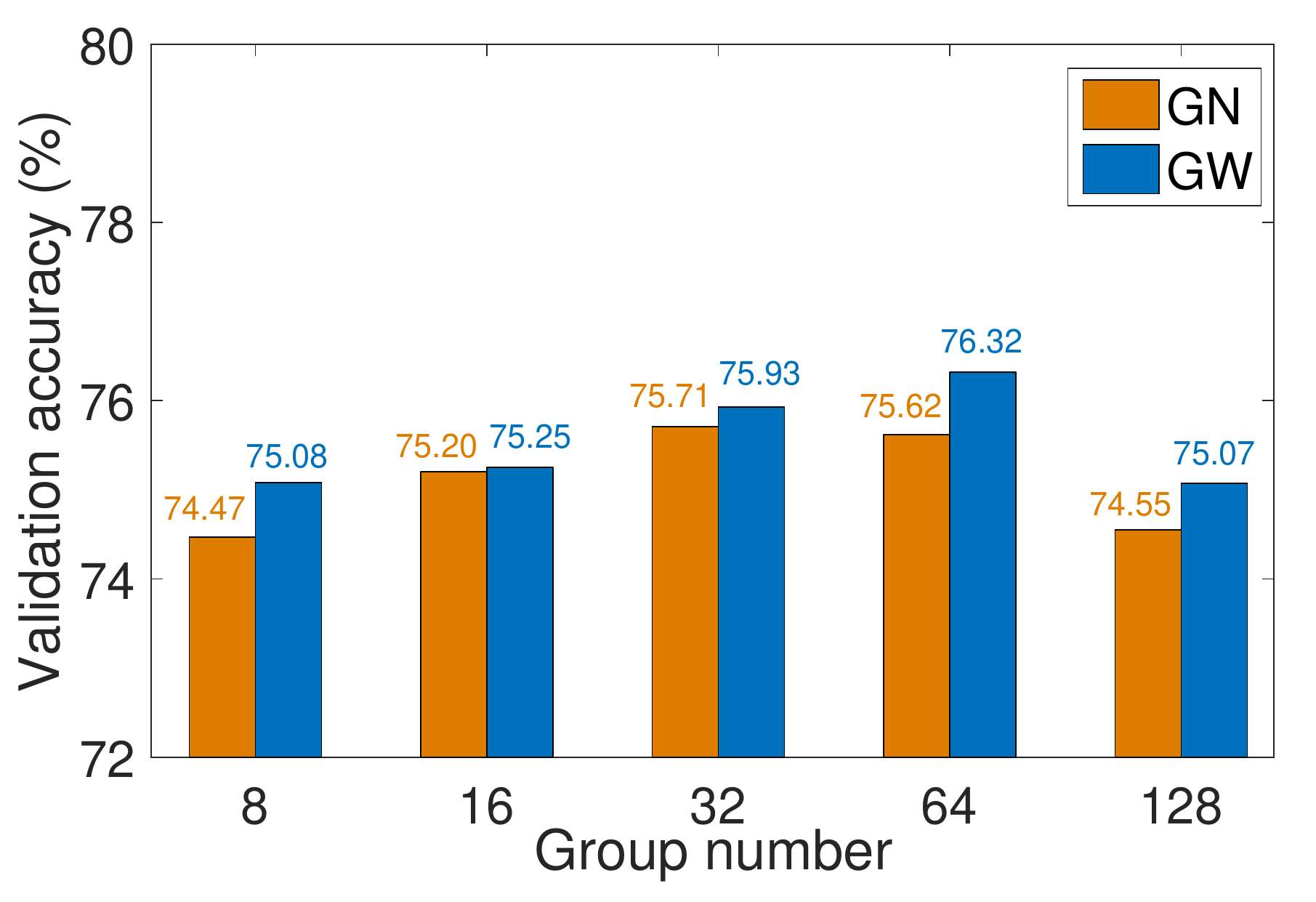}
		\end{minipage}
	}
	\vspace{-0.06in}
	\caption{Effects of group number of GW/GN on ResNet-50 for ImageNet classification. We evaluate the top-1 training and validation accuracies.}
	\label{fig:exp_groupsize}
	\vspace{-0.15in}
\end{figure}

\begin{table*}[t]
	\centering
	\vspace{-0.1in}
	%\begin{footnotesize}
	\begin{tabular}{c|lllll}
		\bottomrule[1pt]
		& ~~S1   & S1-B1 &  S1-B2 &  S1-B3 & S1-B12\\
		\hline
		Baseline (BN) & ~~76.23 & ~~76.23 & ~~76.23 & ~~76.23& ~~76.23  \\
		BW~\cite{2019_CVPR_Huang}     & ~~76.58 $_{(\textcolor{blue}{\uparrow 0.35})}$ & ~~76.68 $_{(\textcolor{blue}{\uparrow 0.45})}$
		&~~76.86 $_{(\textcolor{blue}{\uparrow 0.63})}$  &~~76.53 $_{(\textcolor{blue}{\uparrow 0.30})}$ &~~76.60 $_{(\textcolor{blue}{\uparrow 0.37})}$\\
		BW$_\Sigma$~\cite{2020_CVPR_Huang}     & ~~76.63 $_{(\textcolor{blue}{\uparrow 0.40})}$ & ~~76.80 $_{(\textcolor{blue}{\uparrow 0.57})}$
		&~~76.76 $_{(\textcolor{blue}{\uparrow 0.53})}$  &~~76.52 $_{(\textcolor{blue}{\uparrow 0.29})}$ &~~76.71 $_{(\textcolor{blue}{\uparrow 0.48})}$\\
		GW     & ~~\textbf{76.76} $_{(\textcolor{blue}{\uparrow 0.53})}$  & ~~\textbf{77.62} $_{(\textcolor{blue}{\uparrow 1.39})}$
		&~~\textbf{77.72} $_{(\textcolor{blue}{\uparrow 1.49})}$ &~~\textbf{77.47} $_{(\textcolor{blue}{\uparrow 1.24})}$ &~~\textbf{77.45} $_{(\textcolor{blue}{\uparrow 1.22})}$ \\
		\toprule[1pt]
	\end{tabular}
	\caption{Effects of position when applying GW on ResNet-50 for ImageNet classification. We evaluate the top-1 validation accuracy on five architectures (\textbf{S1}, \textbf{S1-B1}, \textbf{S1-B2}, \textbf{S1-B3} and \textbf{S1-B12}). }
	\label{table:ImageNet-Res50-ablation}
	\vspace{-0.10in}
	%	\end{footnotesize}
\end{table*}

\begin{table*}[t]
	\centering
	%\begin{footnotesize}
	\begin{tabular}{c|llll}
		\bottomrule[1pt]
		Method     & ResNet-50   &  ResNet-101 &  ResNeXt-50 &  ResNeXt-101\\
		\hline
		Baseline (BN)~\cite{2015_ICML_Ioffe} & ~~76.23 & ~~77.69 & ~~77.01 & ~~79.29  \\
		GN~\cite{2018_ECCV_Wu}   & ~~75.71 $_{(\textcolor{red}{\downarrow 0.52})}$ & ~~77.20 $_{(\textcolor{red}{\downarrow 0.49})}$
		& ~~75.69 $_{(\textcolor{red}{\downarrow 1.32})}$  &~~78.00 $_{(\textcolor{red}{\downarrow 1.29})}$   \\	
		BW$_\Sigma$~\cite{2020_CVPR_Huang}     & ~~77.21 $_{(\textcolor{blue}{\uparrow 0.98})}$ & ~~78.27 $_{(\textcolor{blue}{\uparrow 0.58})}$
		&~~77.29 $_{(\textcolor{blue}{\uparrow 0.28})}$  &~~79.43 $_{(\textcolor{blue}{\uparrow 0.14})}$ \\
		GW     & ~~\textbf{77.72} $_{(\textcolor{blue}{\uparrow 1.49})}$  & ~~\textbf{78.71} $_{(\textcolor{blue}{\uparrow 1.02})}$
		&~~\textbf{78.43} $_{(\textcolor{blue}{\uparrow 1.42})}$ &~~\textbf{80.43} $_{(\textcolor{blue}{\uparrow 1.14})}$  \\
		\toprule[1pt]
	\end{tabular}
	\caption{Comparison of validation accuracy on ResNets~\cite{2015_CVPR_He} and ResNeXts~\cite{2017_CVPR_Xie} for ImageNet. Note that we use an additional layer for BW$_\Sigma$ to learn the decorrelated features, as recommended in \cite{2020_CVPR_Huang}.}
	%Here, we apply GW and BW$_\Sigma$ in these models following the \textbf{S1-B2} architecture.}
	\label{table:ImageNet-Res-Step}
	\vspace{-0.12in}
	%	\end{footnotesize}
\end{table*}

%\vspace{-0.05in}
\section{Large-Scale Visual Recognition Tasks}
\vspace{-0.05in}
\label{sec_experiments}

We investigate the effectiveness of our proposed GW on large-scale ImageNet classification~\cite{2015_IJCV_ImageNet}, as well as COCO object detection and segmentation~\cite{2014_ECCV_COCO}. We use the more efficient and numerically stable `ItN' (with $T=5$)~\cite{2019_CVPR_Huang} to calculate the whitening matrix for both GW and BW, in all experiments. Our implementation is based on PyTorch~\cite{2017_NIPS_pyTorch}.
% the Newton's iteration to approximate the Whitening operation, following \TODO{cite}.
\vspace{-0.13in}
\subsection{ImageNet Classification}
\label{subsec_imagenet}
\vspace{-0.03in}
We experiment on the ImageNet dataset with 1,000 classes ~\cite{2015_IJCV_ImageNet}.
We use the official 1.28M training images as a training set, and evaluate the top-1 accuracy  on a single-crop of 224x224 pixels in the validation set with 50k images.
% We adopt the data augmentation   the \cite{2016_TorchImageNet}
We investigate the ResNet~\cite{2015_CVPR_He} and ResNeXt~\cite{2017_CVPR_Xie} models.

\vspace{-0.05in}
\subsubsection{Ablation Study on ResNet-50}
\vspace{-0.05in}
\label{subsec_imagenet_ablation}
%\paragraph{Experimental setup.}
We follow the same experimental setup as described in ~\cite{2015_CVPR_He}, except that we use two GPUs and train over 100 epochs. We apply SGD with a mini-batch size of 256, momentum of 0.9 and weight decay of 0.0001. The initial learning rate is set to 0.1 and divided by 10 at 30, 60  and 90 epochs. Our baseline is the 50-layer ResNet (ResNet-50) trained with BN~\cite{2015_ICML_Ioffe}.
\vspace{-0.12in}
\paragraph{Effects of group number.}
We investigate the effects of group number for GW/GN, which we use to replace the BN of ResNet-50. We vary the group number $g$ ranging in $\{8, 16, 32, 64, 128\}$ (we use the channel number if it is less than the group number in a given layer), and report the training and validation accuracies in Figure~\ref{fig:exp_groupsize}. We can see that GW has consistent improvement over GN in training accuracy, across all values of $g$, which indicates the advantage of the whitening operation over standardization in terms of  optimization. Besides, GW also has better validation accuracy than GN.  We believe this may be because the stronger constraints of GW contribute to generalization.
We also observe that both GN and GW have significantly reduced training accuracy when the group number is too large (\eg, g=128), which is consistent with the previous results in Figure~\ref{fig:MLP_RandomLabel}.

\vspace{-0.12in}
\paragraph{Positions of GW.}
Although GW ($g$=64) provides slight improvement over the BN baseline ($76.32\%$ \textit{vs.} $76.23 \%$), it has a $90\%$ additional time cost\footnote{Note that our implementations are based on the APIs provided by PyTorch and are not finely optimized. For more discussion on time costs, please refer to the \TODO{\SM}~\ref{sup:exp-Time}.} on ResNet-50. Based on the analysis in Section~\ref{sec_theory}, it is reasonable to only partially replace BN with GW in networks, because 1) normalization within a batch or a group of channels both have their advantages in improving the optimization and generalization; 2) whitening can achieve better optimization efficiency and generalization than standardization~\cite{2018_CVPR_Huang}, but at a higher computational cost~\cite{2018_CVPR_Huang,2019_CVPR_Huang,2019_ICLR_Siarohin}.

Here, we investigate the position at which to apply GW ($g$=64) in ResNet-50. ResNet and ResNeXt are both composed primarily of a stem layer and multiple bottleneck blocks~\cite{2015_CVPR_He}. We consider:  1) replacing the BN in the stem layer with GW (referred to as `S1'); and 2) replacing the $1^{st}$, $2^{nd}$, $1^{st}~\&~2^{nd}$, and $3^{rd}$ BNs in all the bottleneck blocks, which are referred to as `B1', `B2', `B12' and `B3', respectively. We investigate five architectures, \textbf{S1}, \textbf{S1-B1}, \textbf{S1-B2}, \textbf{S1-B3} and \textbf{S1-B12}, which have 1, 17, 17, 17 and 33 GW modules, respectively.
We also perform experiments using BW~\cite{2019_CVPR_Huang} and BW$_\Sigma$~\cite{2020_CVPR_Huang} (employing a covariance matrix to estimate the population statistics of BW) for contrast.

We report the results in Table~\ref{table:ImageNet-Res50-ablation}.
BW/BW$_\Sigma$ improve their BN counterparts on all architectures by a clear margin, which demonstrates the advantage of the whitening operation over standardization~\cite{2019_CVPR_Huang}.
GW provides significant improvements over BW/BW$_\Sigma$ on \textbf{S1-B1}, \textbf{S1-B2}, \textbf{S1-B3} and \textbf{S1-B12} (an absolute improvement of $0.9\%$ on average). We attribute this to the advantage of GW in avoiding the estimation of population statistics.
%, while BW (BW$_\Sigma$) requires to estimate the population statistics of whitening matrix (covariance matrix) whose parameters are quadratic to the number of channels.
%  suggests the GW's advantage in avoding the mini-batch estimation problem \footnote{We also compare the recently proposed more effective estimation methods in \TODO{\SM}, where GW still has significantly improvement over BW}. 3)
We also observe that GW  has a slightly worse performance on \textbf{S1-B12} than on \textbf{S1-B1}/\textbf{S1-B2}. We believe there is a trade-off between GW and BN, in terms of affecting the model's representational capacity, optimization efficiency and generalization.

We also investigate the effect of inserting a GW/BW layer after the last average pooling (before the last linear layer) to learn the decorrelated feature representations, as proposed in \cite{2019_CVPR_Huang}.  This can slightly improve the performance ($0.10\%$ on average) when using GW, though the net gain is smaller than using BW ($0.22\%$) or BW$_\Sigma$ ($0.43\%$). Please refer to the \TODO{\SM}~\ref{sup:exp-DF} for details.
%The results are show in Figure ~\ref{fig:exp_position} (b).

\begin{table*}[t]
	\centering
	\vspace{-0.12in}
	%\begin{footnotesize}
	\begin{tabular}{c|lll|lll}
		\bottomrule[1pt]
		& \multicolumn{3}{c| }{2fc head box} &    \multicolumn{3}{c}{4conv1fc head box}  \\
		\hline
		Method     & AP$^{bbox}$   & AP$_{50}^{bbox}$ & AP$_{75}^{bbox}$& AP$^{bbox}$   & AP$_{50}^{bbox}$ & AP$_{75}^{bbox}$ \\
		\hline
		%BN*   &  36.90 & 58.60 & 40 &  36.85 & 57.73 & 39.93  \\
		$BN^{\dag}$  &36.31 &58.39 &38.83  &36.39 &57.22 &39.56 \\
		GN  &  36.62$_{(\textcolor{blue}{\uparrow 0.31})}$ & 58.91$_{(\textcolor{blue}{\uparrow 0.52})}$ & 39.32$_{(\textcolor{blue}{\uparrow 0.49})}$  &  37.86$_{(\textcolor{blue}{\uparrow 1.47})}$ & 58.96$_{(\textcolor{blue}{\uparrow 1.74})}$ & 40.76$_{(\textcolor{blue}{\uparrow 1.20})}$ \\
		GW   & \textbf{38.13}$_{(\textcolor{blue}{\uparrow 1.82})}$ & \textbf{60.63}$_{(\textcolor{blue}{\uparrow 2.24})}$ & \textbf{41.08}$_{(\textcolor{blue}{\uparrow 2.25})}$ &  \textbf{39.60}$_{(\textcolor{blue}{\uparrow 3.21})}$ & \textbf{61.12}$_{(\textcolor{blue}{\uparrow 3.90})}$ & \textbf{43.25}$_{(\textcolor{blue}{\uparrow 3.69})}$ \\
		\toprule[1pt]
	\end{tabular}
	\caption{Detection results ($\%$) on COCO using the Faster R-CNN framework implemented in~\cite{massa2018mrcnn}. We use ResNet-50 as the backbone, combined with FPN. All models are trained by 1x lr scheduling (90k iterations), with a batch size of 16 on eight GPUs.}
	\vspace{-0.05in}
	\label{table:Faster-FC}
\end{table*}

\begin{table*}[t]
	\centering
	%\begin{footnotesize}
	\begin{tabular}{c|lll|lll}
		\bottomrule[1pt]
		Method     & AP$^{bbox}$   & AP$^{bbox}_{50}$ & AP$^{bbox}_{75}$ & AP$^{mask}$ & AP$^{mask}_{50}$ & AP$^{mask}_{75}$\\
		\hline
		$BN^{\dag}$  &  42.24 & 63.00 & 46.19 & 37.53 & 59.82 & 39.96  \\
		%	BN  &  --  & -- & -- & -- & -- & --  \\
		GN  &  42.18$_{(\textcolor{red}{\downarrow 0.06})}$ & 63.22$_{(\textcolor{blue}{\uparrow 0.22})}$ & 46.00$_{(\textcolor{red}{\downarrow 0.19})}$ & 37.54$_{(\textcolor{blue}{\uparrow 0.01})}$ & 60.18$_{(\textcolor{blue}{\uparrow 0.36})}$ & 39.99$_{(\textcolor{blue}{\uparrow 0.03})}$  \\	
		GW  &  \textbf{44.41}$_{(\textcolor{blue}{\uparrow 2.17})}$ & \textbf{65.36}$_{(\textcolor{blue}{\uparrow 2.36})}$ & \textbf{48.67}$_{(\textcolor{blue}{\uparrow 2.48})}$ & \textbf{39.17}$_{(\textcolor{blue}{\uparrow 1.64})}$ & \textbf{62.13}$_{(\textcolor{blue}{\uparrow 2.31})}$ & \textbf{41.95}$_{(\textcolor{blue}{\uparrow 1.99})}$  \\
		%	GW  &  43.61 & 65.78 & 47.45 & 38.96 & 62.28 & 41.59  \\
		\toprule[1pt]
	\end{tabular}
		\caption{Detection and segmentation results ($\%$) on COCO using the Mask R-CNN framework implemented in~\cite{massa2018mrcnn}. We use ResNeXt-101 as the backbone, combined with FPN. All models are trained by 1x lr scheduling (180k iterations), with a batch size of 8 on eight GPUs.}
		\vspace{-0.1in}
		\label{table:Mask-X101-fc}
	%	\vspace{0.1in}
	%	\end{footnotesize}
	\vspace{-0.1in}
\end{table*}

\vspace{-0.08in}
\subsubsection{Validation on Larger Models}
\vspace{-0.05in}
\label{subsec_imagenet_larger}
In this section, we further validate the effectiveness of GW on ResNet-101~\cite{2015_CVPR_He}, ResNeXt-50 and ResNeXt-101~\cite{2017_CVPR_Xie}. We apply GW ($g$=64) in these models following the \textbf{S1-B2} architecture, which achieves the best performance (Table~\ref{table:ImageNet-Res50-ablation}) without significantly increasing the computational cost (it is only increased by roughly $23\%$). For comparison, we also apply BW$_\Sigma$ following the `S1-B2' architecture, combining the learning of decorrelated features~\cite{2020_CVPR_Huang} (BW$_\Sigma$ has a slightly improved performance compared to BW~\cite{2020_CVPR_Huang}).
%method~\footnote{Here, we use the more strong BW method  shown in~\cite{2020_CVPR_Huang}, where the population statistics are estimated by covariance matrix, which has improved performance than~\cite{2019_CVPR_Huang}} following `S1-B2' architecture.
Our baselines are the original networks trained with BN, and we also provide the results trained with GN.

The results are shown in Table~\ref{table:ImageNet-Res-Step}. We can see that 1) our method improves the baseline (BN) by a significant margin (between $1.02\%$ and $1.49\%$); and 2) BW$_\Sigma$ has consistently better performance than BN, but the net gain is reduced on wider networks (RexNeXt-50 and ResNeXt-101), which is probably caused by the difficulty in estimating the population statistics.
We also conduct experiments using more advanced training strategies (\eg,  cosine learning rate decay~\cite{2017_ICLR_Loshchilov}, label smoothing~\cite{2019_CVPR_He} and mixup~\cite{2018_ICLR_Zhang}) and GW again improves the baseline consistently. Please refer to the \TODO{\SM}~\ref{sup:exp-andvanceTraining} for details.
%3) GN has worse perfomance than BN, especially on ResNeXt models.

%\begin{table}[t]
%	\centering
%	\caption{Comparison of validation accuracy ($\%$, single model and
%		single-crop) ImageNet. We use the default cosine learning rate schedule.}
%	\label{table:ImageNet-Res-Cos}
%	\vspace{-0.1in}
%	%\begin{footnotesize}
%	\begin{tabular}{c|llll}
%		\bottomrule[1pt]
%		Method     & ResNet-50~\cite{2015_CVPR_He}   &  ResNet-101~\cite{2015_CVPR_He} &  ResNeXt-50 &  ResNeXt-101\\
%		\hline
%		Baseline (BN) \cite{2015_CVPR_He} & ~~76.62 &  ~~78.48 & ~~77.89 & ~~79.84  \\
%		GN   & ~~76.13 $_{(\textcolor{blue}{\downarrow 0.49})}$ & ~~77.51 $_{(\textcolor{blue}{\downarrow 0.97})}$
%		& ~~76.31 $_{(\textcolor{blue}{\downarrow 1.58})}$  &~~78.62  $_{(\textcolor{blue}{\downarrow 1.22})}$   \\	
%		ItN$_\Sigma$     & ~~77.92 $_{(\textcolor{blue}{\uparrow 1.30})}$ & ~~79.00 $_{(\textcolor{blue}{\uparrow 0.52})}$
%		&~~77.98 $_{(\textcolor{blue}{\uparrow 0.09})}$  & ~~79.96 $_{(\textcolor{blue}{\uparrow 0.12})}$ \\
%		GW (Ours)    & ~~\textbf{78.12} $_{(\textcolor{blue}{\uparrow 1.50})}$  & ~~\textbf{79.31} $_{(\textcolor{blue}{\uparrow 0.83})}$
%		&~~\textbf{78.96} $_{(\textcolor{blue}{\uparrow 1.07})}$ &~~\textbf{80.81} $_{(\textcolor{blue}{\uparrow 0.97})}$  \\
%		\toprule[1pt]
%	\end{tabular}
%	%	\end{footnotesize}
%\end{table}

%\paragraph{deeper and wider:} performance on 101 layer resnet and resnext. For the Resnext we use the 101x8d.
%
%\paragraph{Cosine learning rate schedule:}

%Here we denote the corsponding whitneing algorothima as ZCA$_{\CM{}}^L$
\vspace{0.1in}
\subsection{Object Detection and Segmentation on COCO}
\label{sec_Coco}
%\vspace{-0.05in}
We fine-tune the models trained on ImageNet for object detection and segmentation on the COCO benchmark~\cite{2014_ECCV_COCO}. We experiment on the Faster R-CNN \cite{2015_NIPS_Ren} and Mask R-CNN \cite{2017_ICCV_He} frameworks using the publicly available codebase `maskrcnn-benchmark'~\cite{massa2018mrcnn}.
We train the models on the COCO $train2017$ set and evaluate on the COCO $val2017$ set. We report the standard COCO metrics of average precision (AP), AP$_{50}$, and AP$_{75}$ for bounding box detection (AP$^{bbox}$) and instance segmentation (AP$^{m}$)~\cite{2014_ECCV_COCO}.
%We use the  pre-trained models on ImageNets  shown in Table**, as the back-bone, combining the Feature Pyramid Network (FPN).
For BN, we use its frozen version (indicated by $BN^{\dag}$) when fine-tuning for object detection~\cite{2018_ECCV_Wu}.
%These computer vision tasks in general benefit from higher-resolution input, so the batch size tends to be small in common practice.
\vspace{-0.12in}
\paragraph{Results on Faster R-CNN.}
For the Faster R-CNN framework, we use the  ResNet-50 models pre-trained on ImageNet (Table~\ref{table:ImageNet-Res-Step}) as the backbones, combined with the feature pyramid network (FPN)~\cite{2017_CVPR_Lin}. We consider two setups: 1) we use the  box head consisting of two fully connected layers (`2fc') without a normalization layer, as proposed in~\cite{2017_CVPR_Lin}; 2) following~\cite{2018_ECCV_Wu}, we replace the `2fc' box head with `4conv1fc', which can better leverage GN, and apply GN/GW to the FPN and box head.
We use the default  hypeparameter configurations  from the training scripts provided by the codebase~\cite{massa2018mrcnn} for Faster R-CNN.
The results are reported in Table~\ref{table:Faster-FC}.
The GW pre-trained model improves  $BN^{\dag}$ and GN by $1.82\%$ and $1.51\%$  AP, respectively. By adding GW/GN to the FPN and  `4conv1fc' head box, GW improves $BN^{\dag}$ and GN by $3.21\%$ and $1.74\%$  AP, respectively.

\vspace{-0.12in}
\paragraph{Results on Mask R-CNN.} For the Mask R-CNN framework, we use the ResNeXt-101~\cite{2017_CVPR_Xie} models pre-trained  on  ImageNet (Table~\ref{table:ImageNet-Res-Step}) as the backbones, combined with FPN. We use the `4conv1fc' box head, and apply GN/GW to the FPN, box head and mask head. We again use the default hypeparameter configurations from the training scripts provided by the codebase for Mask R-CNN~\cite{massa2018mrcnn}.
The results are shown in Table~\ref{table:Mask-X101-fc}. GW achieves $44.41\%$ in box AP and $39.17\%$ in mask AP, an improvement over $BN^{\dag}$  of $2.17\%$ and $1.64\%$, respectively.
%, \eg, we train the model for $180k$, with batch size of 8 on 8 GPUs
%Mainly benefits finetuning.
%
%\TODO{We also provides the experiments on ResNet-50 like Faster R-CNN in the \SM}
\vspace{-0.12in}
\paragraph{Small-batch-size training of BNs.}
Here, we further show that the network mixed with BNs and GWs can still work well under small-batch-size scenarios.
%One concern is that the network mixed with BNs and GWs may still have the poor performance under small-batch-size scenarios, due to the exists of BNs.
%of a mixture of BNs and GWs sort of defeats the purpose of mitigating the poor performance of BNs under small
%batch size scenarios.
%\TODO{Training From Scratch}
As illustrated in ~\cite{2017_NIPS_Ioffe}, one main cause of BN's small-batch-size problem is the inaccurate estimation between  training and inference distributions, which is amplified for a network with increased BN layers (these inaccuracies are compounded with depth). We believe inserting GW (which ensures the same distribution between training and inference) between consecutive BN layers will `break' these compounding inaccuracies, thus relieving the small-batch-size problem of BNs in a network.
We train  Faster R-CNN from scratch and use normal BN that is not frozen. We follow the same setup as in the previous experiment (\eg, two images/GPUs). We find that using all BNs only obtains 25.10$\%$ AP, while 28.37$\%$ AP is achieved using our mixture of BNs and GWs (the \textbf{S1-B2} architecture). Note that using all GNs (GWs) obtains 28.19$\%$ (28.79$\%$) AP.  This experiment  further validates that our mixture of BNs and GWs may also help mitigate the small-batch-size problem of the BNs in a network.
%Show results: fine tune benefits (Setup 1), and training benefits(Setup 2)
%, where we train the model for $90k$, with batch size of 16 on 8 GPU

%\vspace{-0.06in}
%\section{Related Work}
%\label{sec_relatedWork}
%\vspace{-0.06in}
%%\paragraph{Analysis of Batch Normalization}
%Batch Normalization perform standadizaiton over mini-batch data, which means that the
%\paragraph{Extension of normalization area}
%\TODO{Over batch dimension, but for small batch size problems}
%

\vspace{-0.05in}
\section{Conclusion and Future Work}
\vspace{-0.05in}
In this paper, we proposed group whitening (GW), which combines the advantages of normalization within a group of channels and the whitening operation. The effectiveness of GW was validated on large-scale visual recognition tasks. Furthermore, we also  analyzed the feature constraints imposed by normalization methods, enabling further understanding of how the batch size (group number) affects the performance of batch (group) normalized networks from the perspective of representational capacity.
This analysis can provide theoretical guidance for  applying GW and other normalization methods in practice.
It would be interesting to  build a unified framework to further investigate the effects of normalization in representation, optimization and generalization, by combining the proposed constraint analysis with the previous conditioning analysis~\cite{1998_NN_Yann,2020_ECCV_Huang} and stochasticity analysis~\cite{2018_ICML_Azizpour,2020_CVPR_Huang}. Our GW has also the potentialities to be used as a basic module in the switchable normalization methods~\cite{2018_arxiv_luo,2019_ICCV_Pan,2020_CVPR_Zhang}  to improve their performance.
 %We hope our analysis will provide a new means of understanding the behaviors of normalization methods.

  %further contrast normalization along batch with other dimensions~\cite{2019_NIPS_Li}    investigate normalization along other dimensions (\eg, positional normalization~\cite{2019_NIPS_Li} and divisive normalization~\cite{2017_ICLR_Ren}) or other normalization operations (\eg, scaling only~\cite{2019_NIPS_Chiley,2019_NIPS_Zhang}), using our constraint analysis. We hope our analysis will provide a new means of understanding the behaviors of normalization methods.
%Further work:
%1. An unified investigation on normalizaiton's affects on the model's representational capacity, optimization and generalization;
% 2) rethink the normalization method in training the small networks (prune-techinqiees, and small). (eventually normlaiziaon affects the model's representational capacity)
%3) rethink the LN methods, why it has better performance by RMS.

{\small
\bibliographystyle{ieee_fullname}
\bibliography{2GW}
}
\appendix
\clearpage
\renewcommand{\thetable}{A\arabic{table}}
\setcounter{table}{0}

\renewcommand{\thefigure}{A\arabic{figure}}
\setcounter{figure}{0}

\input{supp_GW_arxiv}

\end{document}

%% file: supp_GW_arxiv.tex
\section{Algorithms}
\label{sup:algorithm}

\begin{algorithm}[t]
	\algsetup{linenosize=\footnotesize}
	\footnotesize
	\caption{The forward pass of group whitening.}
	\label{alg_forward}
	\begin{algorithmic}[1]
		\begin{small}
			\STATE \textbf{Input}: a input sample $ \mathbf{x} \in \mathbb{R}^{d} $.
            \STATE \textbf{Hyperparameters}: $ \epsilon$, group number $g$.
			\STATE \textbf{Output}: $ \hat{\mathbf{x}} \in \mathbb{R}^{d}$.
            \STATE	Group division: $ \mathbf{X}_G= \Pi(\mathbf{x}; g) \in \mathbb{R}^{g \times c}$.
		    \STATE	 $\mathbf{\mu} = \frac{1}{c} \mathbf{X}_{G} \mathbf{1}$.
			\STATE	 $\mathbf{X}_{C} = \mathbf{X}_{G}-\mathbf{\mu}  \mathbf{1}^T $.
			\STATE	 $\Sigma = \frac{1}{c}\mathbf{X}_{C} \mathbf{X}_{C}^T + \epsilon \mathbf{I}$.
			\STATE  Calculate whitening matrix: $\Sigma^{-\frac{1}{2}} = \psi^f(\Sigma)$.
			\STATE  $\widehat{\mathbf{X}}_G= \Sigma^{-\frac{1}{2}} \mathbf{X}_C$.
			\STATE  Inverse  group division: $\hat{\mathbf{x}}= \Pi^{-1}(\widehat{\mathbf{X}}_G) \in \mathbb{R}^{d}$.
			%			\STATE  update: $\mathbf{\mu}_{E} \leftarrow (1-\lambda) ~\mathbf{\mu}_{E} + \lambda ~\mathbf{\mu}$.
			%			
			%			\STATE update: $\Sigma^{-\frac{1}{2}}_E \leftarrow (1-\lambda) \Sigma^{-\frac{1}{2}}_{E} + \lambda \mathbf{D} \mathbf{U} $.
		\end{small}
	\end{algorithmic}
\end{algorithm}

\begin{algorithm}[t]
	\caption{The corresponding backward pass of Algorithm \ref{alg_forward}.}
	\label{alg_backprop}
	\begin{algorithmic}[1]
		\begin{small}
			\STATE \textbf{Input}: gradient of a sample: $ \D{\Nx{}} \in \mathbb{R}^{d}$, and
			auxiliary data from respective forward pass: (1) $\mathbf{X}_C$; (2) $\Sigma^{-\frac{1}{2}}$.
			\STATE \textbf{Output}: gradient with respect to the input: $ \D{\x{}} \in \mathbb{R}^{d} $.
            \STATE	Group division: $ \D{\NX{G}}= \Pi(\D{\Nx{}}; g) \in \mathbb{R}^{g \times c}$.
			\STATE  $\D{\Sigma^{-\frac{1}{2}}}  = \D{\NX{G}}  \mathbf{X}_C^T$.
			\STATE Calculate gradient with respect to the covariance matrix: $\D{\Sigma} = \psi^b(\D{\Sigma^{-\frac{1}{2}}})$.
			\STATE  $\mathbf{f}=\frac{1}{c} \D{\NX{G}}  \mathbf{1} $.
			\STATE  $\D{\X{G}} =  \Sigma^{-\frac{1}{2}} (\D{\NX{G}}-\mathbf{f} \mathbf{1}^T ) +  \frac{1}{c}  (\D{\Sigma} + \D{\Sigma}^T) \mathbf{X}_C $.
             \STATE  Inverse  group division: $\D{\x{}}= \Pi^{-1}(\D{\X{G}}) \in \mathbb{R}^{d}$.
		\end{small}
	\end{algorithmic}
\end{algorithm}

\begin{figure*}[t]
	\centering
	\vspace{-0.23in}
	\begin{minipage}[c]{.88\linewidth}
		\centering
		\includegraphics[width=12.8cm]{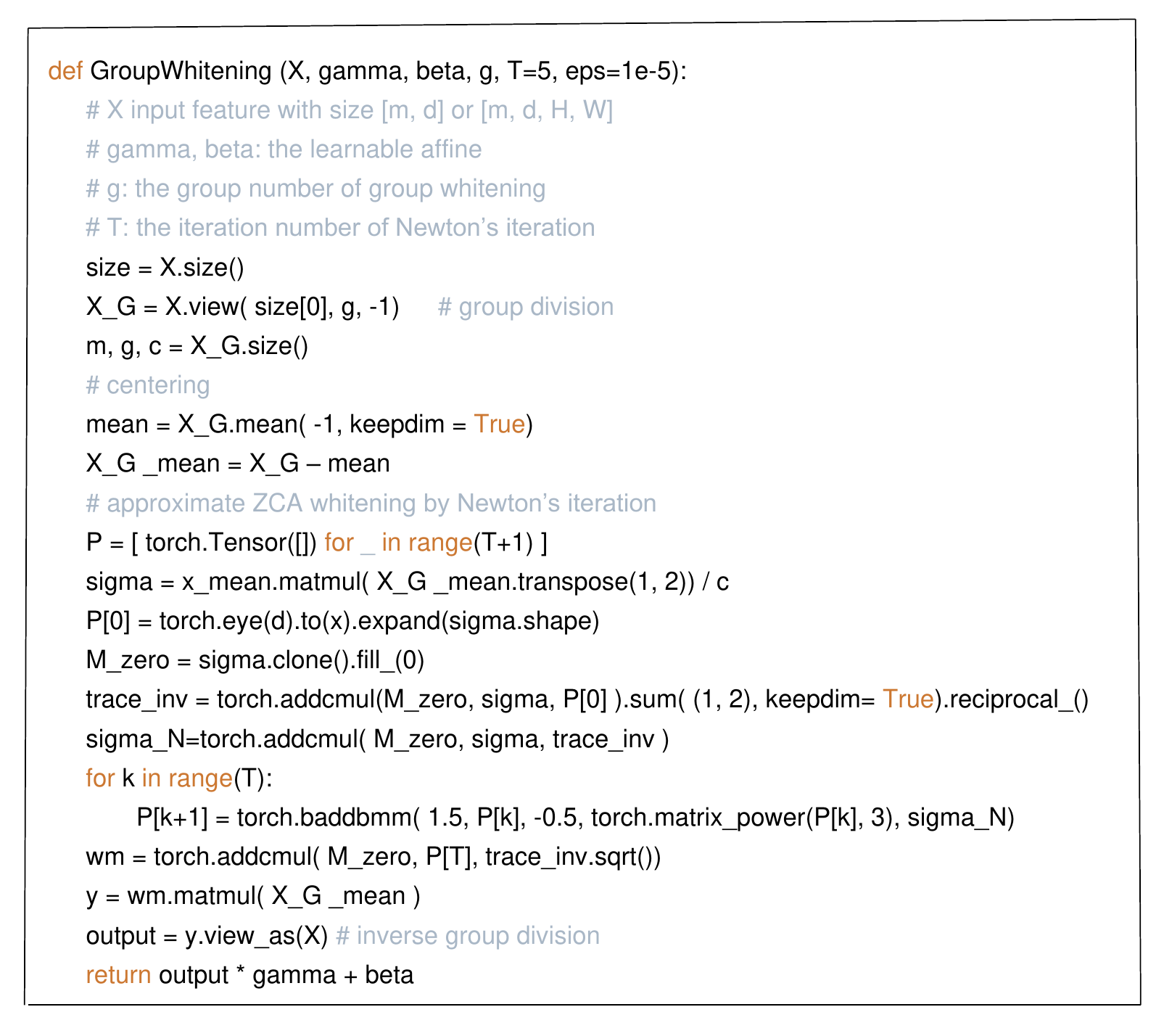}
	\end{minipage}
	\vspace{-0.08in}
	\caption{Python code of GW using ItN whitening, based on PyTorch. }
	\label{fig-sup:code_ItN}
	%\vspace{-0.17in}
\end{figure*}

The forward pass of the proposed group whitening (GW) method is shown in Algorithm~\ref{alg_forward}, and its corresponding backward pass is shown in Algorithm~\ref{alg_backprop}.\footnote{For GW, we also use the extra learnable dimension-wise scale and shift parameters, like BN~\cite{2015_ICML_Ioffe}. We omit this in the algorithms for simplicity.}
Note that  we need to specify the method for calculating the whitening matrix $\Sigma^{-\frac{1}{2}} \psi^f(\Sigma) $ in Line 8 of Algorithm~\ref{alg_forward}, as well as its backward operation $\D{\Sigma} = \psi^b(\D{\Sigma^{-\frac{1}{2}}})$ shown in Line 5 of Algorithm~\ref{alg_backprop}. As stated in the paper, we use zero-phase component analysis (ZCA)   whitening and its efficient approximation by Newton's iteration (`ItN')~\cite{2019_CVPR_Huang}. Here, we provide the details.

\paragraph{ZCA whitening.} ZCA whitening \cite{2018_CVPR_Huang}  calculates the whitening matrix by eigen decomposition as: $\Sigma^{-\frac{1}{2}} = \psi^f_{ZCA} (\Sigma) = \mathbf{D}\Lambda^{-\frac{1}{2}} \mathbf{D}^T $, where  $\Lambda=\mbox{diag}(\sigma_1, \ldots,\sigma_d)$ and $\mathbf{D}=[\mathbf{d}_1, ...,
\mathbf{d}_d]$ are the eigenvalues and associated eigenvectors of $\Sigma$, \ie $\Sigma = \mathbf{D}
\Lambda \mathbf{D}^T$.

The corresponding backward operation $\D{\Sigma} = \psi^b_{ZCA}(\D{\Sigma^{-\frac{1}{2}}})$ is as follows:
\begin{align}
\D{\Lambda}&=\mathbf{D}^T (\D{\Sigma^{-\frac{1}{2}}}) \mathbf{D} (-\frac{1}{2} \Lambda^{-3/2} )\\
\D{ \mathbf{D}}&=  (\D{\Sigma^{-\frac{1}{2}}}+  (\D{\Sigma^{-\frac{1}{2}}})^T)  \mathbf{D} \Lambda^{-1/2} \\
\D{\Sigma}&=\mathbf{D}\{ (\mathbf{K}^T \odot (\mathbf{D}^T \D{ \mathbf{D}} )) + (\D{\Lambda})_{diag}\} \mathbf{D}^T,
\end{align}
where $(\D{\Lambda})_{diag}$ sets the off-diagonal elements of
$\D{\Lambda}$ as zero.

\paragraph{`ItN' whitening.} `ItN' whitening~\cite{2019_CVPR_Huang} calculates the whitening matrix by Newton's iteration as: $\Sigma^{-\frac{1}{2}} = \psi^f_{ItN} (\Sigma) = \frac{\mathbf{P}_T}{\sqrt{tr(\Sigma_d)}}$, where $tr(\Sigma_d)$ indicates the trace of $\Sigma_d$ and $\mathbf{P}_T$ is calculated iteratively as:
  \begin{small}
  {\setlength\abovedisplayskip{3pt}
	\setlength\belowdisplayskip{3pt}
  	\begin{equation}
  	\label{eqn:Iteration}
  	\begin{cases}
  	\mathbf{P}_0=\mathbf{I} \\
  	\mathbf{P}_{k}=\frac{1}{2} (3 \mathbf{P}_{k-1} - \mathbf{P}_{k-1}^{3} \Sigma^{N}_{d}), ~~ k=1,2,...,T.
  	\end{cases}
  	\end{equation}
  }
  \end{small}
 \hspace{-0.05in}Here, $\Sigma^{N}_{d} = \Sigma_d / tr(\Sigma_d)$.

The corresponding backward operation $\D{\Sigma} = \psi^b_{ItN}(\D{\Sigma^{-\frac{1}{2}}})$ is as follows:
\begin{small}
	\begin{align}
	\label{eqn:backward-1}
	\D{\mathbf{P}_{T}}=& \frac{1}{\sqrt{tr(\Sigma)}} \D{\Sigma^{-\frac{1}{2}}} \nonumber \\
	\D{\Sigma_{N}}=& -\frac{1}{2} \sum_{k=1}^{T} (\mathbf{P}_{k-1}^3)^T \D{\mathbf{P}_k}  \nonumber  \\
	\D{\Sigma} =&  \frac{1}{tr(\Sigma)} \D{\Sigma_{N}}
	-\frac{1}{(tr(\Sigma))^2} tr(\D{\Sigma_{N}}^T \Sigma)   \mathbf{I}   \nonumber \\
	& ~~~-\frac{1}{2(tr(\Sigma))^{3/2}} tr((\D{\Sigma^{-\frac{1}{2}}})^T \mathbf{P}_T) \mathbf{I}.
	\end{align}
\end{small}
\hspace{-0.05in}Here, $\D{\mathbf{P}_k}$ can be calculated by the following iterations:
\begin{small}	
	\begin{align}
	\label{eqn:backward-iteration}
	\D{\mathbf{P}_{k-1}}& =\frac{3}{2} \frac{\partial{L}}{\partial{\mathbf{P}_{k}}}
	-\frac{1}{2} \D{\mathbf{P}_k}  (\mathbf{P}_{k-1}^2 \Sigma_{N})^T
	-\frac{1}{2}  (\mathbf{P}_{k-1}^2)^T  \D{\mathbf{P}_k} \Sigma_{N}^T
	\nonumber \\
	&-  \frac{1}{2}(\mathbf{P}_{k-1})^T \D{\mathbf{P}_k} (\mathbf{P}_{k-1} \Sigma_{N})^T
	,  ~~k=T,...,1.
	\end{align}
\end{small}
We also provide the python code of GW using ItN whitening, based on PyTorch~\cite{2017_NIPS_pyTorch}, in Figure~\ref{fig-sup:code_ItN}.

\section{More Results on Effects of Batch Size and Group Number}
\label{sup:batchSize}
\begin{figure*}[]
	\centering
	%\vspace{-0.2in}
  \hspace{-0.0in}	\subfigure[]{
		\begin{minipage}[c]{.23\linewidth}
			\centering
			\includegraphics[width=4.2cm]{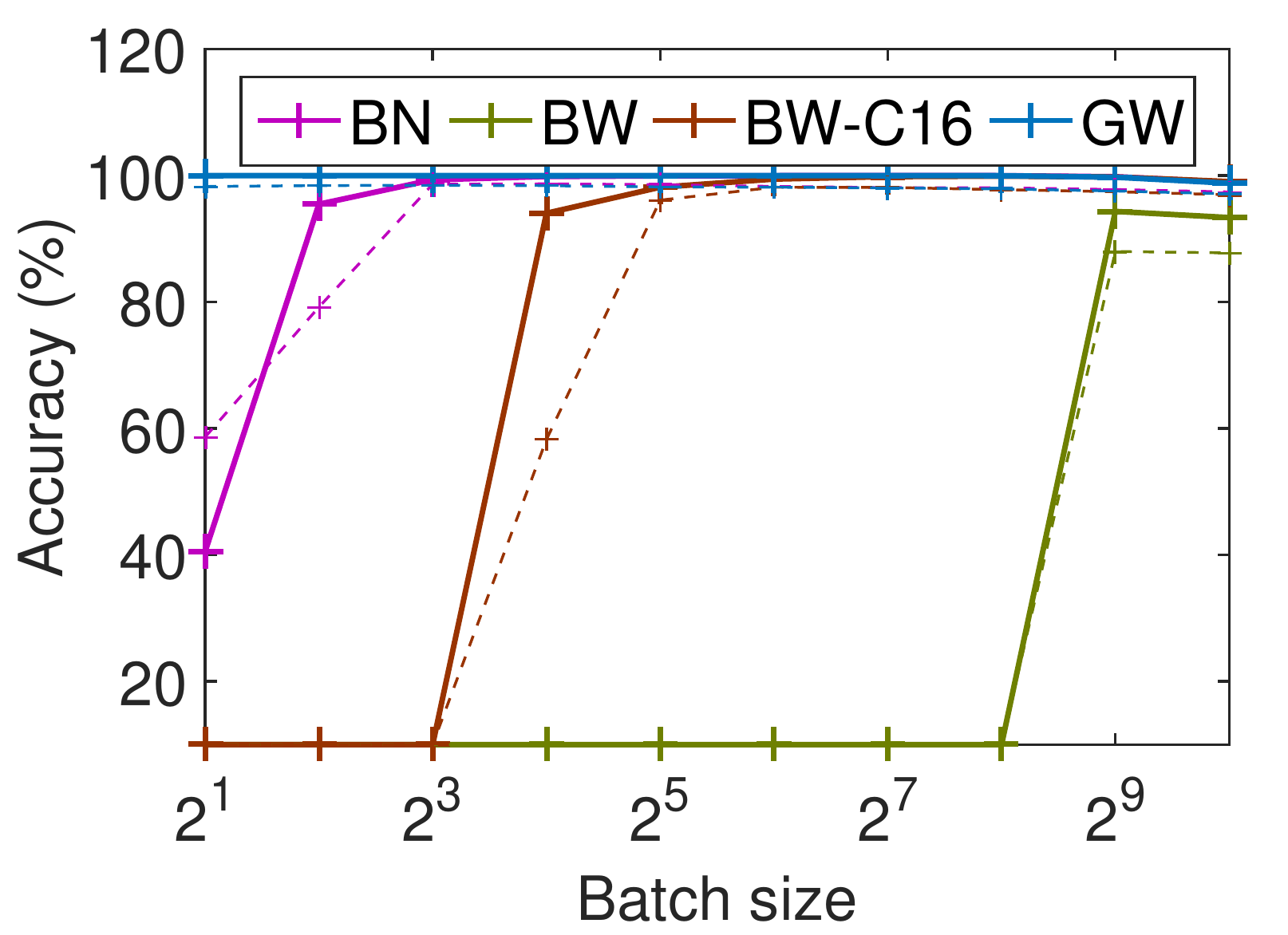}
		\end{minipage}
	}
	\hspace{0.0in}	\subfigure[]{
		\begin{minipage}[c]{.23\linewidth}
			\centering
			\includegraphics[width=4.2cm]{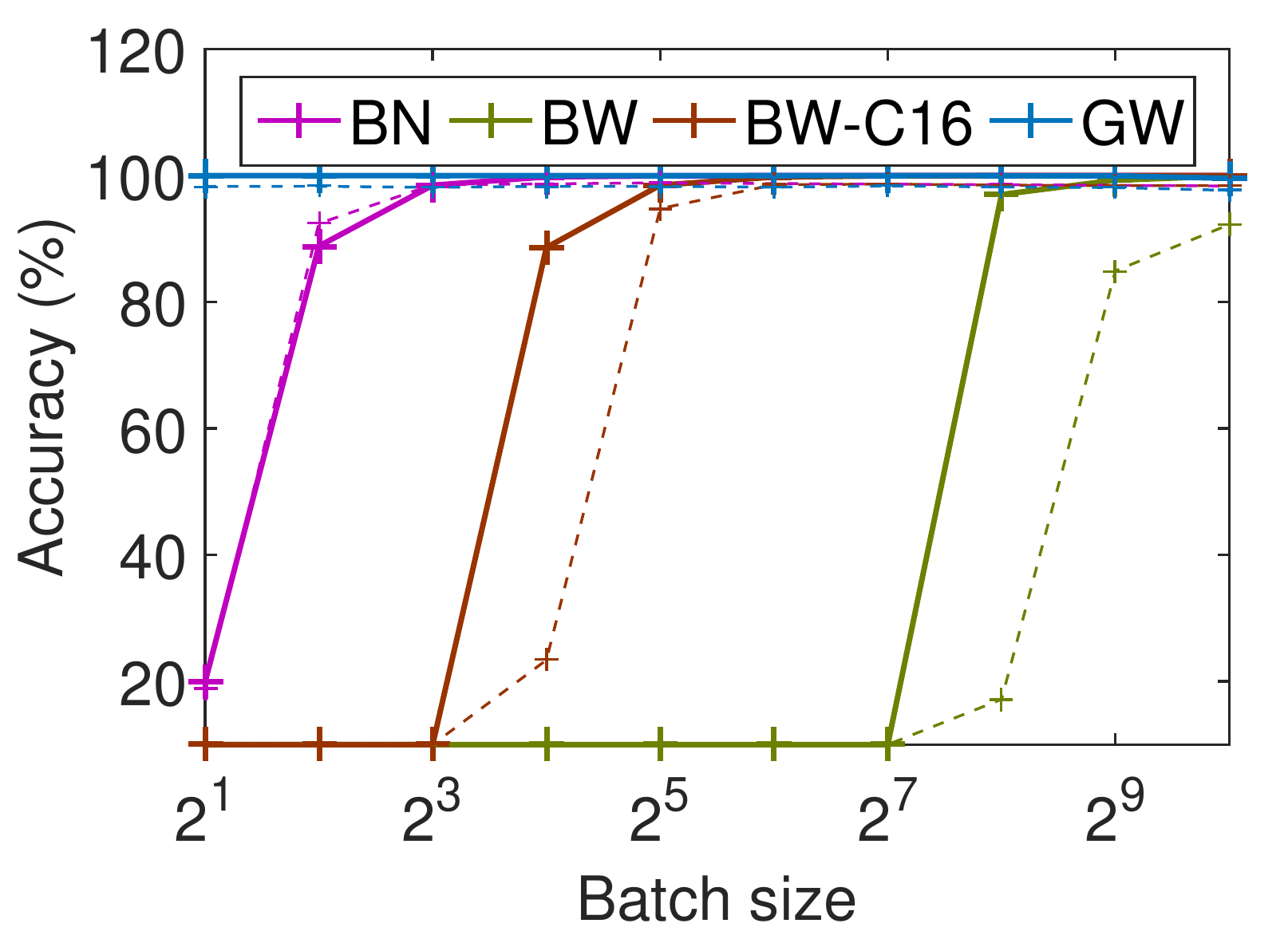}
		\end{minipage}
	}
  \hspace{0.0in}	\subfigure[]{
		\begin{minipage}[c]{.23\linewidth}
			\centering
			\includegraphics[width=4.2cm]{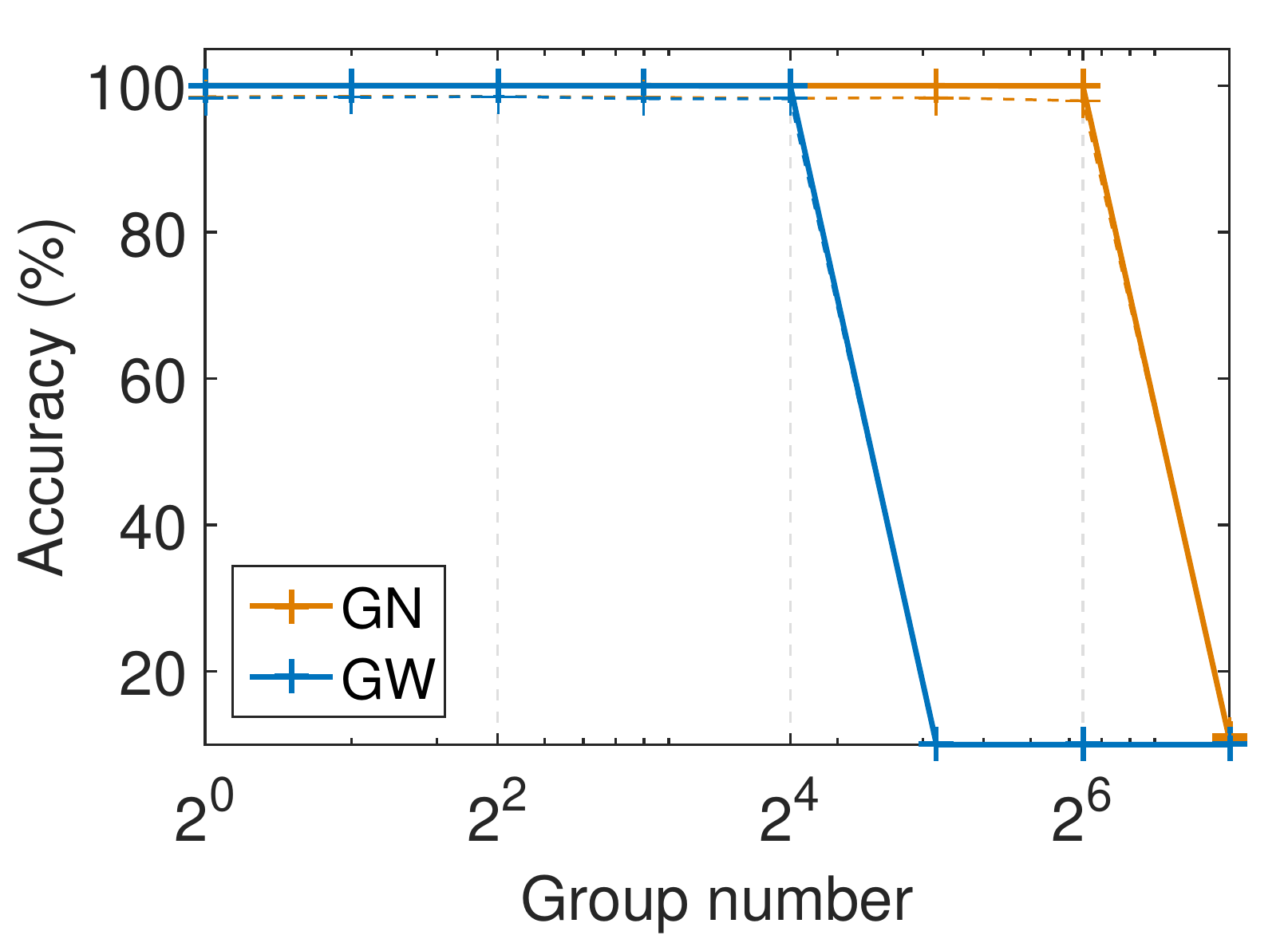}
		\end{minipage}
	}
	\hspace{0.0in}	\subfigure[]{
		\begin{minipage}[c]{.23\linewidth}
			\centering
			\includegraphics[width=4.2cm]{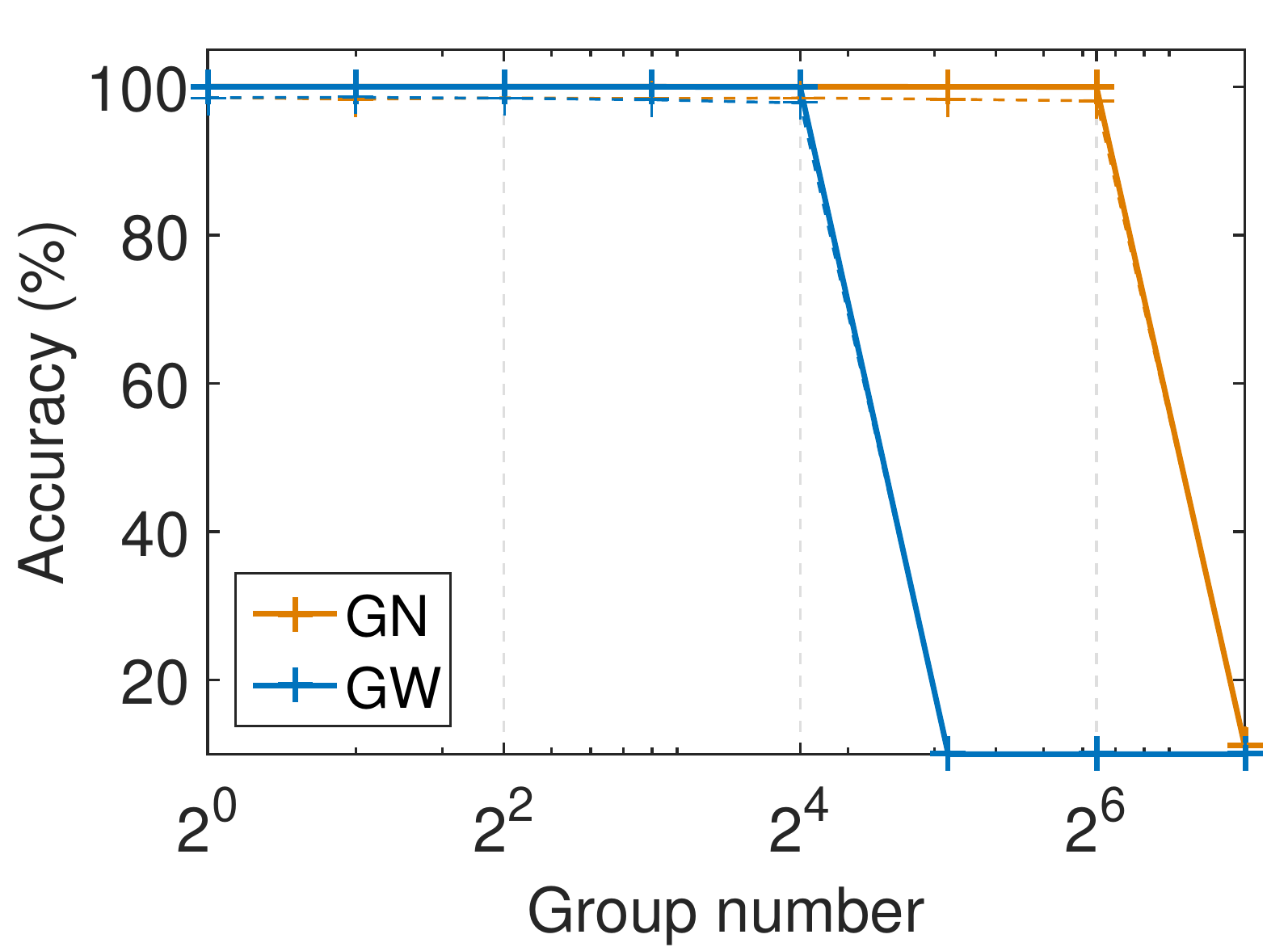}
		\end{minipage}
	}
	\vspace{-0.06in}
	\caption{Effects of batch size (group number) for batch (group) normalized networks. The experimental setup is the same as the one in Figure~\ref{fig:MLP_BatchSize} of the paper.  (a) Effect of batch size using a learning rate of 0.01; (b) Effect of batch size using a learning rate of 0.5; (c) Effect of group number using a learning rate of 0.01; (d) Effect of group number using a learning rate of 0.05;}
	\label{fig-sup:MLP_BatchSize}
	%\vspace{-0.12in}
\end{figure*}

 %The whitening operation has better advantage in improving the conditioning of optimization problem than standardization \TODO{cite},\TODO{SWitening}.
 In Figure~\ref{fig:MLP_BatchSize}  of the paper, we show the effects of  batch size (group number) for batch (group)
normalized networks, where the results are obtained with a learning rate of 0.1. Here, we provide more results using different learning rates, shown in Figure~\ref{fig-sup:MLP_BatchSize}. We  obtain similar observations.

\begin{figure*}[]
	\centering
	%\vspace{-0.2in}
  \hspace{-0.0in}	\subfigure[$\kappa_{90\%}$, one-layer MLP]{
		\begin{minipage}[c]{.23\linewidth}
			\centering
			\includegraphics[width=4.2cm]{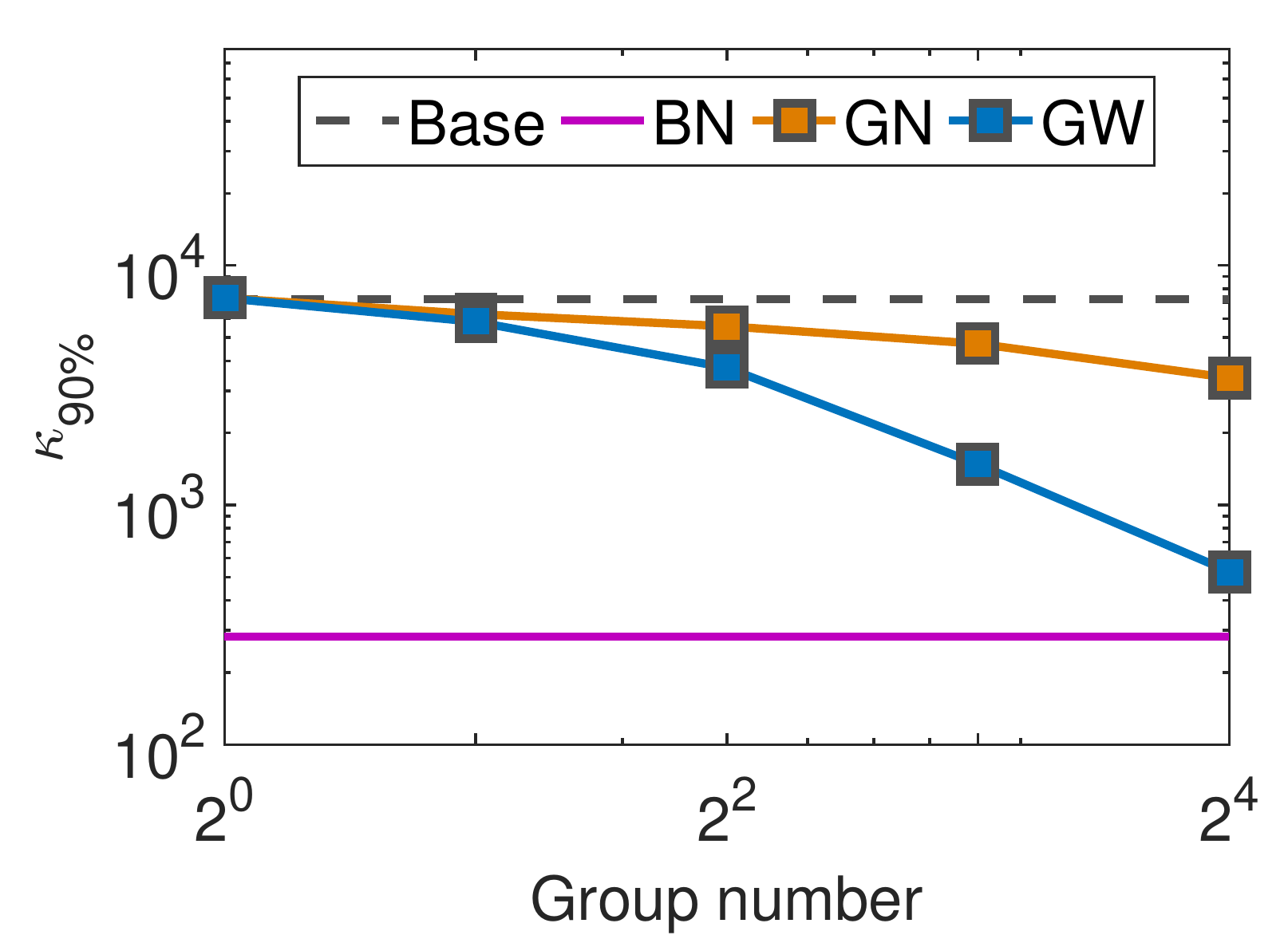}
		\end{minipage}
	}
	\hspace{0.0in}	\subfigure[$\kappa_{90\%}$, two-layer MLP]{
		\begin{minipage}[c]{.23\linewidth}
			\centering
			\includegraphics[width=4.2cm]{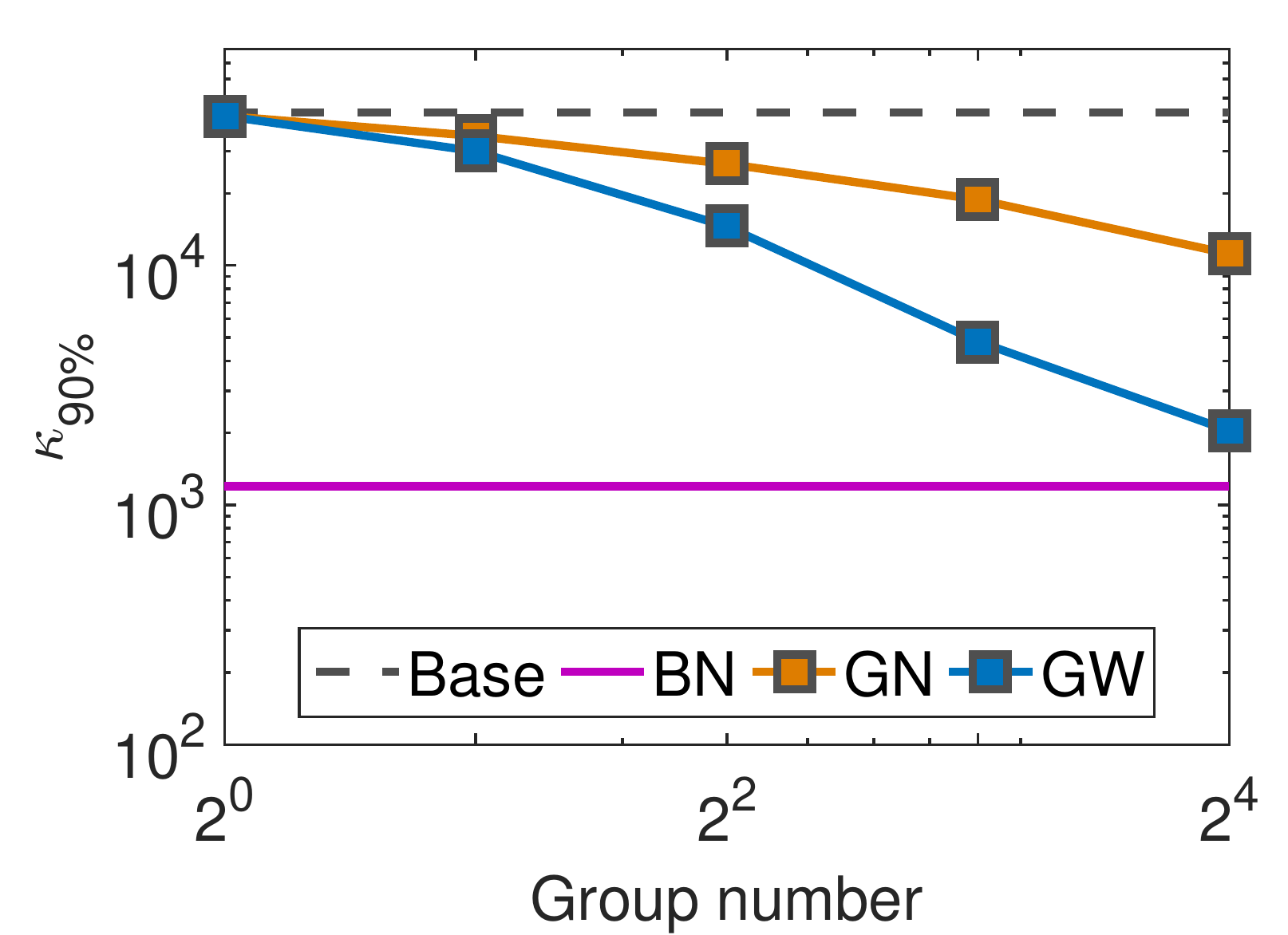}
		\end{minipage}
	}
  \hspace{0.0in}	\subfigure[$\kappa_{90\%}$, three-layer MLP]{
		\begin{minipage}[c]{.23\linewidth}
			\centering
			\includegraphics[width=4.2cm]{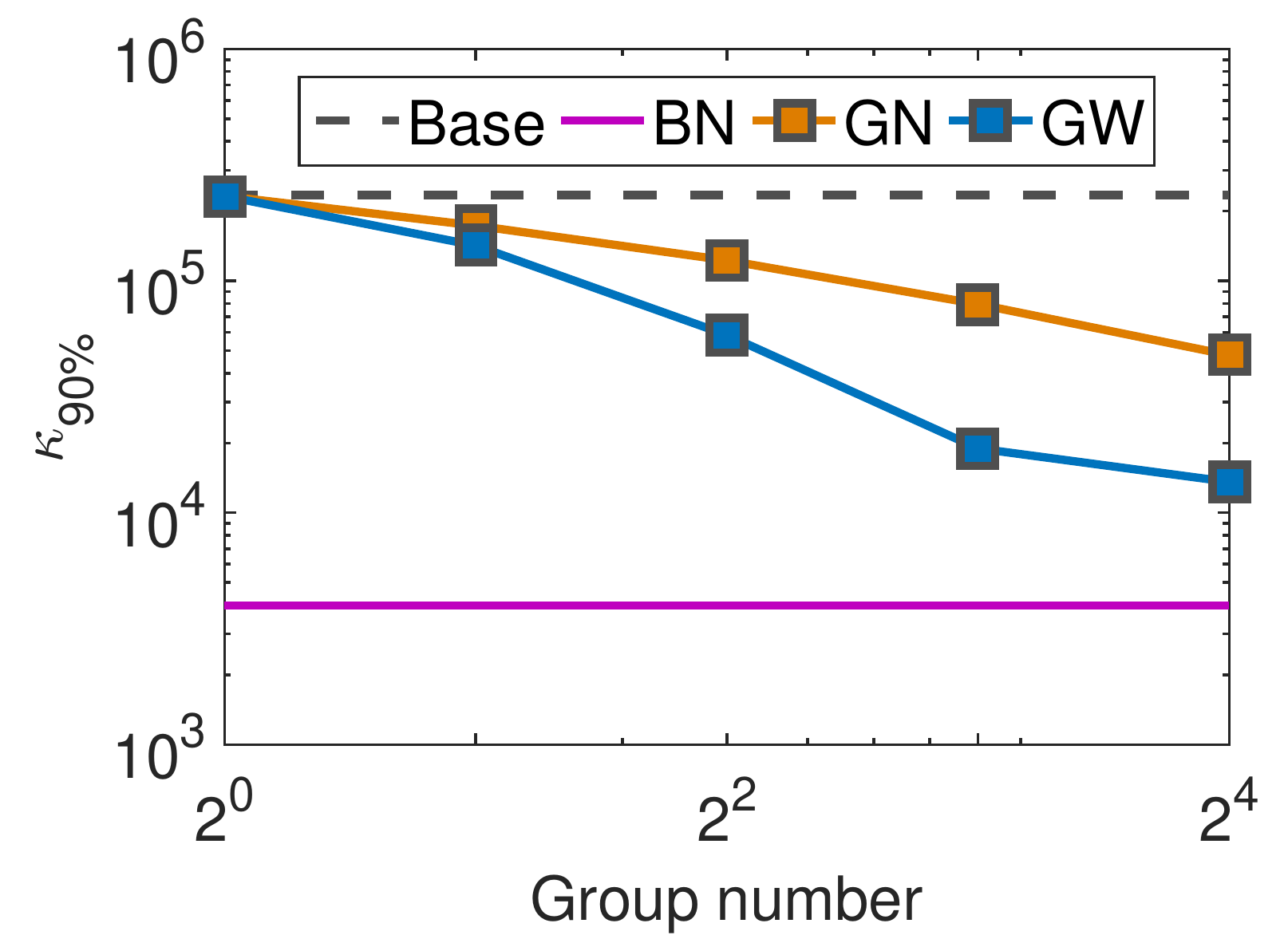}
		\end{minipage}
	}
	\hspace{0.0in}	\subfigure[$\kappa_{90\%}$, four-layer MLP]{
		\begin{minipage}[c]{.23\linewidth}
			\centering
			\includegraphics[width=4.2cm]{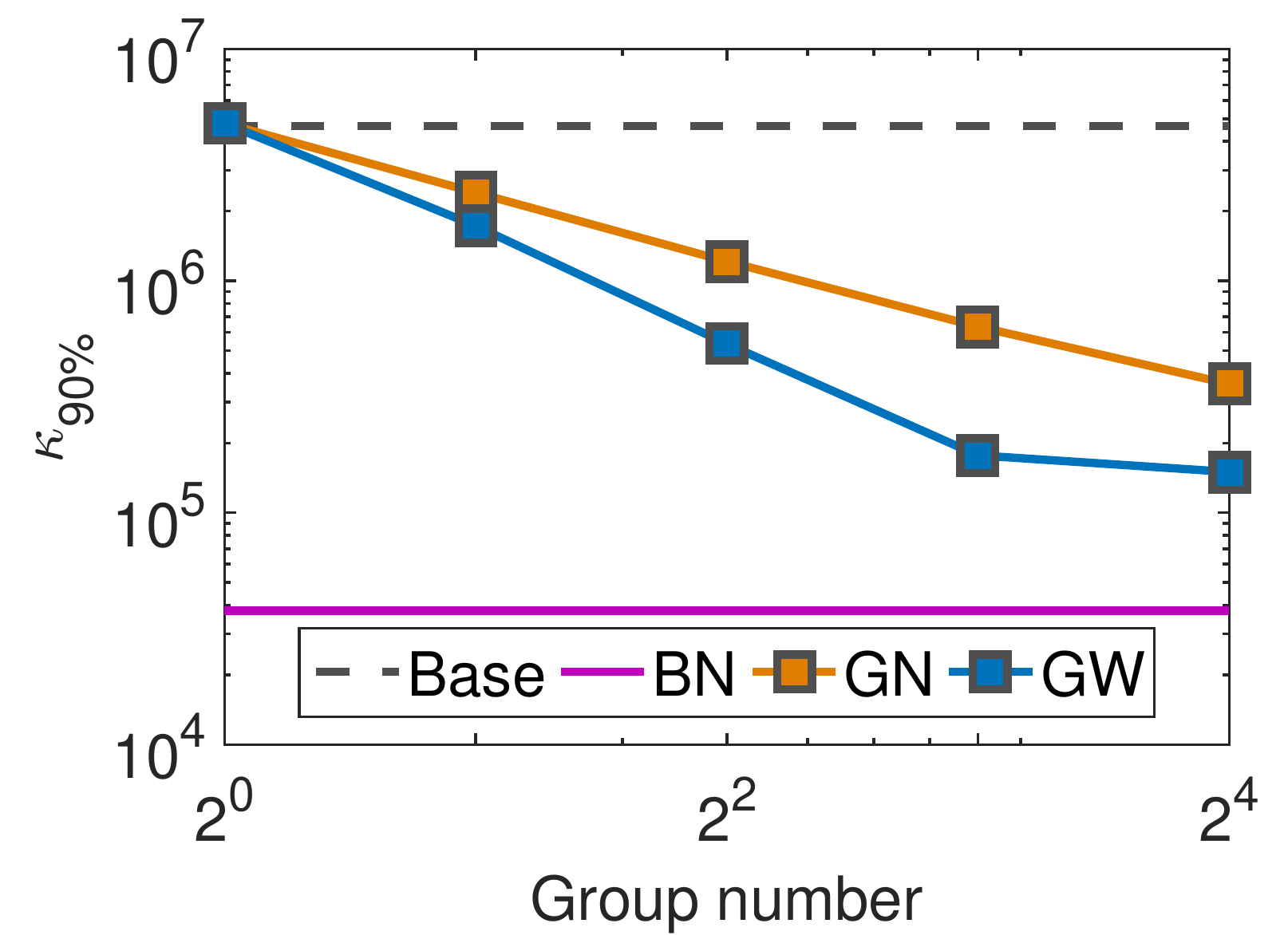}
		\end{minipage}
	}
\\
  \hspace{-0.0in}	\subfigure[$\kappa_{80\%}$, one-layer MLP]{
		\begin{minipage}[c]{.23\linewidth}
			\centering
			\includegraphics[width=4.2cm]{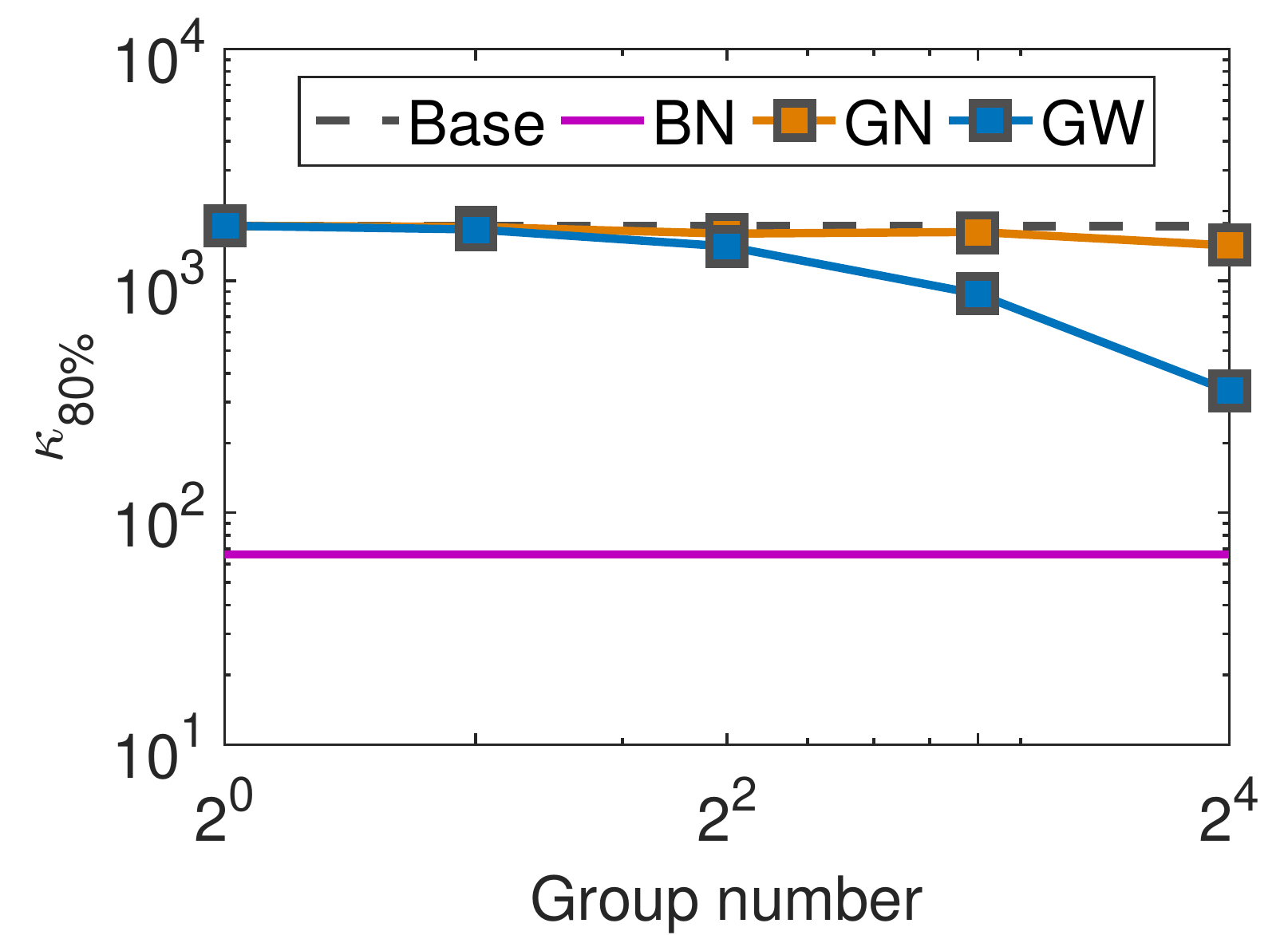}
		\end{minipage}
	}
	\hspace{0.0in}	\subfigure[$\kappa_{80\%}$, two-layer MLP]{
		\begin{minipage}[c]{.23\linewidth}
			\centering
			\includegraphics[width=4.2cm]{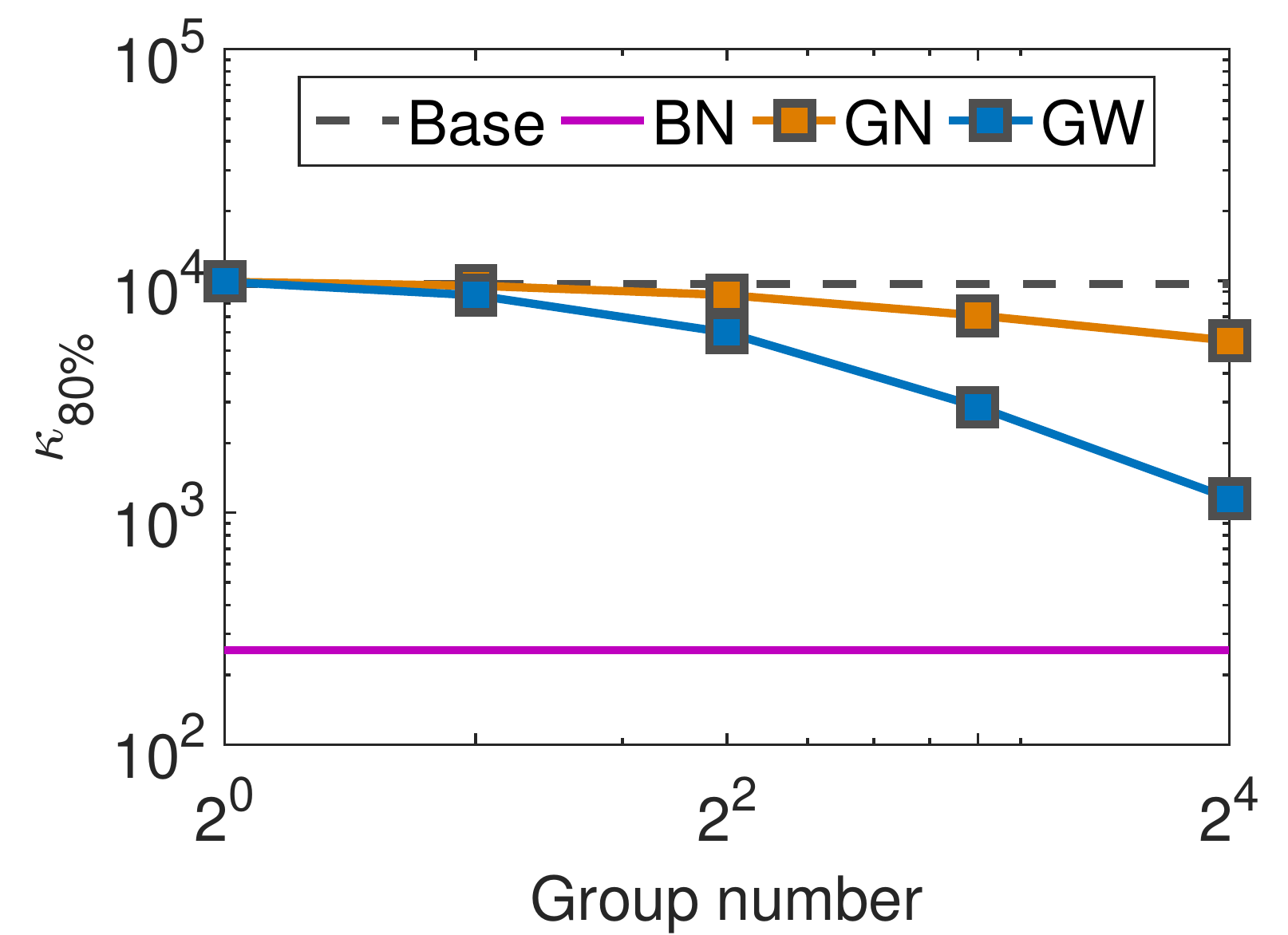}
		\end{minipage}
	}
  \hspace{0.0in}	\subfigure[$\kappa_{80\%}$, three-layer MLP]{
		\begin{minipage}[c]{.23\linewidth}
			\centering
			\includegraphics[width=4.2cm]{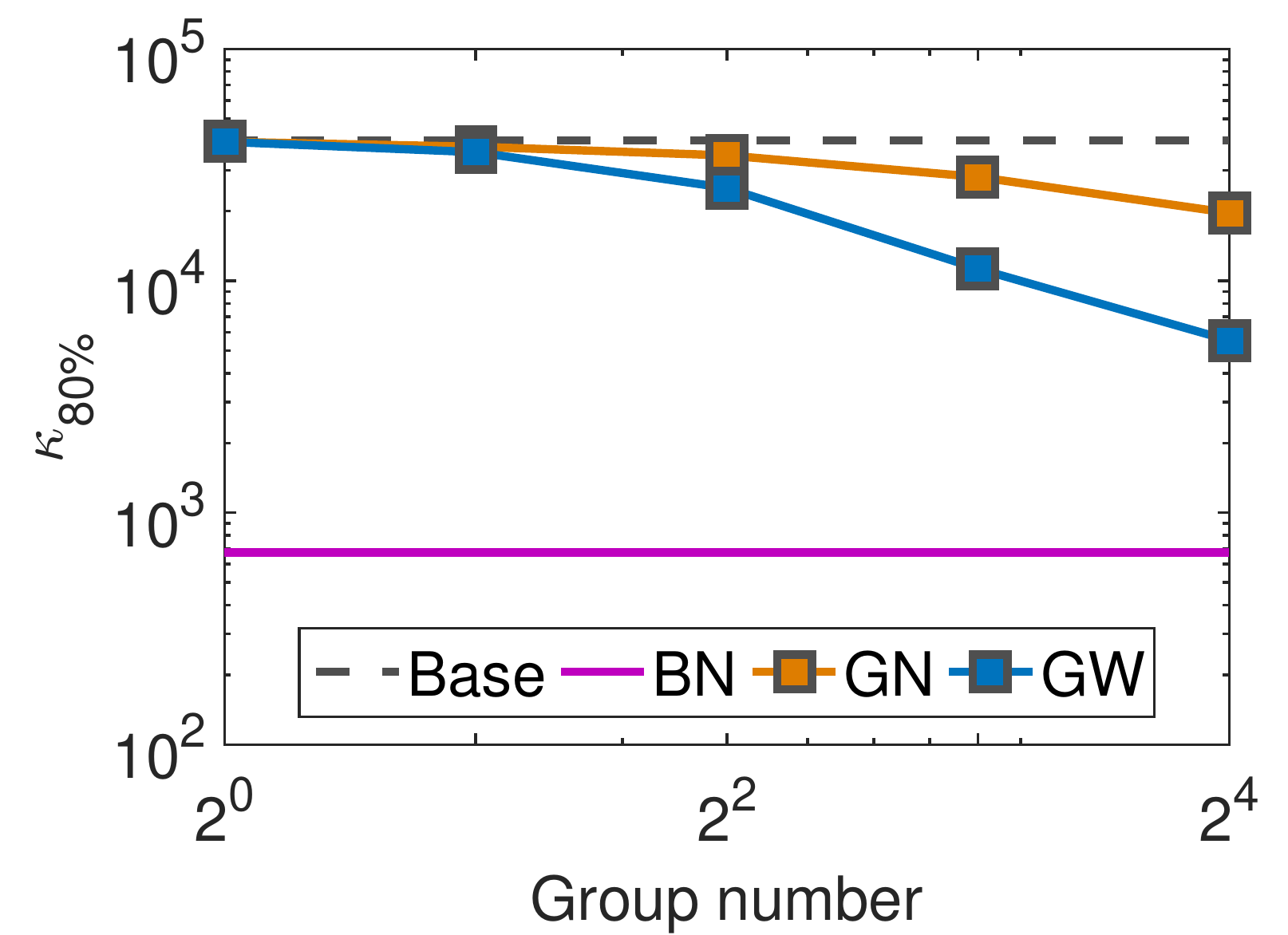}
		\end{minipage}
	}
	\hspace{0.0in}	\subfigure[$\kappa_{80\%}$, four-layer MLP]{
		\begin{minipage}[c]{.23\linewidth}
			\centering
			\includegraphics[width=4.2cm]{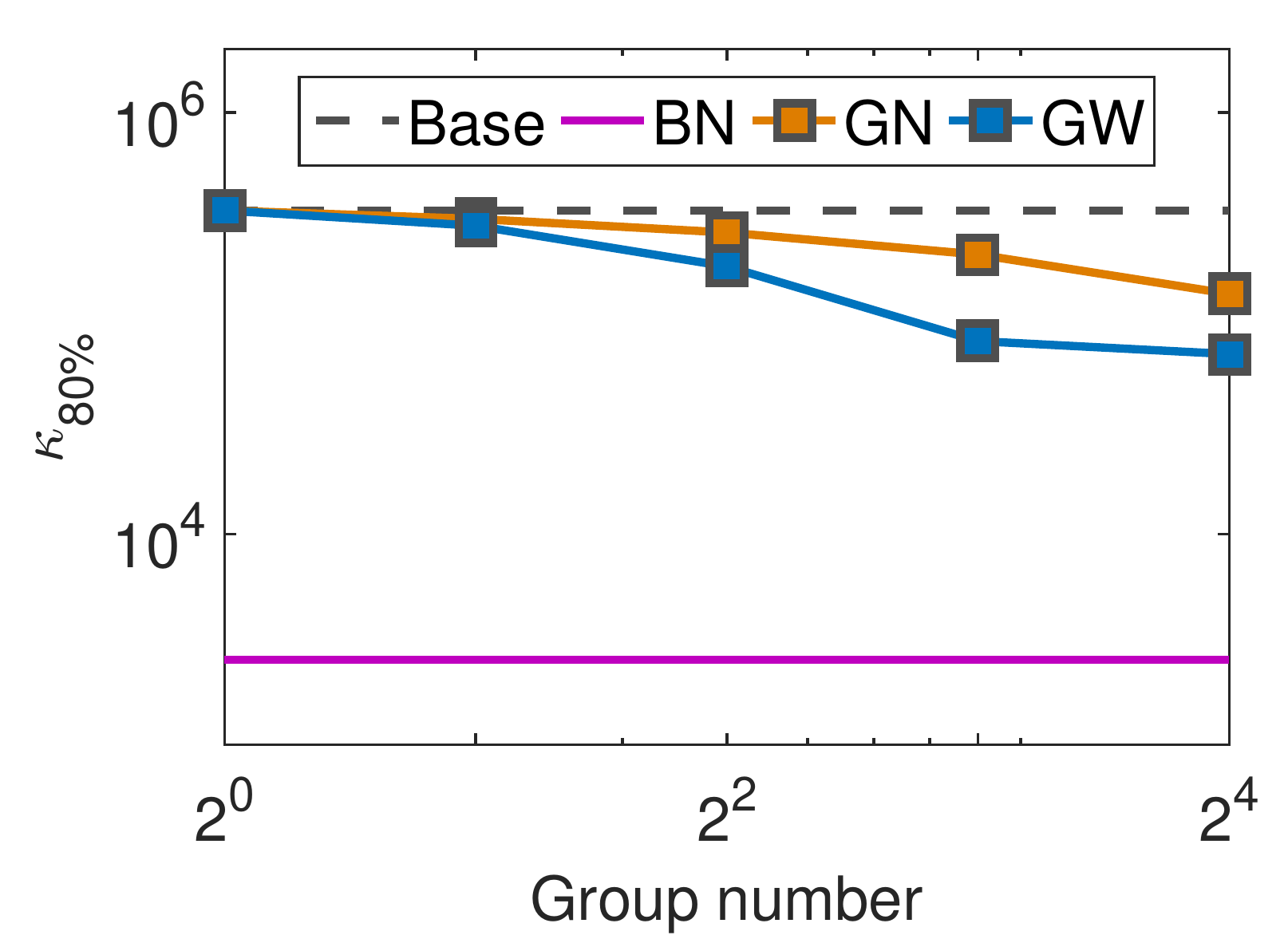}
		\end{minipage}
	}
	\vspace{-0.06in}
	\caption{Conditioning analysis on the normalized output. We simulate the activations $\mathbf{X}=f(\mathbf{X}_0) \in \mathbb{R}^{256 \times 1024}$ using a network $f(\cdot)$, where $\mathbf{X}_0$ is sampled from a Gaussian distribution.  We evaluate the more general condition number with respect to the percentage: $\kappa_{p}=\frac{\lambda_{max}}{\lambda_{p}}$, where $\lambda_{p}$ is the $pd$-th eigenvalue (in descending order) and $d$ is the total number of eigenvalues. We show the $\kappa_{90\%}$ of the covariance matrix of the output in (a)(b)(c)(d), and the $\kappa_{80\%}$ of the covariance matrix of the output  in (e)(f)(g)(h). We use a one-layer MLP as $f(\cdot)$ in (a)/(e); a two-layer MLP in (b)/(f); a three-layer MLP in (c)/(g); and a four-layer MLP in (d)/(h). }
	\label{fig-sup:conditioning}
	%\vspace{-0.12in}
\end{figure*}

\section{More Results on Conditioning Analysis}
\label{sup:conditioning}
In Figure~\ref{fig:conditioning}  of the paper, we perform a conditioning analysis on the normalized output, where we report $\kappa_{90\%}$ and use a one-layer and two-layer multilayer perceptron (MLP) as $f(\cdot)$ to obtain the activations. Here, we provide more results, shown in Figure ~\ref{fig-sup:conditioning}. We  obtain similar observations.

%\TODO{\section{More Figures of the representaion}}

\section{Derivation of Constraint Number of Normalization Methods}
\vspace{-0.1in}
\label{sup:constraint}
In Section~\ref{sec_theory} of the paper, we define the constraint number of a normalization operation, and summarize the constraint number of different normalization methods in Table~\ref{table:Constrains} of the paper.
Here, we provide the details for deriving the constraint number of batch whitening (BW), group normalization (GN)~\cite{2018_ECCV_Wu} and our proposed GW, for the mini-batch input $\mathbf{X} \in \mathbb{R}^{d \times m}$.

\paragraph{Constraint number of BW.} BW~\cite{2018_CVPR_Huang} ensures that the normalized output is centered and whitened, which has the constraints $\Upsilon_{\phi_{BW}}(\widehat{\mathbf{X}})$ as:
 {\setlength\abovedisplayskip{5pt}
	\setlength\belowdisplayskip{5pt}
	\begin{align}
	\label{eqn:constrain-BW-c}
    &\widehat{\mathbf{X}} \mathbf{1} = \mathbf{0}_{d}, ~~~~~and \\
	\label{eqn:constrain-BW-w}
    &\widehat{\mathbf{X}} \widehat{\mathbf{X}}^T  -  m \mathbf{I} = \mathbf{0}_{d \times d},
	\end{align}
}
\hspace{-0.05in}where $\mathbf{0}_{d}$ is a $d$-dimensional column vector of all zeros, and $\mathbf{0}_{d \times d}$ is a $d \times d$ matrix of all zeros. Note that there are $d$ independent equations in the system of equations $\widehat{\mathbf{X}} \mathbf{1} = \mathbf{0}_{d}$.
Let's denote $\mathbf{M} = \widehat{\mathbf{X}} \widehat{\mathbf{X}}^T  -  m \mathbf{I}$. We have $\mathbf{M}^T = \mathbf{M}$, and thus $\mathbf{M}$ is a symmetric matrix. Therefore, there are $d(d+1)/2$ independent equations in the system of equations $\widehat{\mathbf{X}} \widehat{\mathbf{X}}^T  -  m \mathbf{I} = \mathbf{0}_{d \times d}$. We thus have $d(d+1)/2+d$ independent equations in $\Upsilon_{\phi_{BW}}(\widehat{\mathbf{X}})$, and the constraint number of BW is $d(d+3)/2$.

\paragraph{Constraint number of GN.} Given a sample $\mathbf{x} \in \mathbb{R}^{d}$, GN divides the neurons into groups: $\mathbf{Z}=\Pi(\mathbf{x}) \in \mathbb{R}^{g \times c}$, where $g$ is the group number and $d=g c$. The standardization operation is then performed on $\mathbf{Z}$ as:
{\setlength\abovedisplayskip{3pt}
	\setlength\belowdisplayskip{3pt}
	\begin{equation}
	\label{eqn:GN}
\widehat{\mathbf{Z}}
= \Lambda_{g}^{-\frac{1}{2}} (\mathbf{Z} - \mathbf{\mu}_g  \mathbf{1}^T),
	\end{equation}
}
\hspace{-0.04in}where, $\mathbf{\mu}_g = \frac{1}{c} \mathbf{Z} \mathbf{1}$ and  $\Lambda_{g}=\mbox{diag}(\sigma_1^2, \ldots,\sigma_g^2) + \epsilon \mathbf{I}$
. This  ensures that the normalized output $\widehat{\mathbf{Z}}$ for each sample has the constraints:
{\setlength\abovedisplayskip{5pt}
	\setlength\belowdisplayskip{5pt}
	\begin{eqnarray}
	\label{eqn:constrain-GN}
     \sum_{j=1}^{c} \widehat{\mathbf{Z}}_{ij}=0 ~ and ~ \sum_{j=1}^c \widehat{\mathbf{Z}}_{ij}^2 =c, ~ for  ~i=1, ..., g.
	\end{eqnarray}
}
\hspace{-0.05in} In the system of equations~\ref{eqn:constrain-GN}, the number of independent equations is $2g$. Therefore, the constraint number of GN is $2dm$, when given $m$ samples.

 %Here, defining the group division operation as $\Pi: \mathbb{R}^{d \times m} \mapsto \mathbb{R}^{c \times gm}$, where $g$ is the group number and $d=g c$,
%{\setlength\abovedisplayskip{3pt}
%	\setlength\belowdisplayskip{3pt}
%	\begin{eqnarray}
%	\label{eqn:GN-1}
%     Group ~ division:& \mathbf{X}_G= \Pi(\mathbf{x}; g) \in \mathbb{R}^{g \times c}, \\
%     \label{eqn:GN}
%     Standardization:&\widehat{\mathbf{X}}_G = \phi_{LN}(\mathbf{X}_G) = \Sigma^{-\frac{1}{2}}_{g}(\mathbf{X}_G - \mathbf{\mu}_g  \mathbf{1}^T), \\
%     	\label{eqn:GN-2}
%     Inverse ~ group~ division:& \hat{\mathbf{x}}= \Pi^{-1}(\widehat{\mathbf{X}}_G) \in \mathbb{R}^{d},
%	\end{eqnarray}
%}

\paragraph{Constraint number of GW.}
 Given a sample $\mathbf{x} \in \mathbb{R}^{d}$,  GW performs normalization as:
{\setlength\abovedisplayskip{3pt}
	\setlength\belowdisplayskip{3pt}
	\begin{align}
	\label{eqn:GW-1}
    & Group ~ division: \mathbf{X}_G= \Pi(\mathbf{x}; g) \in \mathbb{R}^{g \times c}, \\
     \label{eqn:GW}
    & Whitening:\widehat{\mathbf{X}}_G = \Sigma^{-\frac{1}{2}}_{g}(\mathbf{X}_G - \mathbf{\mu}_g  \mathbf{1}^T), \\
     	\label{eqn:GW-2}
    &  Inverse ~ group~ division:\hat{\mathbf{x}}= \Pi^{-1}(\widehat{\mathbf{X}}_G) \in \mathbb{R}^{d}.
	\end{align}
}
\hspace{-0.05in}The normalization operation ensures that $\widehat{\mathbf{X}}_G \in \mathbb{R}^{g \times c}$ has the following constraints:
{\setlength\abovedisplayskip{5pt}
	\setlength\belowdisplayskip{5pt}
	\begin{align}
	\label{eqn:constrain-GW-c}
    &\widehat{\mathbf{X}}_G \mathbf{1} = \mathbf{0}, ~~~~~and \\
	\label{eqn:constrain-GW-w}
    &\widehat{\mathbf{X}}_G \widehat{\mathbf{X}}_G^T  -  c \mathbf{I} = \mathbf{0}.
	\end{align}
}
\hspace{-0.03in}Following the analysis for BW, the number of  independent equations is $g(g+3)/2$ from Eqns.~\ref{eqn:constrain-GW-c} and~\ref{eqn:constrain-GW-w}. Therefore, the  constraint number of GW is $mg(g+3)/2$,  when given $m$ samples.

\begin{table*}[t]
	\centering
	%\vspace{-0.15in}
	%\begin{footnotesize}
	\begin{tabular}{c|lllll}
		\bottomrule[1pt]
		& ~~S1   & S1-B1 &  S1-B2 &  S1-B3 & S1-B12\\
		\hline
		Baseline (BN) & ~~76.24 & ~~76.24 & ~~76.24 & ~~76.24& ~~76.24  \\
		BW~\cite{2019_CVPR_Huang}     & ~~76.91 $_{(\textcolor{blue}{\uparrow 0.67})}$ & ~~76.94 $_{(\textcolor{blue}{\uparrow 0.70})}$
		&~~76.93 $_{(\textcolor{blue}{\uparrow 0.69})}$  &~~76.78 $_{(\textcolor{blue}{\uparrow 0.54})}$ &~~76.79 $_{(\textcolor{blue}{\uparrow 0.55})}$\\
		BW$_\Sigma$~\cite{2020_CVPR_Huang}     & ~~\textbf{77.09} $_{(\textcolor{blue}{\uparrow 0.85})}$ & ~~77.04 $_{(\textcolor{blue}{\uparrow 0.80})}$
		&~~77.21 $_{(\textcolor{blue}{\uparrow 0.97})}$  &~~77.10 $_{(\textcolor{blue}{\uparrow 0.86})}$ &~~77.11 $_{(\textcolor{blue}{\uparrow 0.87})}$\\
		GW     & ~~76.86 $_{(\textcolor{blue}{\uparrow 0.62})}$  & ~~\textbf{77.63} $_{(\textcolor{blue}{\uparrow 1.39})}$
		&~~\textbf{77.80} $_{(\textcolor{blue}{\uparrow 1.56})}$ &~~\textbf{77.75} $_{(\textcolor{blue}{\uparrow 1.51})}$ &~~\textbf{77.48} $_{(\textcolor{blue}{\uparrow 1.24})}$ \\
		\toprule[1pt]
	\end{tabular}
	\caption{Effects of inserting a GW/BW/BW$_{\Sigma}$ layer after the last average pooling of ResNet-50 to learn decorrelated feature representations for ImageNet classification. We evaluate the top-1 validation accuracy on five architectures  (\textbf{S1}, \textbf{S1-B1}, \textbf{S1-B2}, \textbf{S1-B3} and \textbf{S1-B12}), described in the paper.  Note that we also use an extra BN layer after the last average pooling for the Baseline (BN). }
	\label{table-sup:ImageNet-Res50-ablation}
	%\vspace{-0.1in}
	%\vspace{-0.1in}
	%	\end{footnotesize}
\end{table*}

\section{Investigating Representational Capacity on CNNs}
\label{sub:representation_CNN}
In Section~\ref{sec:representationFeature} of the paper, we empirically show how normalization affects the representational capacity of a network with  experiments conducted on MLPs. Here, we  conduct experiments on convolutional neural networks
(CNNs) for CIFAR-10 classification.
%We investigate how the group number (batch size) affects the performance of group (batch) normalized networks.
Note that the number of neurons to be normalized for GN  is $dHW$, given the convolutional input $\mathbf{X} \in \mathbb{R}^{d \times m \times H \times W}$, where H and W are the height and width of the feature maps. The number of samples to be normalized for BN is $mHW$.
We use a CNN with $n$ convolutional layers, following average pooling and one fully connected layer. We use $d=16$ channels in each layer and vary the depth $n$. We apply stochastic gradient descent (SGD) with a momentum of 0.9. We train over 160 epochs and  divide the learning rate by 5 at 60  and 120 epochs. We evaluate the best training accuracy among the learning rates of $\{0.001, 0.01,0.05,0.1,0.5\}$.

\begin{figure}[t]
	\centering

			\centering
			\includegraphics[width=4.8cm]{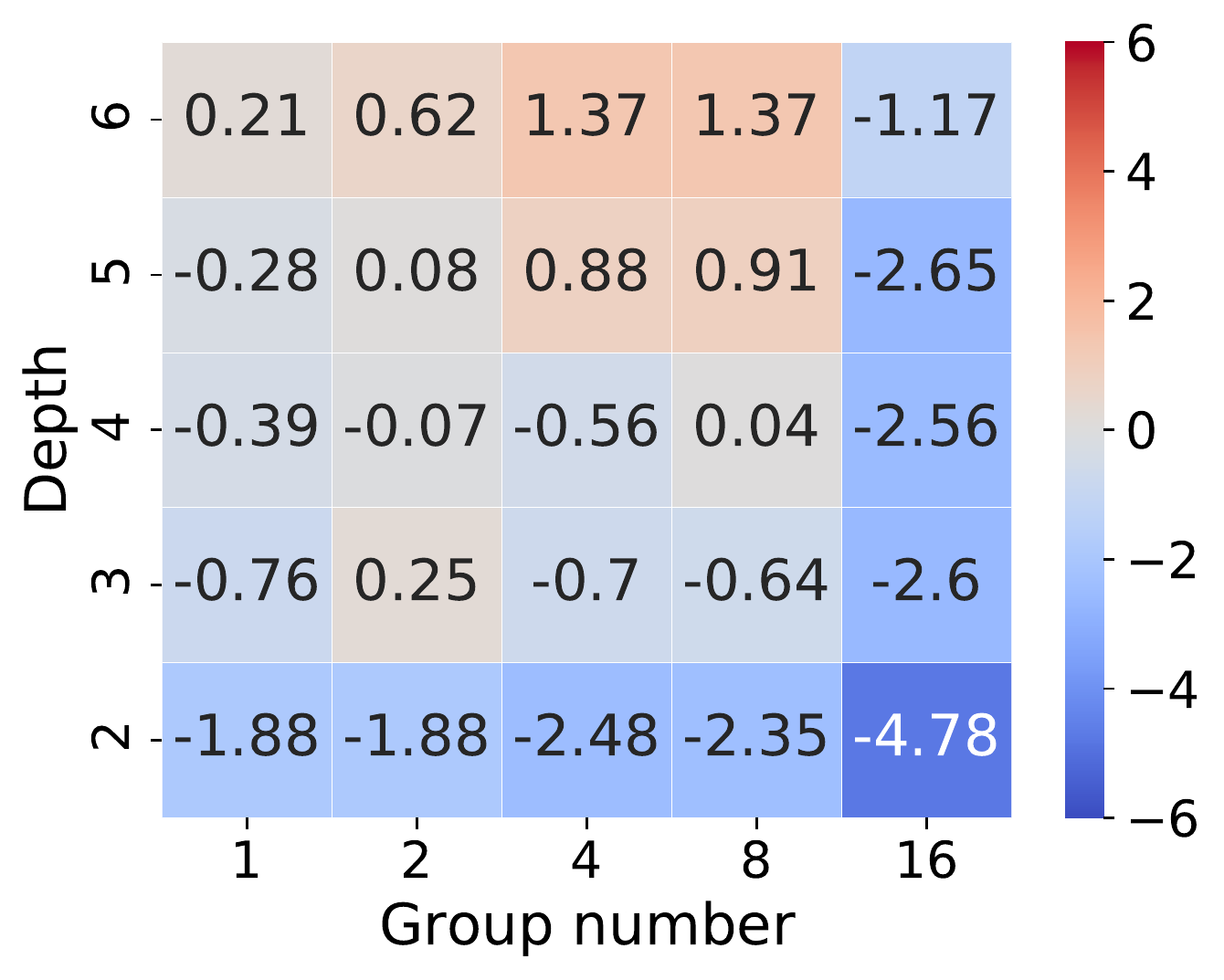}
%\hspace{0.15in}		\subfigure[Validation]{
%		\begin{minipage}[c]{.43\linewidth}
%			\centering
%			\includegraphics[width=4.2cm]{./figures/CNN/test_w16_DeltaGN.pdf}
%		\end{minipage}
%	}	
	\vspace{-0.05in}
	\caption{Effects of group number for GN on CNNs for CIFAR-10 classification. We vary the group number of GN, and evaluate the difference in training accuracy between GN and the model without normalization (`Base'). }
	\label{fig:sup_CNN_debug_GN}
	%\vspace{-0.16in}
\end{figure}

Figure~\ref{fig:sup_CNN_debug_GN} shows the results of GN with varying group number (we use a batch size of 256), where we report the difference in training accuracy between GN and the model without normalization (`Base'). We observe that: 1) GN has significantly degenerated performance when the group number is too large (relative to the channel number), \eg, GN has worse performance than `Base' when $g=d=16$; 2) The net gain of GN over `Base' is amplified as the depth increases. These observations are consistent with the experiments on the MLPs  shown in Section ~\ref{sec:representationFeature} of the paper.

\begin{figure}[t]
	\centering
			\includegraphics[width=6.5cm]{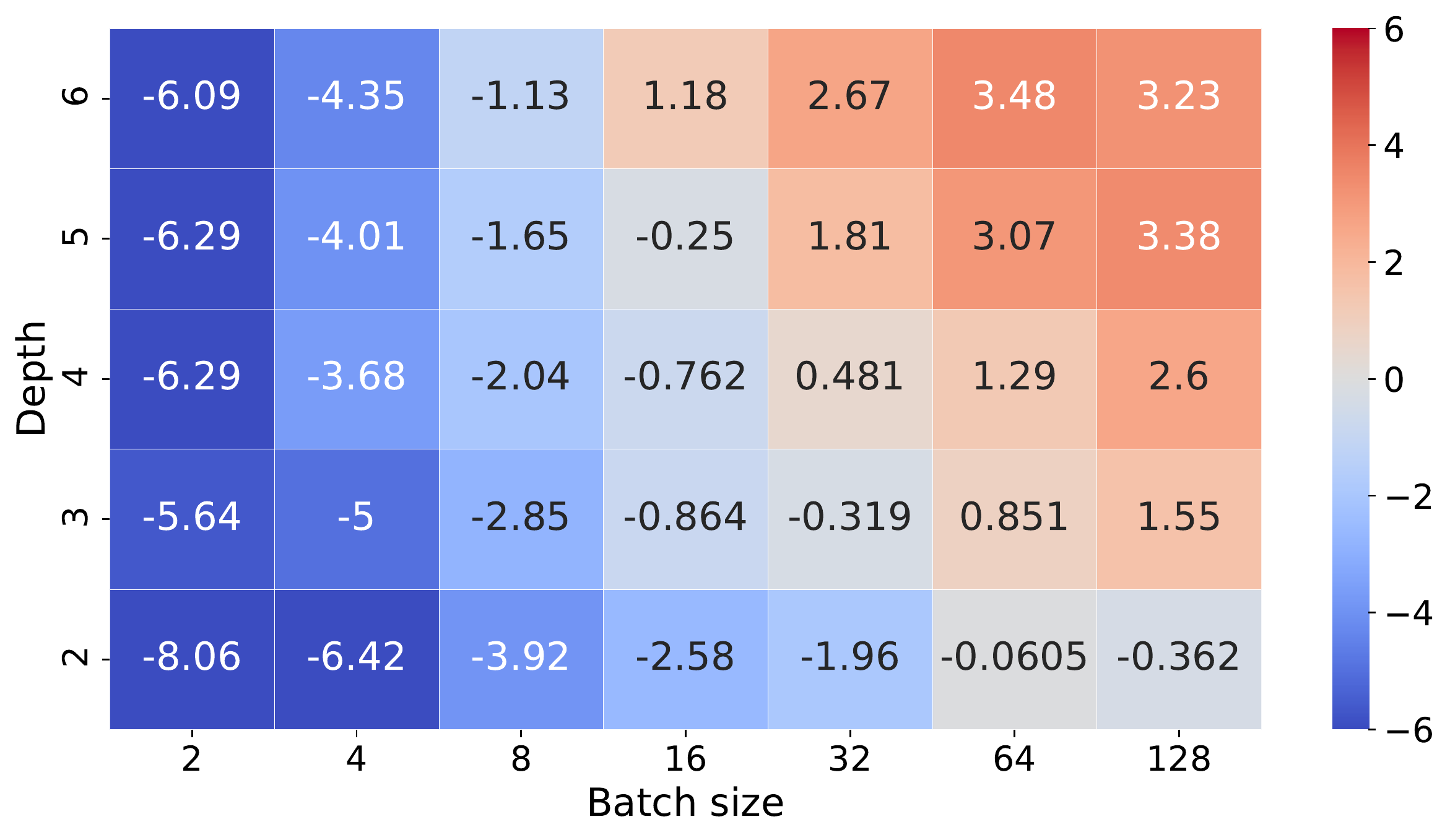}
%\hspace{0.15in}		\subfigure[Validation]{
%		\begin{minipage}[c]{.43\linewidth}
%			\centering
%			\includegraphics[width=6.2cm]{./figures/CNN/test_w16_DeltaBN.pdf}
%		\end{minipage}
%	}	
	\vspace{-0.05in}
	\caption{Effects of batch size for BN on CNNs for CIFAR-10 classification. We vary the batch size of BN, and evaluate the difference in training accuracy between BN and `Base'. }
	\label{fig:sup_CNN_debug_BN}
	%\vspace{-0.16in}
\end{figure}

Figure~\ref{fig:sup_CNN_debug_BN} gives the results of BN with varying batch size, wher we report the difference in accuracy between BN and `Base'. We observe that: 1) BN has significantly degenerated performance when the batch size is too small, \eg, BN has worse performance than `Base' when $m=2$ or $m=4$; 2) The net gain of BN over `Base' is amplified as the depth increases. These observations suggest that there is also a trade-off for BN between the benefits of normalization on optimization and its constraints on representation.

\begin{table*}[t]
	\centering
	%\vspace{-0.15in}
	%\begin{footnotesize}
	\begin{tabular}{c|lllll}
		\bottomrule[1pt]
		& ~~S1   & S1-B1 &  S1-B2 &  S1-B3 & S1-B12\\
		\hline
		Baseline (BN) & ~~419 & ~~419 & ~~419 & ~~419& ~~419  \\
		%BW$_\Sigma$~\cite{2020_CVPR_Huang}     & ~~436 $_{(\textcolor{blue}{\Delta 4.1\%})}$ & ~~578 $_{(\textcolor{blue}{\Delta 37.9\%})}$		&~~565 $_{(\textcolor{blue}{\Delta 34.8\%})}$  &~~998 $_{(\textcolor{blue}{\Delta 138\%})}$ &~~730 $_{(\textcolor{blue}{\Delta 74.2\%})}$\\
		GW &~~437 $_{(\textcolor{blue}{\Delta 4.3\%})}$ &~~518 $_{(\textcolor{blue}{\Delta 23.6\%})}$  &~~514 $_{(\textcolor{blue}{\Delta 22.7\%})}$ &~~634 $_{(\textcolor{blue}{\Delta 51.3\%})}$ &~~589 $_{(\textcolor{blue}{\Delta 40.6\%})}$   \\
		\toprule[1pt]
	\end{tabular}
	\caption{Time costs ($ms$) of five architectures when applying GW on ResNet-50 (\textbf{S1}, \textbf{S1-B1}, \textbf{S1-B2}, \textbf{S1-B3} and \textbf{S1-B12}). Note that $\textcolor{blue}{\Delta~ x\%}$ indicates the additional time cost is $x\%$, compared to the baseline.}
	\label{table-sup:ImageNet-TimeCost-ablation}
	\vspace{-0.1in}
	%	\vspace{-0.1in}
	%	\end{footnotesize}
\end{table*}

\section{More Experimental Results on ImageNet}

\subsection{Learning Decorrelated Feature Representations}
\label{sup:exp-DF}
%\subsection{Learning decorrelated representation}
As described in Section~\ref{subsec_imagenet_ablation} of the paper, we investigate the effect of inserting a GW/BW layer after the last average pooling (before the last linear layer) to learn the decorrelated feature representations, as proposed in \cite{2019_CVPR_Huang}. We provide the results in Table~\ref{table-sup:ImageNet-Res50-ablation}. This can slightly improve the performance ($0.10\%$ on average) when using GW (comparing Table~\ref{table-sup:ImageNet-Res50-ablation} to Table~\ref{table:ImageNet-Res50-ablation} of the paper). We note that $BW_{\Sigma}$ benefits the most from this kind of architecture.
%, with a average gain of $0.43\%$ (comparing Table~\ref{table-sup:ImageNet-Res50-ablation} to Table 2 of the paper)

\begin{table*}[t]
	\centering
	%\begin{footnotesize}
	\begin{tabular}{c|llll}
		\bottomrule[1pt]
		Method     & ResNet-50   &  ResNet-101 &  ResNeXt-50 &  ResNeXt-101\\
		\hline
		Baseline (BN)~\cite{2015_ICML_Ioffe} & ~~419 & ~~672 & ~~574 & ~~912  \\
		%GN~\cite{2018_ECCV_Wu}   & ~~526 $_{(\textcolor{blue}{\Delta 25.3\%})}$ & ~~822 $_{(\textcolor{blue}{\Delta 22.3\%})}$     & ~~720 $_{(\textcolor{blue}{\Delta 25.4\%})}$  &~~1061 $_{(\textcolor{blue}{\Delta 16.3\%})}$   \\	
		BW$_\Sigma$~\cite{2019_CVPR_Huang}     & ~~550 $_{(\textcolor{blue}{\Delta 31.3\%})}$ & ~~882 $_{(\textcolor{blue}{\Delta 30.9\%})}$
		&~~798 $_{(\textcolor{blue}{\Delta 39.0\%})}$  &~~1587 $_{(\textcolor{blue}{\Delta 74.0\%})}$ \\
		GW     & ~~514 $_{(\textcolor{blue}{\Delta 22.7\%})}$  & ~~810 $_{(\textcolor{blue}{\Delta 20.5\%})}$
		&~~738 $_{(\textcolor{blue}{\Delta 28.6\%})}$ &~~1180 $_{(\textcolor{blue}{\Delta 29.3\%})}$  \\
		\toprule[1pt]
	\end{tabular}
	\caption{Time costs ($ms$) of ResNets~\cite{2015_CVPR_He} and ResNeXts~\cite{2017_CVPR_Xie} for ImageNet classification. Note that $\textcolor{blue}{\Delta~ x\%}$ indicates the additional time cost is $x\%$, compared to the baselines.}
	%Here, we apply GW and BW$_\Sigma$ in these models following the \textbf{S1-B2} architecture.}
	\label{table-sup:ImageNet-TimeCost-arch}
	\vspace{-0.1in}
	%\vspace{-0.15in}
	%	\end{footnotesize}
\end{table*}

\subsection{Running Time Comparison}
\label{sup:exp-Time}
In this section, we compare the wall-clock time of the models described in Section~\ref{subsec_imagenet} of the paper.
We run the experiments on GPUs (NVIDIA Tesla V100).
All implementations are based on the API provided by PyTorch, with CUDA (version number: 9.0).
We use the same experimental setup as described in Section~\ref{subsec_imagenet} of  the paper.
We evaluate the training time for each iteration, averaged over 100 iterations.
The ResNets-50 baseline (BN) costs $419~ms$. Replacing the BNs of ResNet-50 with our GWs ($g$=64) costs $796~ms$, a $90\%$ additional time cost on ResNet-50. This is one factor that drives us to investigate the position at which to apply GW.

Table~\ref{table-sup:ImageNet-TimeCost-ablation} shows the time costs of five architectures, \textbf{S1}, \textbf{S1-B1}, \textbf{S1-B2}, \textbf{S1-B3} and \textbf{S1-B12}, which have 1, 17, 17, 17 and 33 GW modules, respectively. Note that applying GW in the \textbf{S1-B3} architecture results in a clearly increased computational cost, compared to  \textbf{S1-B1}/\textbf{S1-B2}. This is because the channel number of the third normalization layer is $4\times$ larger than that of the first/second normalization layer, in the bottleneck blocks~\cite{2015_CVPR_He}.

Table~\ref{table-sup:ImageNet-TimeCost-arch} shows the time costs of ResNets~\cite{2015_CVPR_He} and ResNeXts~\cite{2017_CVPR_Xie} (the corresponding models in Table~\ref{table:ImageNet-Res-Step} of the paper) for ImageNet classification.

\subsection{Results on Advanced Training Strategies}
\label{sup:exp-andvanceTraining}
In Section~\ref{subsec_imagenet_larger} of the paper, we show the effectiveness of our GW on ResNets~\cite{2015_CVPR_He} and ResNeXts~\cite{2017_CVPR_Xie}, under the standard training strategy (\eg, using learning rate step decay). Here, we also conduct experiments using more advanced training strategies: 1) We train over 100 epochs with cosine learning rate decay~\cite{2017_ICLR_Loshchilov}; 2) We add the label smoothing tricks with a smoothing constant $\varepsilon=0.1$~\cite{2019_CVPR_He}; 3) We use mixup training with $\alpha=0.2$ in the Beta distribution~\cite{2018_ICLR_Zhang}. The results are shown in Table~\ref{table:sup-ImageNet-Res-Cos}, where GW improves the baselines consistently.

%\begin{table*}[t]
%	\centering
%	%\begin{footnotesize}
%	\begin{tabular}{c|llll}
%		\bottomrule[1pt]
%		Method     & ResNet-50   &  ResNet-101 &  ResNeXt-50 &  ResNeXt-101\\
%		\hline
%		Baseline (BN)~\cite{2015_ICML_Ioffe} & ~~77.16 & ~~79.13 & ~~78.84 & ~~80.93  \\
%		GN~\cite{2018_ECCV_Wu}   & ~~76.09 $_{(\textcolor{red}{\downarrow 1.07 })}$ & ~~78.29 $_{(\textcolor{red}{\downarrow 0.84 })}$
%		& ~~76.90 $_{(\textcolor{red}{\downarrow 1.94})}$  &~~79.71 $_{(\textcolor{red}{\downarrow 1.22})}$   \\	
%		BW$_\Sigma$~\cite{2020_CVPR_Huang}     & ~~78.29 $_{(\textcolor{blue}{\uparrow 1.13})}$ & ~~79.73 $_{(\textcolor{blue}{\uparrow 0.60})}$
%		&~~79.55 $_{(\textcolor{blue}{\uparrow 0.71})}$  &~~-- $_{(\textcolor{blue}{\uparrow })}$ \\
%		GW     & ~~\textbf{78.46} $_{(\textcolor{blue}{\uparrow 1.30})}$  & ~~\textbf{80.12} $_{(\textcolor{blue}{\uparrow 0.99 })}$
%		&~~\textbf{80.07} $_{(\textcolor{blue}{\uparrow 1.23})}$ &~~\textbf{81.64} $_{(\textcolor{blue}{\uparrow 0.71})}$  \\
%		\toprule[1pt]
%	\end{tabular}
%	\caption{Comparison of validation accuracy on ResNets~\cite{2015_CVPR_He} and ResNeXts~\cite{2017_CVPR_Xie} for ImageNet on more advanced training strategies. Note that we use an additional layer for BW$_\Sigma$ to learn the decorrelated feature, as recommended in \cite{2020_CVPR_Huang}.}
%	%Here, we apply GW and BW$_\Sigma$ in these models following the \textbf{S1-B2} architecture.}
%	\label{table:ImageNet-Res-Step}
%	%\vspace{-0.1in}
%	%	\end{footnotesize}
%\end{table*}

\section{Additional Experiments on Neural Machine Translation}
\label{sup:exp-NLP}
%\TODO{Experiment description and WMT datasets}
 Our GW does indeed generalize layer normalization (LN), which is a widely used technique in NLP tasks. We thus believe our GW has the potential to improve the performance of LN in NLP tasks.
 We conduct additional experiments to apply our GW on Transformer~\cite{2017_NIPS_Vaswani} (where LN is the default normalization) for machine translation tasks using \textit{fairseq-py}~\cite{2019_fairseq_Ott}. We evaluate on the public   IWSLT14 German-to-English (De-EN) dataset using BLEU (higher is better). We use the  hyper-parameters recommended  in \textit{fairseq-py}~\cite{2019_fairseq_Ott} for Transformer and train over 50 epochs with five random seeds. The baseline LN has a BLEU score of $35.02 \pm 0.09$.  GW  (replacing all the LNs with GWs) has a BLEU score of $35.27\pm0.06$. Note that the  hyperparameters were designed for LN, and may not be optimal for GW.

\begin{table}[t]
	\centering
	%\begin{footnotesize}
	\begin{tabular}{c|ll}
		\bottomrule[1pt]
		Method     & ResNet-50   &   ResNeXt-50 \\
		\hline
		Baseline (BN)~\cite{2015_ICML_Ioffe} & ~~77.16 &  ~~78.84   \\
		GN~\cite{2018_ECCV_Wu}   & ~~76.09 $_{(\textcolor{red}{\downarrow 1.07 })}$
		& ~~76.90 $_{(\textcolor{red}{\downarrow 1.94})}$     \\	
		BW$_\Sigma$~\cite{2020_CVPR_Huang}     & ~~78.29 $_{(\textcolor{blue}{\uparrow 1.13})}$
		&~~79.55 $_{(\textcolor{blue}{\uparrow 0.71})}$   \\
		GW     & ~~\textbf{78.46} $_{(\textcolor{blue}{\uparrow 1.30})}$
		&~~\textbf{80.07} $_{(\textcolor{blue}{\uparrow 1.23})}$ \\
		\toprule[1pt]
	\end{tabular}
	\caption{Comparison of validation accuracy on 50-layer ResNet~\cite{2015_CVPR_He} and ResNeXt~\cite{2017_CVPR_Xie} for ImageNet on more advanced training strategies (\eg,  cosine learning rate decay~\cite{2017_ICLR_Loshchilov}, label smoothing~\cite{2019_CVPR_He} and mixup~\cite{2018_ICLR_Zhang}). Note that we use an additional layer for BW$_\Sigma$ to learn the decorrelated feature, as recommended in \cite{2020_CVPR_Huang}.}
	%Here, we apply GW and BW$_\Sigma$ in these models following the \textbf{S1-B2} architecture.}
	\label{table:sup-ImageNet-Res-Cos}
	%\vspace{-0.1in}
	%	\end{footnotesize}
\end{table}

%% file: CVPR_GW_Camera.bbl
\begin{thebibliography}{10}\itemsep=-1pt

\bibitem{2019_ICLR_Arora}
Sanjeev Arora, Zhiyuan Li, and Kaifeng Lyu.
\newblock Theoretical analysis of auto rate-tuning by batch normalization.
\newblock In {\em ICLR}, 2019.

\bibitem{2018_CoRR_Andrei}
Andrei Atanov, Arsenii Ashukha, Dmitry Molchanov, Kirill Neklyudov, and Dmitry
  Vetrov.
\newblock Uncertainty estimation via stochastic batch normalization.
\newblock In {\em ICLR Workshop}, 2018.

\bibitem{2018_ICML_Azizpour}
Hossein Azizpour, Mattias Teye, and Kevin Smith.
\newblock Bayesian uncertainty estimation for batch normalized deep networks.
\newblock In {\em International Conference on Machine Learning (ICML)}, 2018.

\bibitem{2016_CoRR_Ba}
Lei~Jimmy Ba, Ryan Kiros, and Geoffrey~E. Hinton.
\newblock Layer normalization.
\newblock {\em arXiv preprint arXiv:1607.06450}, 2016.

\bibitem{2005_NumerialAlg}
Dario~A. Bini, Nicholas~J. Higham, and Beatrice Meini.
\newblock Algorithms for the matrix pth root.
\newblock {\em Numerical Algorithms}, 39(4):349--378, Aug 2005.

\bibitem{2018_NIPS_Bjorck}
Johan Bjorck, Carla Gomes, and Bart Selman.
\newblock Understanding batch normalization.
\newblock In {\em NeurIPS}, 2018.

\bibitem{2020_arxiv_Cai}
Tianle Cai, Shengjie Luo, Keyulu Xu, Di He, Tie-yan Liu, and Liwei Wang.
\newblock Graphnorm: A principled approach to accelerating graph neural network
  training.
\newblock {\em arXiv preprint arXiv:2009.03294}, 2020.

\bibitem{2019_ICML_Cai}
Yongqiang Cai, Qianxiao Li, and Zuowei Shen.
\newblock A quantitative analysis of the effect of batch normalization on
  gradient descent.
\newblock In {\em ICML}, 2019.

\bibitem{2017_NIPS_Cho}
Minhyung Cho and Jaehyung Lee.
\newblock Riemannian approach to batch normalization.
\newblock In {\em NeurIPS}, 2017.

\bibitem{2016_CoRR_Cooijmans}
Tim Cooijmans, Nicolas Ballas, C{\'{e}}sar Laurent, and Aaron~C. Courville.
\newblock Recurrent batch normalization.
\newblock In {\em ICLR}, 2017.

\bibitem{2020_arxiv_Daneshmand}
Hadi Daneshmand, Jonas Kohler, Francis Bach, Thomas Hofmann, and Aurelien
  Lucchi.
\newblock Theoretical understanding of batch-normalization: A markov chain
  perspective.
\newblock {\em arXiv preprint arXiv:2003.01652}, 2020.

\bibitem{2015_NIPS_Desjardins}
Guillaume Desjardins, Karen Simonyan, Razvan Pascanu, and koray kavukcuoglu.
\newblock Natural neural networks.
\newblock In {\em NeurIPS}, 2015.

\bibitem{2019_ICML_Ghorbani}
Behrooz Ghorbani, Shankar Krishnan, and Ying Xiao.
\newblock An investigation into neural net optimization via hessian eigenvalue
  density.
\newblock In {\em ICML}, 2019.

\bibitem{2017_ICCV_He}
Kaiming He, Georgia Gkioxari, Piotr Doll{\'{a}}r, and Ross~B. Girshick.
\newblock Mask {R-CNN}.
\newblock In {\em ICCV}, 2017.

\bibitem{2015_CVPR_He}
Kaiming He, Xiangyu Zhang, Shaoqing Ren, and Jian Sun.
\newblock Deep residual learning for image recognition.
\newblock In {\em CVPR}, 2016.

\bibitem{2019_CVPR_He}
Tong He, Zhi Zhang, Hang Zhang, Zhongyue Zhang, Junyuan Xie, and Mu Li.
\newblock Bag of tricks for image classification with convolutional neural
  networks.
\newblock In {\em CVPR}, 2019.

\bibitem{2018_arxiv_Hoffer}
Elad Hoffer, Ron Banner, Itay Golan, and Daniel Soudry.
\newblock Norm matters: efficient and accurate normalization schemes in deep
  networks.
\newblock In {\em NeurIPS}, 2018.

\bibitem{2016_CoRR_Huang_a}
Gao Huang, Zhuang Liu, and Kilian~Q. Weinberger.
\newblock Densely connected convolutional networks.
\newblock In {\em CVPR}, 2017.

\bibitem{2020_ECCV_Huang}
Lei Huang, Jie Qin, Li Liu, Fan Zhu, and Ling Shao.
\newblock Layer-wise conditioning analysis in exploring the learning dynamics
  of dnns.
\newblock In {\em ECCV}, 2020.

\bibitem{2018_CVPR_Huang}
Lei Huang, Dawei Yang, Bo Lang, and Jia Deng.
\newblock Decorrelated batch normalization.
\newblock In {\em CVPR}, 2018.

\bibitem{2020_CVPR_Huang}
Lei Huang, Lei Zhao, Yi Zhou, Fan Zhu, Li Liu, and Ling Shao.
\newblock An investigation into the stochasticity of batch whitening.
\newblock In {\em CVPR}, 2020.

\bibitem{2019_CVPR_Huang}
Lei Huang, Yi Zhou, Fan Zhu, Li Liu, and Ling Shao.
\newblock Iterative normalization: Beyond standardization towards efficient
  whitening.
\newblock In {\em CVPR}, 2019.

\bibitem{2017_NIPS_Ioffe}
Sergey Ioffe.
\newblock Batch renormalization: Towards reducing minibatch dependence in
  batch-normalized models.
\newblock In {\em NeurIPS}, 2017.

\bibitem{2015_ICML_Ioffe}
Sergey Ioffe and Christian Szegedy.
\newblock Batch normalization: Accelerating deep network training by reducing
  internal covariate shift.
\newblock In {\em ICML}, 2015.

\bibitem{2019_arxiv_Jacot}
Arthur Jacot, Franck Gabriel, and Cl{\'e}ment Hongler.
\newblock Freeze and chaos for dnns: an ntk view of batch normalization,
  checkerboard and boundary effects.
\newblock {\em arXiv preprint arXiv:1907.05715}, 2019.

\bibitem{2019_NIPS_Karakida}
Ryo Karakida, Shotaro Akaho, and Shun-ichi Amari.
\newblock The normalization method for alleviating pathological sharpness in
  wide neural networks.
\newblock In {\em NeurIPS}. 2019.

\bibitem{2018_AS_Kessy}
Agnan Kessy, Alex Lewin, and Korbinian Strimmer.
\newblock Optimal whitening and decorrelation.
\newblock {\em The American Statistician}, 72(4):309--314, 2018.

\bibitem{2019_AISTATS_Kohler}
Jonas Kohler, Hadi Daneshmand, Aurelien Lucchi, Thomas Hofmann, Ming Zhou, and
  Klaus Neymeyr.
\newblock Exponential convergence rates for batch normalization: The power of
  length-direction decoupling in non-convex optimization.
\newblock In {\em AISTATS}, 2019.

\bibitem{2020_ECCV_Kolesnikov}
Alexander Kolesnikov, Lucas Beyer, Xiaohua Zhai, Joan Puigcerver, Jessica Yung,
  Sylvain Gelly, and Neil Houlsby.
\newblock Big transfer (bit): General visual representation learning.
\newblock In {\em ECCV}, 2020.

\bibitem{2016_ICASSP_Laurent}
C{\'{e}}sar Laurent, Gabriel Pereyra, Philemon Brakel, Ying Zhang, and Yoshua
  Bengio.
\newblock Batch normalized recurrent neural networks.
\newblock In {\em ICASSP}, 2016.

\bibitem{1998_NN_Yann}
Yann LeCun, L{\'e}on Bottou, Genevieve~B. Orr, and Klaus-Robert M\"{u}ller.
\newblock Effiicient backprop.
\newblock In {\em Neural Networks: Tricks of the Trade}, pages 9--50, 1998.

\bibitem{2020_ICLR_Li}
Zhiyuan Li and Sanjeev Arora.
\newblock An exponential learning rate schedule for batch normalized networks.
\newblock In {\em ICLR}, 2020.

\bibitem{2019_AISTATS_Lian}
Xiangru Lian and Ji Liu.
\newblock Revisit batch normalization: New understanding and refinement via
  composition optimization.
\newblock In {\em AISTATS}, 2019.

\bibitem{2017_CVPR_Lin}
Tsung{-}Yi Lin, Piotr Doll{\'{a}}r, Ross~B. Girshick, Kaiming He, Bharath
  Hariharan, and Serge~J. Belongie.
\newblock Feature pyramid networks for object detection.
\newblock In {\em CVPR}, 2017.

\bibitem{2014_ECCV_COCO}
Tsung-Yi Lin, Michael Maire, Serge Belongie, James Hays, Pietro Perona, Deva
  Ramanan, Piotr Doll{\'a}r, and C.~Lawrence Zitnick.
\newblock Microsoft coco: Common objects in context.
\newblock In {\em ECCV}, 2014.

\bibitem{2017_ICLR_Loshchilov}
Ilya Loshchilov and Frank Hutter.
\newblock {SGDR:} stochastic gradient descent with restarts.
\newblock In {\em ICLR}, 2017.

\bibitem{2018_arxiv_luo}
Ping Luo, Jiamin Ren, and Zhanglin Peng.
\newblock Differentiable learning-to-normalize via switchable normalization.
\newblock {\em arXiv preprint arXiv:1806.10779}, 2018.

\bibitem{2019_ICLR_Luo2}
Ping Luo, Xinjiang Wang, Wenqi Shao, and Zhanglin Peng.
\newblock Towards understanding regularization in batch normalization.
\newblock In {\em ICLR}, 2019.

\bibitem{massa2018mrcnn}
Francisco Massa and Ross Girshick.
\newblock {maskrcnn-benchmark: Fast, modular reference implementation of
  Instance Segmentation and Object Detection algorithms in PyTorch}.
\newblock \url{https://github.com/facebookresearch/maskrcnn-benchmark}, 2018.
\newblock Accessed: 09-26-2019.

\bibitem{2014_NeurIPS_Montufar}
Guido~F Montufar, Razvan Pascanu, Kyunghyun Cho, and Yoshua Bengio.
\newblock On the number of linear regions of deep neural networks.
\newblock In {\em NeurIPS}, 2014.

\bibitem{2019_fairseq_Ott}
Myle Ott, Sergey Edunov, Alexei Baevski, Angela Fan, Sam Gross, Nathan Ng,
  David Grangier, and Michael Auli.
\newblock fairseq: A fast, extensible toolkit for sequence modeling.
\newblock In {\em Proceedings of NAACL-HLT 2019: Demonstrations}, 2019.

\bibitem{2019_ICCV_Pan}
Xingang Pan, Xiaohang Zhan, Jianping Shi, Xiaoou Tang, and Ping Luo.
\newblock Switchable whitening for deep representation learning.
\newblock In {\em ICCV}, 2019.

\bibitem{2017_NIPS_pyTorch}
Adam Paszke, Sam Gross, Soumith Chintala, Gregory Chanan, Edward Yang, Zachary
  DeVito, Zeming Lin, Alban Desmaison, Luca Antiga, and Adam Lerer.
\newblock Automatic differentiation in {PyTorch}.
\newblock In {\em NeurIPS Autodiff Workshop}, 2017.

\bibitem{2017_ICLR_Ren}
Mengye Ren, Renjie Liao, Raquel Urtasun, Fabian~H. Sinz, and Richard~S. Zemel.
\newblock Normalizing the normalizers: Comparing and extending network
  normalization schemes.
\newblock In {\em ICLR}, 2017.

\bibitem{2015_NIPS_Ren}
Shaoqing Ren, Kaiming He, Ross Girshick, and Jian Sun.
\newblock Faster {R-CNN}: Towards real-time object detection with region
  proposal networks.
\newblock In {\em NeurIPS}, 2015.

\bibitem{2015_IJCV_ImageNet}
Olga Russakovsky, Jia Deng, Hao Su, Jonathan Krause, Sanjeev Satheesh, Sean Ma,
  Zhiheng Huang, Andrej Karpathy, Aditya Khosla, Michael Bernstein,
  Alexander~C. Berg, and Li Fei-Fei.
\newblock {ImageNet Large Scale Visual Recognition Challenge}.
\newblock {\em International Journal of Computer Vision (IJCV)},
  115(3):211--252, 2015.

\bibitem{2018_NIPS_shibani}
Shibani Santurkar, Dimitris Tsipras, Andrew Ilyas, and Aleksander Madry.
\newblock How does batch normalization help optimization?
\newblock In {\em NeurIPS}, 2018.

\bibitem{2020_arxiv_Shao}
Wenqi Shao, Shitao Tang, Xingang Pan, Ping Tan, Xiaogang Wang, and Ping Luo.
\newblock Channel equilibrium networks for learning deep representation.
\newblock In {\em ICML}, 2020.

\bibitem{2018_ACCV_Alexander}
Alexander Shekhovtsov and Boris Flach.
\newblock Stochastic normalizations as bayesian learning.
\newblock In {\em ACCV}, 2018.

\bibitem{2019_ICLR_Siarohin}
Aliaksandr Siarohin, Enver Sangineto, and Nicu Sebe.
\newblock Whitening and coloring batch transform for gans.
\newblock In {\em ICLR}, 2019.

\bibitem{2019_ICCV_Singh}
Saurabh Singh and Abhinav Shrivastava.
\newblock Evalnorm: Estimating batch normalization statistics for evaluation.
\newblock In {\em ICCV}, 2019.

\bibitem{2016_CoRR_Szegedy}
Christian Szegedy, Sergey Ioffe, and Vincent Vanhoucke.
\newblock Inception-v4, inception-resnet and the impact of residual connections
  on learning.
\newblock In {\em AAAI}, 2017.

\bibitem{2015_CoRR_Szegedy}
Christian Szegedy, Vincent Vanhoucke, Sergey Ioffe, Jonathon Shlens, and
  Zbigniew Wojna.
\newblock Rethinking the inception architecture for computer vision.
\newblock In {\em CVPR}, 2016.

\bibitem{2018_ICML_Teye}
Mattias Teye, Hossein Azizpour, and Kevin Smith.
\newblock {B}ayesian uncertainty estimation for batch normalized deep networks.
\newblock In {\em ICML}, 2018.

\bibitem{1999_TNN_Vapnik}
Vladimir~N Vapnik.
\newblock An overview of statistical learning theory.
\newblock {\em IEEE transactions on neural networks}, 10(5):988--999, 1999.

\bibitem{2017_NIPS_Vaswani}
Ashish Vaswani, Noam Shazeer, Niki Parmar, Jakob Uszkoreit, Llion Jones,
  Aidan~N Gomez, Lukasz Kaiser, and Illia Polosukhin.
\newblock Attention is all you need.
\newblock In {\em NeurIPS}, 2017.

\bibitem{2018_NIPS_Wang}
Guangrun Wang, Jiefeng Peng, Ping Luo, Xinjiang Wang, and Liang Lin.
\newblock Kalman normalization: Normalizing internal representations across
  network layers.
\newblock In {\em NeurIPS}, 2018.

\bibitem{2019_arxiv_Wei}
Mingwei Wei, James Stokes, and David~J. Schwab.
\newblock Mean-field analysis of batch normalization.
\newblock {\em arXiv preprint arXiv:1903.02606}, 2019.

\bibitem{2011_NIPS_Wiesler}
Simon Wiesler and Hermann Ney.
\newblock A convergence analysis of log-linear training.
\newblock In {\em {NeurIPS}}, 2011.

\bibitem{2018_ECCV_Wu}
Yuxin Wu and Kaiming He.
\newblock Group normalization.
\newblock In {\em ECCV}, 2018.

\bibitem{2017_CVPR_Xie}
Saining Xie, Ross~B. Girshick, Piotr Doll{\'{a}}r, Zhuowen Tu, and Kaiming He.
\newblock Aggregated residual transformations for deep neural networks.
\newblock In {\em CVPR}, 2017.

\bibitem{2020_ICML_Xionghuan}
Huan Xiong, Lei Huang, Mengyang Yu, Li Liu, Fan Zhu, and Ling Shao.
\newblock On the number of linear regions of convolutional neural networks.
\newblock In {\em ICML}, 2020.

\bibitem{2019_NIPS_Xu}
Jingjing Xu, Xu Sun, Zhiyuan Zhang, Guangxiang Zhao, and Junyang Lin.
\newblock Understanding and improving layer normalization.
\newblock In {\em NeurIPS}, 2019.

\bibitem{2019_ICLR_Yang}
Greg Yang, Jeffrey Pennington, Vinay Rao, Jascha Sohl{-}Dickstein, and
  Samuel~S. Schoenholz.
\newblock A mean field theory of batch normalization.
\newblock In {\em ICLR}, 2019.

\bibitem{2020_ICLR_Ye}
Chengxi Ye, Matthew Evanusa, Hua He, Anton Mitrokhin, Tom Goldstein, James~A.
  Yorke, Cornelia Fermuller, and Yiannis Aloimonos.
\newblock Network deconvolution.
\newblock In {\em ICLR}, 2020.

\bibitem{2018_ICLR_Yu}
Adams~Wei Yu, David Dohan, Minh-Thang Luong, Rui Zhao, Kai Chen, Mohammad
  Norouzi, and Quoc~V Le.
\newblock Qanet: Combining local convolution with global self-attention for
  reading comprehension.
\newblock In {\em ICLR}, 2018.

\bibitem{2016_CoRR_Zagoruyko}
Sergey Zagoruyko and Nikos Komodakis.
\newblock Wide residual networks.
\newblock In {\em BMVC}, 2016.

\bibitem{2017_ICLR_Zhang}
Chiyuan Zhang, Samy Bengio, Moritz Hardt, Benjamin Recht, and Oriol Vinyals.
\newblock Understanding deep learning requires rethinking generalization.
\newblock In {\em ICLR}, 2017.

\bibitem{2019_ICLR_Zhang}
Guodong Zhang, Chaoqi Wang, Bowen Xu, and Roger~B. Grosse.
\newblock Three mechanisms of weight decay regularization.
\newblock In {\em ICLR}, 2019.

\bibitem{2018_ICLR_Zhang}
Hongyi Zhang, Moustapha Cisse, Yann~N. Dauphin, and David Lopez-Paz.
\newblock mixup: Beyond empirical risk minimization.
\newblock In {\em ICLR}, 2018.

\bibitem{2020_CVPR_Zhang}
Ruimao Zhang, Zhanglin Peng, Lingyun Wu, Zhen Li, and Ping Luo.
\newblock Exemplar normalization for learning deep representation.
\newblock In {\em CVPR}, 2020.

\end{thebibliography}
